\documentclass{article}

\PassOptionsToPackage{numbers,sort&compress}{natbib}
\usepackage[preprint]{neurips_2026}
\usepackage[utf8]{inputenc}
\usepackage[T1]{fontenc}
\usepackage{hyperref}
\usepackage{url}
\usepackage{booktabs}
\usepackage{subcaption}
\usepackage{amsfonts}
\usepackage{nicefrac}
\usepackage{microtype}
\usepackage{xcolor}
\usepackage{graphicx}
\usepackage{algorithm}
\usepackage{algorithmic}
\usepackage{float}
\usepackage{makecell}
\usepackage{wrapfig}
\usepackage{multicol}
\usepackage{multirow}
\usepackage{bm}
\usepackage{textcomp}
\usepackage{amsmath}
\usepackage{amssymb}
\usepackage{mathtools}
\usepackage{amsthm}
\usepackage[capitalize,noabbrev]{cleveref}
\DeclareMathOperator{\rank}{rank}
\DeclareMathOperator{\Ker}{Ker}
\DeclareMathOperator{\pdet}{pdet}
\DeclareMathOperator{\Img}{Img}
\DeclareMathOperator{\Proj}{Proj}

\theoremstyle{plain}
\newtheorem{theorem}{Theorem}[section]
\newtheorem{proposition}[theorem]{Proposition}
\newtheorem{lemma}[theorem]{Lemma}
\newtheorem{corollary}[theorem]{Corollary}
\theoremstyle{definition}

\newtheorem{assumption}[theorem]{Assumption}
\theoremstyle{remark}
\newtheorem{remark}[theorem]{Remark}

\newcommand{\R}{\mathbb{R}}

\title{Determinism of Randomness: Prompt-Residual Seed Shaping for Diffusion Generation}

\author{%
  Song Yan\textsuperscript{1}\And Wei Zhai\textsuperscript{1}\thanks{Corresponding author.}\And Chenfeng Wang\textsuperscript{2}  \And 
  Xinliang Bi\textsuperscript{3} \And 
  Jian Yang\textsuperscript{4} \And 
  Yancheng Cai\textsuperscript{5} \And 
  Yusen Zhang\textsuperscript{3} \And 
  Yunwei Lan\textsuperscript{1} \And 
  Tao Zhang\textsuperscript{6} \And 
  Guanye Xiong\textsuperscript{3} \And 
  Min Li\textsuperscript{3}\footnotemark[1]\And 
  Zheng-Jun Zha\textsuperscript{1} \\
  \\
  \textsuperscript{1}USTC\quad
  \textsuperscript{2}Li Auto Inc.\quad
  \textsuperscript{3}Xi'an High-tech Research Institute \\
  \textsuperscript{4}Wechat Vision \quad
  \textsuperscript{5}Cambridge University\quad
  \textsuperscript{6}HUST 
   \\ % 如果单位短，可以用 \quad 放在同一行
\texttt{gary\_144@mail.ustc.edu.cn} 
}

\begin{document}
\maketitle

\begin{abstract}
Diffusion models start generation from an isotropic Gaussian latent, yet changing only the random seed can lead to large differences in prompt faithfulness, composition, and visual quality.
We study this seed sensitivity through the semantic map from initial noise to generated meaning.
Although the sampling flow is locally invertible, the subsequent semantic projection is many-to-one, inducing a degenerate pullback semi-metric on the latent space: most local directions are nearly semantic-invariant, while semantic-sensitive variation is concentrated in a much smaller horizontal subspace.
This provides an explanatory geometric view of the seed lottery.
Motivated by this view, we introduce a training-free prompt-residual seed-shaping procedure.
Rather than claiming to recover the exact horizontal space, the method uses a single high-noise cold-start prompt residual as a model-coupled proxy, injects only its tangential component, and retracts the seed to the original Gaussian radius shell.
This keeps the initialization prior-compatible while adding only one conditional/unconditional probe before standard sampling.
Across multiple generation benchmarks, the method improves alignment and quality metrics over standard sampling, supporting both the practical value of the proxy and the explanatory relevance of semantic anisotropy.
\end{abstract}

% =========================================
\section{Introduction}
% =========================================

Diffusion models are increasingly used in settings where generation is expected to be controllable, prompt-faithful, and stable under nuisance factors.
A persistent source of instability is the initial random seed.
With the prompt and model fixed, two seeds of comparable Euclidean norm can yield sharply different semantic correctness, compositional structure, typography, and artifact rates.
This phenomenon, often described as a \emph{Seed Lottery}, leads to repeated sampling, cherry-picking, and inflated evaluation variance.

\begin{wrapfigure}{r}{0.38\textwidth}
    \centering
    \includegraphics[width=0.35\textwidth]{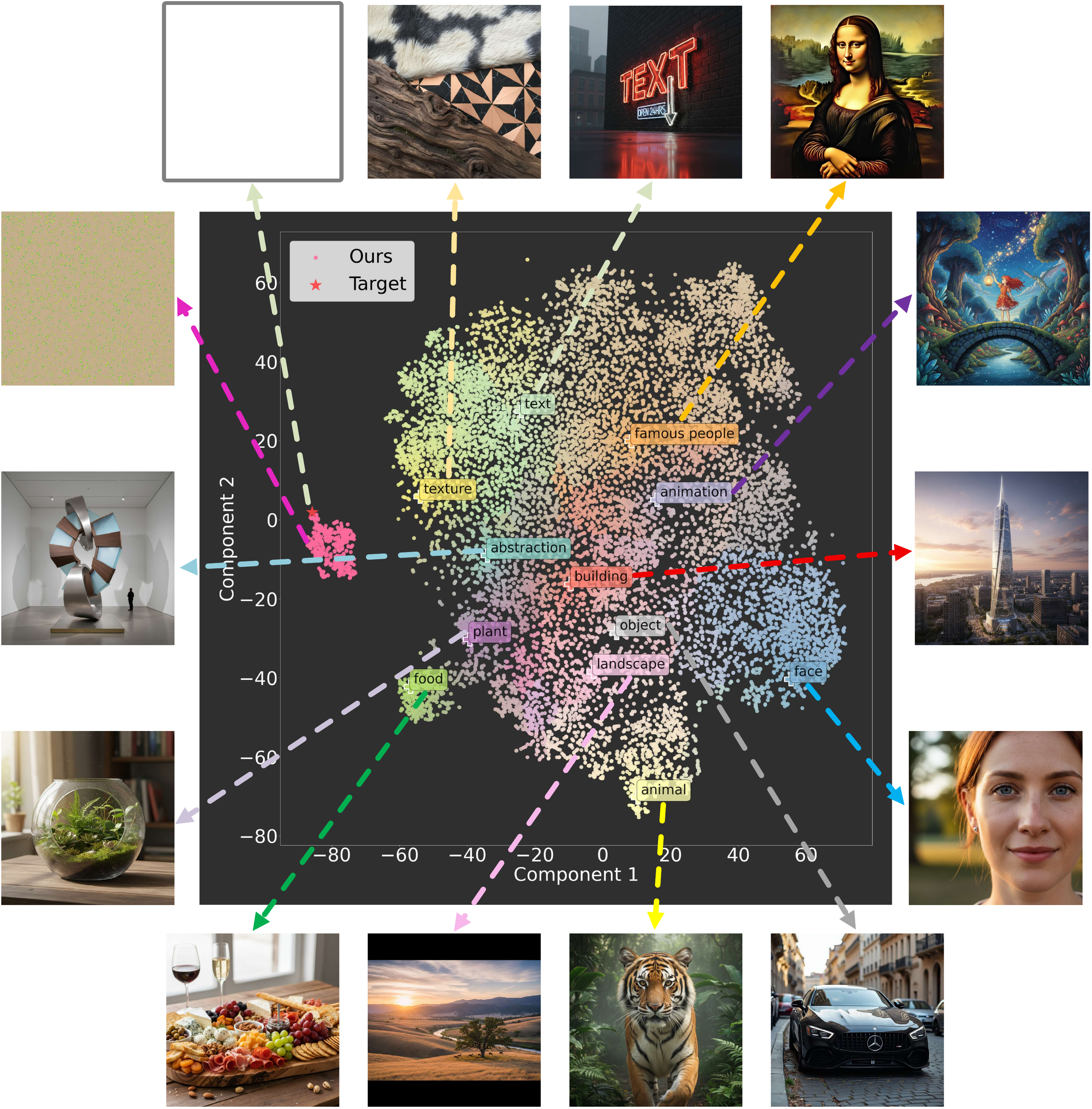}
    \caption{Semantic distribution of 20,000 random noises in FLUX.1 Dev. Details are in Appendix~\ref{VMsspa}.}
    \label{ssp}
\end{wrapfigure}

\textbf{A paradox: isotropic priors, anisotropic outcomes.}
Standard pipelines sample the initial latent from an isotropic Gaussian prior $z_T\sim\mathcal N(0,I)$, which assigns no privileged Euclidean direction.
If this prior geometry were the geometry relevant to semantic behavior, seed perturbations of comparable size would have comparable effects.
In practice, they do not: as illustrated in Figure~\ref{ssp}, some directions induce large semantic changes or failures, whereas many others produce only minor variation.
The issue is therefore not randomness per se, but the mismatch between an isotropic input law and an anisotropic semantic response.
\textbf{Core viewpoint: semantics induces a degenerate latent geometry.}
The sampling flow itself cannot explain this collapse of degrees of freedom.
Under standard local uniqueness conditions for the continuous probability-flow ODE, the map $\Psi_{T\to0}$ is locally invertible on regular intervals.
This continuous flow alone does not explain semantic many-to-one collapse; semantic identification, however, is many-to-one.
We model the effective semantic outcome by
\begin{equation}
    \Phi=\pi\circ\Psi_{T\to 0}:\mathcal Z\cong\mathbb R^d\to\mathcal M,
\end{equation}
where $\pi$ is an operational semantic projection, such as CLIP/DINO features followed by a local chart projection; Appendix~\ref{app:pi_sensitivity} studies the sensitivity to this choice.
On regular regions with $\rank(D\Phi_z)=k\ll d$, the level sets $\Phi^{-1}(x)$ form high-dimensional fibers of semantic-invariant variation.
Pulling back the metric on $\mathcal M$ gives
\begin{equation}
    g_{\mathrm{lat}}=\Phi^\ast g_{\mathcal M},
\end{equation}
which vanishes along fiber directions and is informative only on a low-dimensional horizontal complement.
Thus Euclidean seed distance is not the right measure of semantic stability; local behavior is governed by the spectrum of $D\Phi_z$ on semantic-sensitive directions.

\textbf{From explanatory geometry to an algorithmic approximation.}
The geometric framework identifies the type of direction that can affect semantic outcomes, but it does not make $D\Phi_z$ or the exact horizontal space available to the sampler.
We therefore separate the theoretical object from the implemented update: the algorithm uses the pretrained model's prompt residual as a \emph{model-coupled proxy} for a prompt-sensitive horizontal component, rather than claiming to recover the true horizontal direction.
The estimator uses a single high-noise cold-start probe: near the start of sampling, the scheduler-scaled seed is close to the native trajectory, making this proxy both cheap and stable.

Concretely, our \emph{Prompt-Residual Seed Shaping} procedure has two parts.
First, it estimates a \textbf{cold-start prompt-residual proxy direction} by querying the pretrained model at a high-noise timestep using scheduler-consistent scaling of the initial Gaussian seed.
Second, it applies \textbf{tangential injection and spherical retraction}: the radial component is removed, a small prompt-dependent step is applied, and the result is projected back to the original radius shell. Our contributions are:

1. We systematically derive a degenerate pullback geometry of the semantic map $\Phi=\pi\circ\Psi$, showing that semantic projection creates large semantic-invariant fibers and low-dimensional sensitive directions. 
This provides, to the best of our knowledge, the first geometric explanation of seed lottery under isotropic Gaussian initialization.

2. Guided by this view, we propose prompt-residual seed shaping, a training-free initialization method that uses a single high-noise model probe as a prompt-sensitive proxy direction. 
The method performs tangential injection followed by spherical retraction, preserving the sampled Gaussian radius while suggesting a broader route toward geometry-aware optimization throughout the generation process.

3. We validate both the theory and the algorithm through degeneracy diagnostics, proxy-direction analyses, and targeted ablations. 
Experiments show that the proposed seed shaping improves generation quality and alignment across heterogeneous backbones.

% =========================================
\section{Related Works}
% =========================================
\subsection{Seed Sensitivity, Initial-Condition Interventions, and Latent Geometry}

Diffusion models exhibit a persistent \emph{Seed Lottery} \cite{Qi2024Not,SDXL}.
This sensitivity is often attributed to early denoising and structure commitment, where the initial latent acts as a coarse scaffold for the final generation \cite{parmar2023aliasing,blattmann2023stable}.
Consequently, several works intervene directly on the initial condition, including attention-based initial-noise refinement \cite{DBLP:conf/cvpr/GuoLCLY024} and supervised noise adjustment \cite{zhou2025golden}.

This seed dependence is also related to broader geometric views of generative latent spaces.
Prior work shows that decoder-induced pullback metrics can make latent spaces curved and anisotropic \cite{arvanitidis2018latent}, and Riemannian formulations have been explored for diffusion and flow models on non-Euclidean domains \cite{kim2023riemannian,chen2023stiefel}.
At the representation level, manifold/fiber perspectives separate semantic factors from nuisance variation \cite{parde2023visual}, while Jacobian and dynamical analyses connect local spectra to stability and anisotropy \cite{Biroli2024Dynamical,lim2023geometry}.

\section{Explanatory Geometry: Latent Space Degeneracy}
\label{sec:pre}
\begin{figure}[htbp]
\centering
\includegraphics[width=0.85\linewidth]{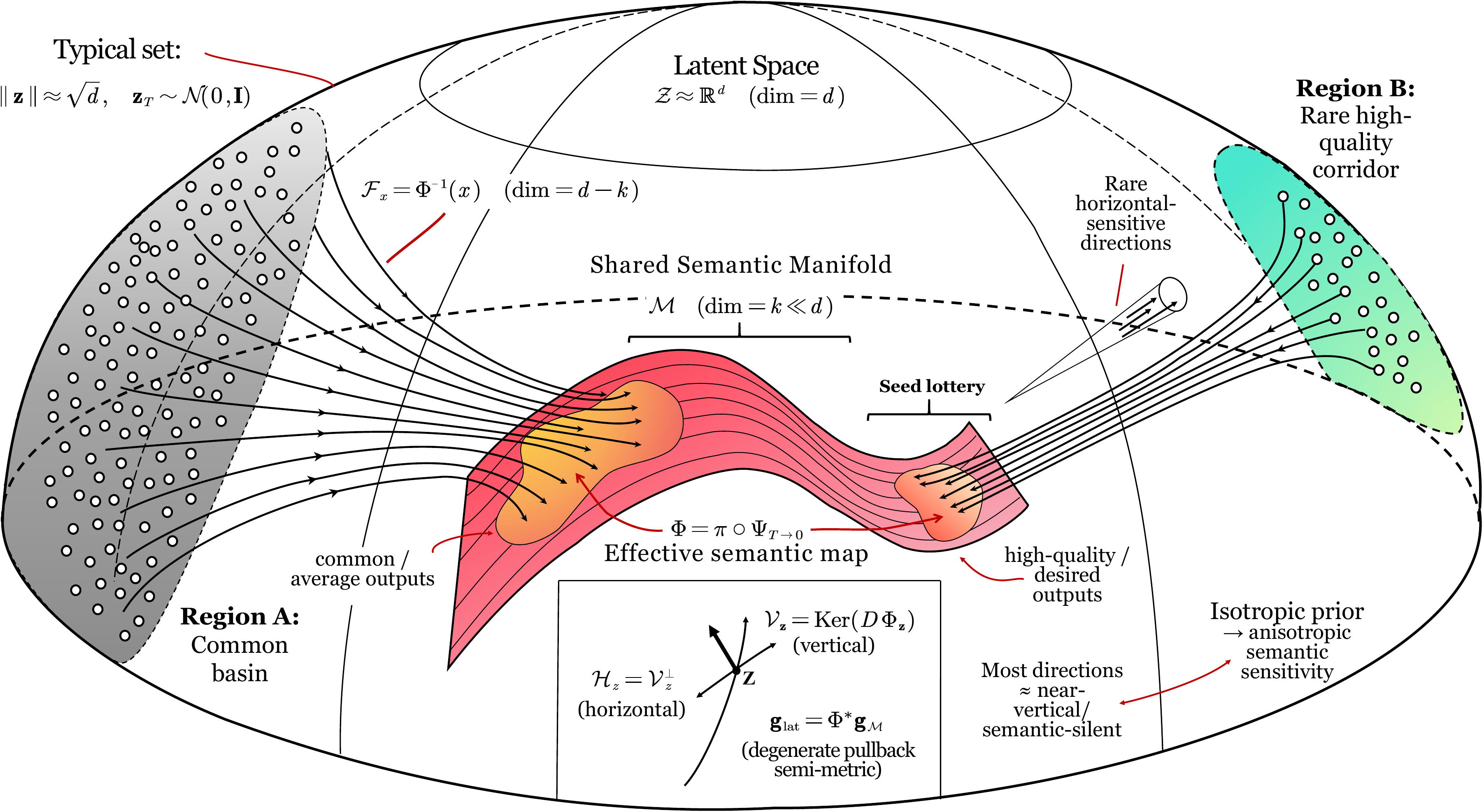}
\caption{\textbf{Geometric overview of latent-space degeneracy.}
Although the sampling flow is locally invertible, the semantic projection is many-to-one. Large latent fibers therefore map to similar semantic outcomes: directions tangent to a fiber are semantic-invariant, while transverse directions control semantic change. This mismatch between an isotropic Gaussian prior and an anisotropic semantic response underlies seed lottery.}
\label{Global}
\end{figure}

\subsection{Semantic map and fiber structure}
\label{sec:semantic_geometry}

Let $z_t$ follow the probability-flow ODE
\begin{equation}
\label{eq:pfode}
\frac{d z_t}{dt}=v_\theta(z_t,t;y),
\qquad z_0=\Psi_{T\to0}(z_T).
\end{equation}
Under standard local Lipschitz regularity, $\Psi_{T\to0}$ is locally invertible on regular intervals. Hence the collapse of degrees of freedom relevant to semantic outcomes does not come from the ODE flow alone, but from the semantic identification applied after generation. We model the effective semantic outcome by
\begin{equation}
\label{eq:semantic_map}
\Phi:=\pi\circ\Psi_{T\to0}:\mathcal Z\simeq\mathbb R^d\to\mathcal M,
\end{equation}
where $\pi$ is an operational differentiable semantic projection, such as CLIP/DINO features followed by a local chart projection.
\begin{assumption}[Regular semantic map]
\label{assump:submersion_main}
Locally in a semantic chart, $\Phi$ is $C^2$ on a dense regular set
$\mathcal Z_{\rm reg}=\{z:\rank(D\Phi_z)=k\}$ with $k\ll d$, and the singular set has negligible Gaussian mass.
When applying coarea, the chart is taken on the effective $k$-dimensional image manifold.
\end{assumption}

At a regular point, write $J_z:=D\Phi_z$ and define the ambient vertical and horizontal spaces
\begin{equation}
\label{eq:HV_split}
\mathcal V_z:=\Ker(J_z),
\qquad
\mathcal H_z:=\mathcal V_z^\perp .
\end{equation}
The fiber $\mathcal F_x=\Phi^{-1}(x)$ consists of seeds with the same semantic outcome, and $\mathcal V_z$ is its tangent space at regular points. Directions in $\mathcal V_z$ are semantic-invariant to first order, while directions in $\mathcal H_z$ are the semantic-sensitive transverse directions depicted in Figure~\ref{Global}.

\subsection{Degenerate pullback geometry}
\label{sec:ambient_pullback}

The semantic map induces the pullback semi-metric
\begin{equation}
\label{eq:pullback_metric}
g_{\rm lat}(z)(u,v):=\langle J_zu,J_zv\rangle_{g_{\mathcal M}}.
\end{equation}
In local coordinates, with semantic metric matrix $G_{\mathcal M}(\Phi(z))$,
\begin{equation}
\label{eq:pullback_matrix}
G_{\rm lat}(z)=J_z^\top G_{\mathcal M}(\Phi(z))J_z.
\end{equation}
\begin{theorem}[Degenerate geometry]
\label{thm:degenerate_geometry_main}
For $z\in\mathcal Z_{\rm reg}$, $g_{\rm lat}$ vanishes exactly on $\mathcal V_z$ and is positive definite on $\mathcal H_z$. Equivalently, $G_{\rm lat}(z)$ has rank $k$ and nullspace $\mathcal V_z$.
\end{theorem}
\noindent\textbf{Proof.} Appendix~\ref{app:proofs:degenerate_geometry}. \qed

Let $\sigma_1(z)\ge\cdots\ge\sigma_k(z)>0$ be the nonzero singular values of $J_z$ and define
\begin{equation}
\label{eq:kappa_eff}
\kappa_{\rm eff}(z):=\frac{\sigma_1(z)^2}{\sigma_k(z)^2}.
\end{equation}
The local semantic effect of a small perturbation is controlled by its horizontal component.
\begin{lemma}[Local semantic sensitivity]
\label{lem:local_sensitivity_main}
If $\Phi$ is $C^2$ near a regular point $z$ and $J_z$ is locally Lipschitz, then for sufficiently small $\Delta z$,
\begin{align}
d_{\mathcal M}(\Phi(z+\Delta z),\Phi(z))
&\le c_2\sigma_1(z)\|\Proj_{\mathcal H_z}\Delta z\|+C\|\Delta z\|^2, \label{eq:local_sens_upper_refined}\\
d_{\mathcal M}(\Phi(z+\Delta z),\Phi(z))
&\ge c_1\sigma_k(z)\|\Proj_{\mathcal H_z}\Delta z\|-C\|\Delta z\|^2. \label{eq:local_sens_lower_refined}
\end{align}
If $\|\Proj_{\mathcal H_z}\Delta z\|\ge\beta\|\Delta z\|$, then, for small enough $\|\Delta z\|$,
\begin{equation}
\label{eq:local_sens_cone}
d_{\mathcal M}(\Phi(z+\Delta z),\Phi(z))
\ge \frac{c_1}{2}\sigma_k(z)\|\Proj_{\mathcal H_z}\Delta z\|.
\end{equation}
\end{lemma}
\noindent\textbf{Proof.} Appendix~\ref{app:proofs:local_sensitivity}. \qed

\subsection{Coarea view and interpretation}
\label{sec:global_overview}

The horizontal--vertical split explains local anisotropy; the coarea formula explains why seed mass is unevenly distributed across semantic fibers. Define
\begin{equation}
\label{eq:jk}
J_k(\Phi,z):=\prod_{i=1}^k\sigma_i(z)=\sqrt{\det(J_zJ_z^\top)}.
\end{equation}
\begin{theorem}[Coarea density transport]
\label{thm:coarea_main}
If $\Phi$ is locally Lipschitz on $\mathcal Z_{\rm reg}$ and $Z\sim p_{\mathcal Z}$ has a density, then for regular values $x\in\mathcal M$,
\begin{equation}
\label{eq:coarea_main}
p_{\mathcal M}(x)=\int_{\mathcal F_x}\frac{p_{\mathcal Z}(z)}{J_k(\Phi,z)}\,d\mathcal H^{d-k}(z).
\end{equation}
\end{theorem}
\noindent\textbf{Proof.} Appendix~\ref{app:proofs:coarea}. \qed
\begin{corollary}[Local compression energy]
\label{cor:energy_surrogate}
For the Gaussian prior, the local coarea weight motivates
\begin{equation}
\label{eq:eff_energy}
E_{\rm eff}(z):=\tfrac12\|z\|^2+\tfrac12\log\pdet(J_zJ_z^\top),
\end{equation}
up to an additive constant.
\end{corollary}
\noindent\textbf{Proof.} Appendix~\ref{app:proofs:energy_surrogate}. \qed

\paragraph{Mechanism.}
Figure~\ref{Global} summarizes the mechanism. Although the Gaussian prior samples seeds at comparable norm, the semantic map collapses high-dimensional fibers to the same or nearby meanings. Fiber-tangent motion mainly changes nuisance details, whereas transverse motion can enter different semantic basins. Seed lottery therefore reflects variation in local fiber geometry and horizontal conditioning: most directions are nearly semantic-silent, while a few corridors have strong semantic response. Section~\ref{sec:method} turns this view into a fixed-radius intervention that preserves the Gaussian norm while seeking prompt-sensitive transverse motion.
% =========================================

\section{Algorithm: Prompt-Residual Seed Shaping}
\label{sec:method}

\subsection{From ambient degeneracy to a radius-shell prompt update}
\label{sec:prompt_shell_objective}

We implement the geometric picture on the Gaussian radius shell.  For $R=\|z_T\|$, keeping $z_T$ on $\mathbb S_R^{d-1}$ preserves the sampled Gaussian radius and isolates directional effects.  Define
\begin{align}
\hat z&:=z/\|z\|,
& Q_z&:=I-\hat z\hat z^\top,
\label{eq:shell_projector}\\
A_z&:=J_zQ_z,
& J_z&:=D\Phi_z .
\label{eq:shell_jacobian}
\end{align}
The ambient fiber split induces the shell split
\begin{equation}
\label{eq:shell_HV_split}
\mathcal V_z^{\rm sh}:=\{u\in T_z\mathbb S_R^{d-1}:J_zu=0\},
\qquad
\mathcal H_z^{\rm sh}:=(\mathcal V_z^{\rm sh})^\perp\cap T_z\mathbb S_R^{d-1}=\Img(Q_zJ_z^\top).
\end{equation}
Equivalently, the shell pullback metric is
\begin{equation}
\label{eq:shell_pullback_metric}
G_{\rm sh}(z)=Q_zJ_z^\top G_{\mathcal M}(\Phi(z))J_zQ_z,
\end{equation}
whose null directions are shell-vertical and whose active directions lie in $\mathcal H_z^{\rm sh}$.  For a local prompt-semantic potential
\begin{equation}
\label{eq:prompt_objective}
F_y(z):=\ell_y(\Phi(z)),
\end{equation}
the shell gradient is horizontal:
\begin{lemma}[Shell prompt-gradient]
\label{lem:shell_prompt_gradient}
In local coordinates, absorbing the semantic metric into $\nabla_\Phi\ell_y$,
\begin{equation}
\label{eq:shell_prompt_gradient}
\nabla_{\mathbb S_R}F_y(z)=Q_zJ_z^\top\nabla_\Phi\ell_y(\Phi(z))\in\mathcal H_z^{\rm sh}.
\end{equation}
\end{lemma}
\begin{proof}
For $u\in T_z\mathbb S_R^{d-1}$, $DF_y(z)[u]=\langle J_z^\top\nabla_\Phi\ell_y,u\rangle$ and $u=Q_zu$, so the tangent representative is $Q_zJ_z^\top\nabla_\Phi\ell_y\in\Img(Q_zJ_z^\top)$.
\end{proof}
Since $J_z$ and $\ell_y$ are unavailable at inference time, we approximate this shell-horizontal prompt direction with a model residual.

\subsection{Setup and cold-start proxy}
\label{sec:method:setup}
\label{sec:method:single_probe}

Given prompt $y$, standard sampling draws $z_T\sim\mathcal N(0,I_d)$.  We use a single high-noise cold-start probe at $t^\star$:
\begin{equation}
\label{eq:cfg_residual}
g_\theta(z_t,t,y):=\epsilon_\theta(z_t,t,y)-\epsilon_\theta(z_t,t,\emptyset),
\end{equation}
with $\epsilon_\theta$ replaced by velocity for flow backbones, and
\begin{equation}
\label{eq:single_probe_proxy}
z_{t^\star}^{\rm cs}:=\rho(t^\star)z_T,
\qquad
v_{\rm proxy}:=g_\theta(z_{t^\star}^{\rm cs},t^\star,y),
\end{equation}
where $\rho(t)$ is the native scheduler scaling.  Algorithm~\ref{alg:sdno} then injects only the tangential component and retracts back to $\mathbb S_R^{d-1}$.

\paragraph{Residual as a noisy shell-gradient proxy.}
At the queried timestep, let $\Phi$ denote the downstream semantic map from the latent state to the final semantic outcome.  For score- or $\epsilon$-parameterized diffusion,
\begin{equation}
\label{eq:score_factorized_bridge}
g_\theta(z_t,t,y)=-\sigma_t\nabla_{z_t}\log p_\theta(y\mid z_t),
\end{equation}
up to score approximation, scalar factors, and sign convention.  For flow backbones we absorb parameterization mismatch into a residual term.  If locally
$\log p_\theta(y\mid z)=\ell_y(\Phi(z))+r_y(z)$, then
\begin{equation}
\label{eq:factorization_to_proxy}
g_\theta=c_tJ_z^\top\nabla_\Phi\ell_y+(c_t\nabla_z r_y+\xi_t)=:J_z^\top a_t(y)+e_t,
\qquad J_z:=D\Phi_z .
\end{equation}
Projecting to the shell gives
\begin{equation}
\label{eq:shell_proxy_decomposition}
Q_zg_\theta(z,t,y)=c_t\nabla_{\mathbb S_R}F_y(z)+Q_ze_t,
\end{equation}
so the implemented direction is a noisy shell-gradient proxy.  Appendix~\ref{app:proofs:score_factorized_bridge} gives the derivation.

\begin{lemma}[Shell cone condition for the prompt residual]
\label{lem:score_horizontal_main}
Let $q_z:=Q_zg_\theta(z,t,y)$.  If Eq.~\eqref{eq:factorization_to_proxy} holds, then
\begin{equation}
\label{eq:shell_vertical_error_identity}
\Proj_{\mathcal V_z^{\rm sh}}q_z
=
\Proj_{\mathcal V_z^{\rm sh}}(Q_ze_t),
\qquad
\|\Proj_{\mathcal V_z^{\rm sh}}q_z\|
\le
\|Q_ze_t\|.
\end{equation}
If, moreover,
\begin{equation}
\label{eq:shell_cone_condition}
\|\Proj_{\mathcal V_z^{\rm sh}}q_z\|
\le
\rho\,
\|\Proj_{\mathcal H_z^{\rm sh}}q_z\|,
\end{equation}
then for $u=q_z/\|q_z\|$,
\begin{equation}
\label{eq:shell_angle_bound}
\|\Proj_{\mathcal V_z^{\rm sh}}u\|
\le
\frac{\rho}{\sqrt{1+\rho^2}},
\qquad
\angle(u,\mathcal H_z^{\rm sh})
\le
\arctan\rho .
\end{equation}
\end{lemma}
\noindent\textbf{Proof.} Appendix~\ref{app:proofs:score_horizontal}. \qed

\begin{remark}[Empirical form of the cone condition]
\label{rem:empirical_cone_condition}
The exact spaces depend on the unobserved $J_z=D\Phi_z$.  In Section~\ref{sec:exp:proxy_diagnostics}, we estimate a finite-dimensional shell-horizontal section $\widehat{\mathcal H}_{z,k}^{\rm sh}$ by radius-preserving finite differences of $\pi\circ\Psi$ around the initial seed, and measure
\begin{equation}
\label{eq:rho_hat_main}
\widehat\rho_k(z,y)
:=
\frac{
\|\Proj_{\widehat{\mathcal V}_{z,k}^{\rm sh}} q_z\|
}{
\|\Proj_{\widehat{\mathcal H}_{z,k}^{\rm sh}} q_z\|+\varepsilon
},
\qquad
\widehat{\mathcal V}_{z,k}^{\rm sh}:=
(\widehat{\mathcal H}_{z,k}^{\rm sh})^\perp\cap T_z\mathbb S_R^{d-1}.
\end{equation}
Small $\widehat\rho_k$ is an empirical test of Eq.~\eqref{eq:shell_cone_condition}.
\end{remark}

\subsection{Local bridge and finite-step update}
\label{sec:method:bridge}

The proxy is evaluated at $z_{t^\star}^{\rm cs}=\rho(t^\star)z_T$.  Under the Lipschitz and boundedness assumptions in Appendix~\ref{app:proofs:cold_start_deviation}, with $h=T-t^\star$,
\begin{equation}
\label{eq:cold_start_state_bound}
\|z_{t^\star}^{\rm traj}-z_{t^\star}^{\rm cs}\|
\le
(B_v+L_\rho\|z_T\|)\frac{e^{L_vh}-1}{L_v}
=\mathcal O(T-t^\star),
\end{equation}
where the fraction is interpreted as $h$ if $L_v=0$.
\begin{proposition}[Cold-start state deviation]
\label{prop:cold_start_deviation}
Under the assumptions stated above, Eq.~\eqref{eq:cold_start_state_bound} holds.
\end{proposition}
\noindent\textbf{Proof.} Appendix~\ref{app:proofs:cold_start_deviation}. \qed

\begin{proposition}[Cold-start shell-proxy transfer]
\label{prop:cold_start_shell_transfer}
Let $\widetilde G_{t^\star}(z):=Q_zg_\theta(\rho(t^\star)z,t^\star,y)$.  If $\widetilde G_{t^\star}$ and $\nabla_{\mathbb S_R}F_y$ are locally Lipschitz with constants $L_G^{\rm sh}$ and $L_F^{\rm sh}$, and if $\widetilde G_{t^\star}(z^{\rm ref})=c_{t^\star}\nabla_{\mathbb S_R}F_y(z^{\rm ref})+r_{\rm ref}$, then
\begin{equation}
\label{eq:cold_start_shell_transfer}
\widetilde G_{t^\star}(z_T)=c_{t^\star}\nabla_{\mathbb S_R}F_y(z_T)+r_T,
\quad
\|r_T\|\le \|r_{\rm ref}\|+(L_G^{\rm sh}+c_{t^\star}L_F^{\rm sh})\|z_T-z^{\rm ref}\|.
\end{equation}
When Eq.~\eqref{eq:cold_start_state_bound} controls the reference discrepancy, the extra error is $\mathcal O(T-t^\star)$.
\end{proposition}
\noindent\textbf{Proof.} Appendix~\ref{app:proofs:cold_start_shell_transfer}. \qed

Lemma~\ref{lem:score_horizontal_main} controls vertical leakage; the following stronger bridge interprets the residual as a noisy ascent direction for $F_y$.  Our diagnostics directly test the former and provide operational evidence for the latter.
\begin{theorem}[Local algorithmic bridge]
\label{thm:algorithmic_bridge}
Let $R=\|z_T\|$, $F_y(z)=\ell_y(\Phi(z))$, and
\begin{equation}
\label{eq:algorithmic_direction}
\bar g_T:=Q_{z_T}g_\theta(z_{t^\star}^{\rm cs},t^\star,y),
\qquad
u_0:=\frac{\bar g_T}{\|\bar g_T\|}\quad (\bar g_T\neq0).
\end{equation}
Assume $\nabla_{\mathbb S_R}F_y(z_T)\neq0$ and, for $c_{t^\star}>0$ and $0\le\rho_c<1$,
\begin{equation}
\label{eq:algorithmic_cone_assumption}
\bar g_T=c_{t^\star}\nabla_{\mathbb S_R}F_y(z_T)+r_T,
\qquad
\|r_T\|\le \rho_c c_{t^\star}\|\nabla_{\mathbb S_R}F_y(z_T)\|.
\end{equation}
For the $\varepsilon$-stabilized implementation, also assume $\|\bar g_T\|\ge m_g>0$.  For
\begin{equation}
\label{eq:bridge_retraction}
z_T^\star=R\frac{z_T+\delta u_0}{\|z_T+\delta u_0\|},
\end{equation}
if $F_y$ is $C^2$ near the shell segment and $\delta$ is small, then
\begin{equation}
\label{eq:algorithmic_bridge_gain}
F_y(z_T^\star)-F_y(z_T)
\ge
\delta\frac{1-\rho_c}{1+\rho_c}\|\nabla_{\mathbb S_R}F_y(z_T)\|-C_F\delta^2-O(\delta\varepsilon/m_g).
\end{equation}
With exact normalization the last term is absent; Proposition~\ref{prop:cold_start_shell_transfer} adds an $\mathcal O(\delta(T-t^\star))$ cold-start term.
\end{theorem}
\noindent\textbf{Proof.} Appendix~\ref{app:proofs:algorithmic_bridge}. \qed

The theorem is explanatory rather than a runtime certificate because $J_z$, $e_t$, and $F_y$ are unobserved.  We therefore test the proxy using radius-preserving probes
\begin{equation}
\label{eq:radius_probe_main}
z_T(\theta;u)=\|z_T\|\left(\cos\theta\frac{z_T}{\|z_T\|}+\sin\theta\,u\right),
\qquad u\perp z_T,
\end{equation}
and semantic displacement
\begin{equation}
\label{eq:semantic_displacement_main}
D_\pi(z_T,u;\theta)=d_\pi\bigl(\pi(\Psi(z_T(\theta;u))),\pi(\Psi(z_T))\bigr).
\end{equation}

\subsection{Tangential injection and spherical retraction}
\label{sec:method:inject}

Let $\hat z=z_T/\|z_T\|$ and $\hat v=v_{\rm proxy}/(\|v_{\rm proxy}\|+\varepsilon)$.  The implementation is
\begin{equation}
\label{eq:orthogonalize}
\tilde v_\perp=\hat v-(\hat v^\top\hat z)\hat z,
\qquad
\hat v_\perp=\frac{\tilde v_\perp}{\|\tilde v_\perp\|+\varepsilon},
\qquad
z_T^\star=\|z_T\|\frac{z_T+\delta\hat v_\perp}{\|z_T+\delta\hat v_\perp\|+\varepsilon}.
\end{equation}
\begin{proposition}[Retraction as projection onto the radius shell]
\label{prop:retraction_opt}
Ignoring the numerical stabilizer, $\|z_T\|(z_T+\delta\hat v_\perp)/\|z_T+\delta\hat v_\perp\|$ is the Euclidean projection of $z_T+\delta\hat v_\perp$ onto $\{z:\|z\|=\|z_T\|\}$; the implemented $+\varepsilon$ denominator adds only an $O(\varepsilon)$ radial perturbation.
\end{proposition}
\noindent\textbf{Proof.} Appendix~\ref{app:proofs:retraction_opt}. \qed

\begin{lemma}[First-order shell retraction]
\label{lem:small_delta}
Let $R=\|z_T\|$ and $u\perp z_T$, $\|u\|=1$.  For $z_T^+=R(z_T+\delta u)/\|z_T+\delta u\|$,
\begin{equation}
\label{eq:first_order_retraction}
z_T^+=z_T+\delta u-\frac{\delta^2}{2R^2}z_T+O\!\left(\frac{\delta^3}{R^2}\right),
\end{equation}
and, for any $C^2$ prompt-semantic objective,
\begin{equation}
\label{eq:first_order_objective_expansion}
F_y(z_T^+)-F_y(z_T)=\delta\langle\nabla_{\mathbb S_R}F_y(z_T),u\rangle+O(\delta^2).
\end{equation}
\end{lemma}
\noindent\textbf{Proof.} Appendix~\ref{app:proofs:small_delta}. \qed

\begin{algorithm}[htbp]
\small
\caption{Prior-Compatible Prompt-Residual Seed Shaping}
\label{alg:sdno}
\begin{algorithmic}[1]
\REQUIRE Prompt $\mathbf y$, model $M$, probe timestep $t^\star$, rescale $\rho(\cdot)$, strength $\delta$
\ENSURE Output $x_0$
\STATE Sample $\mathbf z_T\sim\mathcal N(0,I)$.
\STATE Set $\mathbf z_{t^\star}^{\rm cs}\gets \rho(t^\star)\mathbf z_T$ and compute $\mathbf r\gets M(\mathbf z_{t^\star}^{\rm cs},t^\star,\mathbf y)-M(\mathbf z_{t^\star}^{\rm cs},t^\star,\emptyset)$.
\STATE Normalize, remove the radial component, and re-normalize:
$\hat{\mathbf v}\gets \mathbf r/(\|\mathbf r\|+\varepsilon)$,
$\mathbf v_\perp\gets \hat{\mathbf v}-\frac{\langle\hat{\mathbf v},\mathbf z_T\rangle}{\|\mathbf z_T\|^2+\varepsilon}\mathbf z_T$,
$\hat{\mathbf v}_\perp\gets \mathbf v_\perp/(\|\mathbf v_\perp\|+\varepsilon)$.
\STATE Retract to the original shell:
$\widetilde{\mathbf z}_T\gets \mathbf z_T+\delta\hat{\mathbf v}_\perp$,
$\mathbf z_T^\star\gets\|\mathbf z_T\|\widetilde{\mathbf z}_T/(\|\widetilde{\mathbf z}_T\|+\varepsilon)$.
\STATE $x_0\gets \mathrm{Sampler}(M,\mathbf z_T^\star,\mathbf y)$; \textbf{return} $x_0$.
\end{algorithmic}
\end{algorithm}

\section{Experiment and Analysis}
\label{sec:exp}

\subsection{Prompt-residual proxy diagnostics}
\label{sec:exp:proxy_diagnostics}

We test the empirical cone condition in Remark~\ref{rem:empirical_cone_condition}.
For each prompt--seed pair, let $q_z:=Q_zg_\theta(z,t^\star,y)$.
We form a radius-preserving local section from the normalized residual direction and $m$ random tangent controls.
For tangent $u$, we estimate
$
s(u)=
\frac{\pi(\Psi(\operatorname{Ret}_z(\epsilon u)))-\pi(\Psi(z))}{\epsilon},
\operatorname{Ret}_z(\xi)=R\frac{z+\xi}{\|z+\xi\|}.
$
The top-$k$ right singular directions define $\widehat{\mathcal H}_{z,k}^{\rm sh}$.
We report $\widehat\rho_k$ from Eq.~\eqref{eq:rho_hat_main}, horizontal energy
$
E_{H,k}
=
\frac{\|\Proj_{\widehat{\mathcal H}_{z,k}^{\rm sh}}q_z\|^2}{\|q_z\|^2+\varepsilon},
$
and semantic gain over random tangent controls.

Table~\ref{tab:sectional_cone_main} reports the main $k=16$ setting.
For FLUX, SD3.5M, and Z-Image, $\widehat\rho_{16}<0.2$ and horizontal energy exceeds $96\%$; semantic gain is also above random controls with percentile $1.0$.
Thus the residual is not only tangent, but concentrated in the estimated semantic-horizontal shell, empirically instantiating Lemma~\ref{lem:score_horizontal_main}.

\begin{table}[htbp]
\centering
\caption{\textbf{Finite-difference sectional cone diagnostics for the prompt residual.}
All models use official inference settings. Detailed $k=10$ results and protocols are in Appendix~\ref{app:sectional_cone_diagnostics}.}
\label{tab:sectional_cone_main}
\small
\setlength{\tabcolsep}{2pt}
\resizebox{\linewidth}{!}{%
\begin{tabular}{lcccc lcccc}
\toprule
Model &
$\widehat\rho_{16}\downarrow$ &
Horizontal energy $\uparrow$ &
Semantic gain $\uparrow$ &
Percentile $\uparrow$ &
Model &
$\widehat\rho_{16}\downarrow$ &
Horizontal energy $\uparrow$ &
Semantic gain $\uparrow$ &
Percentile $\uparrow$ \\
\midrule
FLUX~\cite{flux2024}
& 0.110 & 98.8\% & 2.124 & 1.000
&
Z-Image~\cite{team2025zimage}
& 0.159 & 97.5\% & 2.420 & 1.000 \\

SDXL~\cite{SDXL}
& 0.189 & 97.2\% & 1.976 & 1.000
&
WAN~\cite{DBLP:journals/corr/abs-2503-20314}
& 0.114 & 98.2\% & 2.560 & 1.000 \\

SD3.5M~\cite{esser2024scalingrectifiedflowtransformers}
& 0.198 & 96.2\% & 2.467 & 1.000
&
TRELLIS~\cite{DBLP:conf/cvpr/XiangLXDWZC0Y25}
& 0.158 & 97.4\% & 2.133 & 1.000 \\
\bottomrule
\end{tabular}}
\end{table}

Since $\Phi$ and $J_z$ are unavailable at runtime, these diagnostics are not certificates; they validate the proxy's geometric premise.
Additional percentiles, residual concentration, and high-noise consistency are in Appendix~\ref{app:proxy_diagnostics}.

\subsection{Latent degeneracy visualization}
\label{sec:exp:degeneracy_viz_main}

Our theory predicts anisotropic semantic sensitivity on the Gaussian radius shell.
For $R=\|z_T\|$ and $\hat z=z_T/R$, we probe tangent directions by
$
\hat u=
\frac{u-\langle u,\hat z\rangle\hat z}
{\|u-\langle u,\hat z\rangle\hat z\|}$ and
$z_T(\theta)=R(\cos\theta\,\hat z+\sin\theta\,\hat u),
$
, so semantic changes are directional rather than radial.
The diagnostics show heavy-tailed CLIP displacement and stronger movement along the prompt-residual proxy than random tangent controls, supporting the horizontal--vertical view of seed lottery.
Full protocols and plots are in Appendix~\ref{app:degeneracy_viz}, with detailed proxy diagnostics in Appendix~\ref{app:proxy_diagnostics}.

% ---------------------------------------------------------
\subsection{Main results across image, video, and 3D generation}
\label{sec:exp:quant}

Tables~\ref{tab:full_sdxl_main_exp}, \ref{tab:full_video_generation}, and \ref{tab:full_3d_generation} summarize the main image, video, and 3D results.
The method improves SDXL and FLUX metrics over standard sampling and available initial-condition baselines, and transfers to WAN video and TRELLIS 3D generation without retraining. Human evaluation results are reported in Appendix~\ref{US}.

\begin{table}[t]
\centering
\caption{\textbf{Main generation results across image, video, and 3D generation.} See Appendix~\ref{tab:afull_sdxl_main_exp} for PickScore and VQAScore $\Delta [95\% CI]$  and Win Rate.}
\label{tab:all_generation_results}

\begin{minipage}[t]{0.5\linewidth}
\vspace{0pt}
\centering
\subcaption{\textbf{Image generation.}}
\label{tab:full_sdxl_main_exp}

\tiny
\renewcommand{\arraystretch}{0.52}
\setlength{\tabcolsep}{5pt}
\resizebox{\linewidth}{!}{%
\begin{tabular}{@{}lllccc@{}}
\toprule
Model & Dataset & Method & Pick $\uparrow$ & ImgR $\uparrow$ & CLIP $\uparrow$ \\
\midrule
SDXL & Pick-a-Pic & Standard
& 17.049 & -1.971 & 16.223 \\
SDXL & Pick-a-Pic & Initno$^\dagger$ \cite{DBLP:conf/cvpr/GuoLCLY024} (Training-free)
& 17.050 & -1.969 & 16.237 \\
SDXL & Pick-a-Pic & NPNet \cite{zhou2025golden} (Training-based)
& 17.051 & -1.968 & 16.250 \\
SDXL & Pick-a-Pic & Ours (Training-free)
& \textbf{17.371} & \textbf{-1.852} & \textbf{16.642} \\
\midrule
SDXL & DrawBench & Standard
& 17.416 & -2.084 & 16.618 \\
SDXL & DrawBench & Initno$^\dagger$ \cite{DBLP:conf/cvpr/GuoLCLY024} (Training-free)
& 17.419 & -2.085 & 16.632 \\
SDXL & DrawBench & NPNet \cite{zhou2025golden} (Training-based)
& 17.420 & -2.084 & 16.640 \\
SDXL & DrawBench & Ours (Training-free)
& \textbf{17.619} & \textbf{-1.984} & \textbf{16.957} \\
\midrule
SDXL & HPD & Standard
& 16.748 & -1.955 & 14.986 \\
SDXL & HPD & Initno$^\dagger$ \cite{DBLP:conf/cvpr/GuoLCLY024} (Training-free)
& 16.765 & -1.952 & 15.010 \\
SDXL & HPD & NPNet \cite{zhou2025golden} (Training-based)
& 16.780 & -1.948 & 15.040 \\
SDXL & HPD & Ours (Training-free)
& \textbf{17.041} & \textbf{-1.875} & \textbf{15.851} \\
\midrule
FLUX & Pick-a-Pic & Standard
& 17.134 & -1.942 & 16.295 \\
FLUX & Pick-a-Pic & NPNet \cite{zhou2025golden} (Training-based)
& 17.140 & -1.936 & 16.305 \\
FLUX & Pick-a-Pic & Ours (Training-free)
& \textbf{17.732} & \textbf{-1.905} & \textbf{16.914} \\
\midrule
FLUX & DrawBench & Standard
& 17.275 & -2.051 & 16.642 \\
FLUX & DrawBench & NPNet \cite{zhou2025golden} (Training-based)
& 17.282 & -2.044 & 16.655 \\
FLUX & DrawBench & Ours (Training-free)
& \textbf{17.626} & \textbf{-1.968} & \textbf{17.013} \\
\midrule
FLUX & HPD & Standard
& 16.730 & -1.951 & 15.021 \\
FLUX & HPD & NPNet \cite{zhou2025golden}
& 16.750 & -1.944 & 15.900 \\
FLUX & HPD & Ours (Training-free)
& \textbf{17.239} & \textbf{-1.874} & \textbf{16.731} \\
\bottomrule
\end{tabular}%
}
\end{minipage}
\hfill
\begin{minipage}[t]{0.45\linewidth}
\vspace{0pt}
\centering

\subcaption{\textbf{Video generation.}}
\label{tab:full_video_generation}
\vspace{0.5mm}

\tiny
\renewcommand{\arraystretch}{1.0}
\setlength{\tabcolsep}{7pt}
\resizebox{\linewidth}{!}{%
\begin{tabular}{@{}lccccc@{}}
\toprule
Method & AQ $\uparrow$ & AS $\uparrow$ & BC $\uparrow$ & DD $\uparrow$ & IQ $\uparrow$ \\
\midrule
Standard & 0.572 & 0.220 & 0.963 & 0.644 & 0.644 \\
Ours    & \textbf{0.651} & \textbf{0.341} & \textbf{0.985} & \textbf{0.701} & \textbf{0.664} \\
\midrule
Method & MS $\uparrow$ & OC $\uparrow$ & SC $\uparrow$ & TF $\uparrow$ & VQA $\uparrow$ \\
\midrule
Standard & 0.972 & 0.244 & 0.937 & 0.984 & 0.534 \\
Ours    & \textbf{0.997} & \textbf{0.303} & \textbf{0.994} & \textbf{0.997} & \textbf{0.600} \\
\bottomrule
\end{tabular}%
}

\vspace{2mm}

\subcaption{\textbf{3D generation.}}
\label{tab:full_3d_generation}
\vspace{2mm}

\tiny
\renewcommand{\arraystretch}{1.0}
\setlength{\tabcolsep}{4pt}
\resizebox{\linewidth}{!}{%
\begin{tabular}{@{}lccccc@{}}
\toprule
Method & FD$_I$ $\downarrow$ & KD$_I$ $\downarrow$ & FD$_D$ $\downarrow$ & KD$_D$ $\downarrow$ & VQA $\uparrow$ \\
\midrule
Standard & 29.544 & 0.011 & 340.530 & 0.802 & 0.876 \\
Ours     & \textbf{29.403} & \textbf{0.008} & \textbf{335.425} & \textbf{0.733} & \textbf{0.911} \\
\bottomrule
\end{tabular}%
}

\vspace{0.5mm}
{\tiny $I$: Inception; $D$: DINOv2.}
\end{minipage}
\end{table}

\vspace{-1mm}
\begin{table}[t]
\centering
\caption{\textbf{Robustness, diagnostic behavior, and operating cost on FLUX.}}
\label{tab:robust_boundary_combined}

\begin{minipage}[t]{0.52\linewidth}
\vspace{0pt}
\centering
\subcaption{\textbf{Seed robustness and probe diagnostic.}
HPSv3 is the reported score; Q20 is the 20th percentile over matched prompt--seed cases.
$p$ denotes prompts, $s/p$ denotes seeds per prompt, and ``diag.'' denotes diagnostic.}
\label{tab:robust_proxy}

\vspace{-1mm}
\scriptsize
\renewcommand{\arraystretch}{0.88}
\setlength{\tabcolsep}{8pt}
\resizebox{\linewidth}{!}{%
\begin{tabular}{@{}llcc@{}}
\toprule
Block & Setting & HPSv3 $\uparrow$ & Gain / note \\
\midrule
Seed robust.
& 48p, 32s/p & 10.668 & $+0.146$, Q20: $8.697{\to}8.784$ \\
& 48p, 8s/p  & 10.667 & $+0.201$, Q20: $8.613{\to}8.897$ \\
\midrule
Probe diag.
& Standard & 10.466 & -- \\
& $t^\star=900$ & 10.667 & $+0.201$ (default) \\
\bottomrule
\end{tabular}}
\end{minipage}
\hfill
\begin{minipage}[t]{0.45\linewidth}
\vspace{0pt}
\centering
\subcaption{\textbf{Operating boundary and overhead.}
Mod., high, and ext. denote moderate, high, and extreme guidance stress. Guidance rows report HPSv3 ranges; cost rows report runtime / peak memory.}
\label{tab:boundary_cost}

\vspace{-1mm}
\scriptsize
\renewcommand{\arraystretch}{0.8}
\setlength{\tabcolsep}{6pt}
\resizebox{\linewidth}{!}{%
\begin{tabular}{@{}llc@{}}
\toprule
Block & Setting & Metric / reading \\
\midrule
Guidance
& Mod. CFG & HPSv3 11.027--11.129 (stable gain) \\
& High CFG & HPSv3 6.00--6.31 (stress-test degradation) \\
& Ext. CFG & HPSv3 $-1.44$--$-1.92$ (unstable stress test) \\
\midrule
Cost
& Standard & 9.50--9.53s / 34.65 GB \\
& Ours & 9.62--9.66s / 34.65 GB \\
\bottomrule
\end{tabular}}
\end{minipage}
\end{table}

% ---------------------------------------------------------
\subsection{Seed robustness and proxy diagnostics}
\label{sec:exp:robust_proxy}

Matched prompt--seed HPSv3 results in Table~\ref{tab:robust_proxy} improve both mean and Q20, indicating fewer brittle low-quality seeds.
The $t^\star=900$ cold-start residual already outperforms standard sampling, so the method acts as a single-probe initialization proxy rather than cross-timestep optimization.

\subsection{Ablation: component choice, probe stage, and strength}
\label{sec:exp:ablate}
\begin{figure}[htbp]
    \centering
    \includegraphics[width=0.85\textwidth]{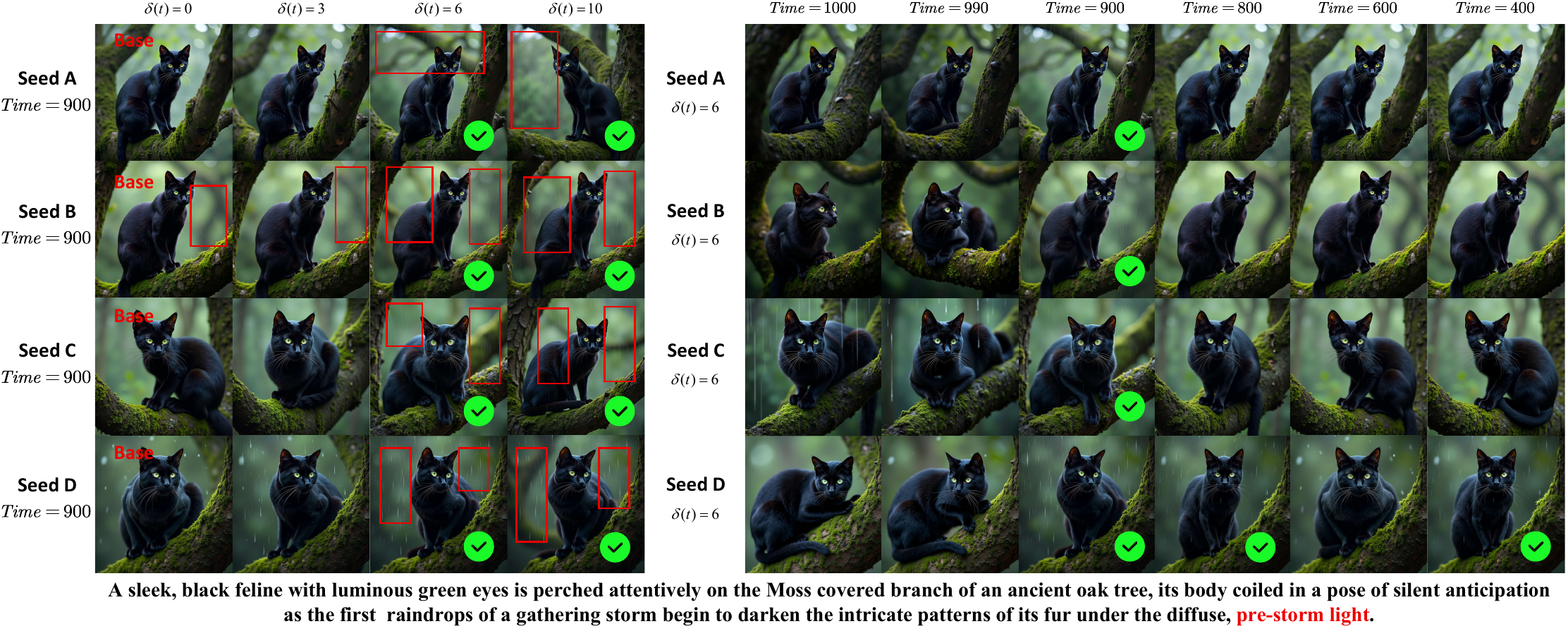}
    \caption{\textbf{Probe stage and injection strength.}}
    \label{time}
\end{figure}
\begin{figure*}[htbp]
    \centering
    \includegraphics[width=0.85\textwidth]{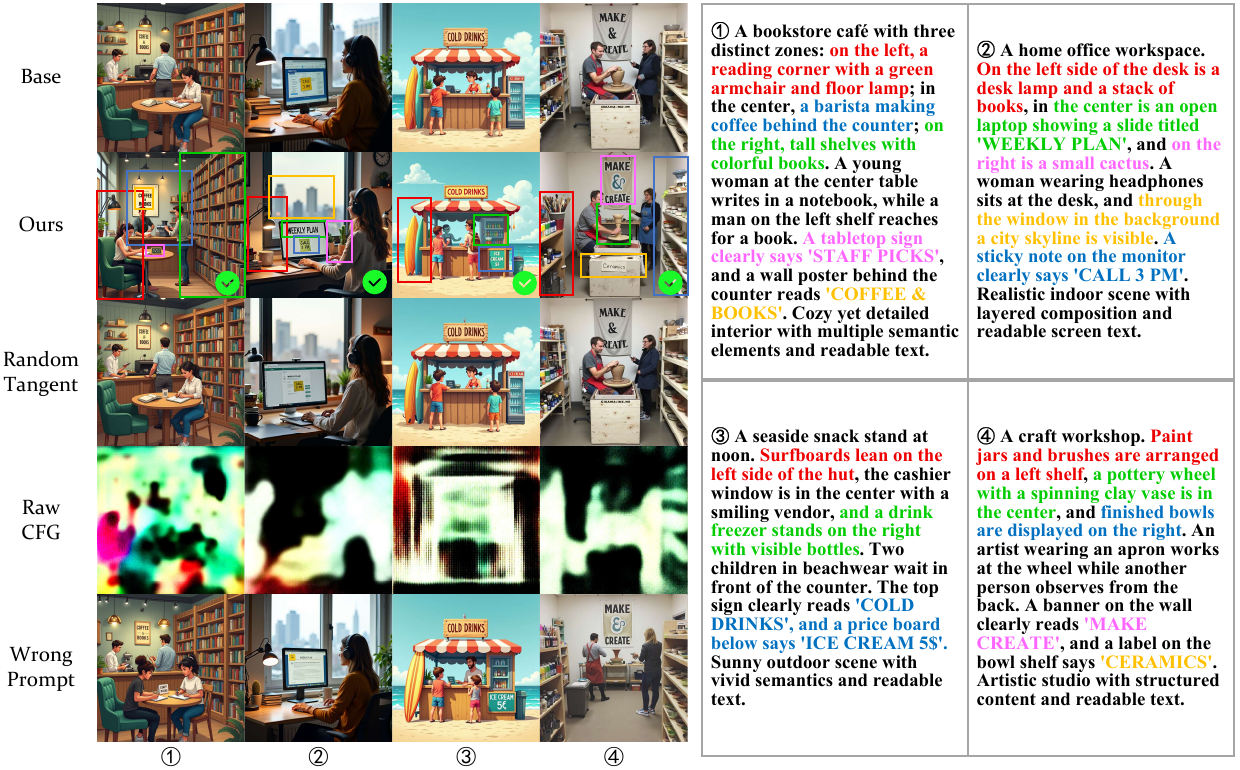}
    \caption{\textbf{Qualitative component ablation of seed shaping on FLUX.}}
    \label{fig:component_ablation_qual}
\end{figure*}

Figure~\ref{time} ablates $t^\star$ and $\delta$.
Moderate $\delta$ works best: $\delta=0$ has no effect, while large updates can over-perturb structure; with $\delta=6$, the high-noise probe around $t^\star=900$ is most stable.
This supports the method as a bounded initialization-side correction. Figure~\ref{fig:component_ablation_qual} compares Base, Random Tangent, Raw CFG, Wrong Prompt, and the full method on matched FLUX cases.
Random Tangent removes prompt information, Raw CFG removes tangential projection and shell retraction, and Wrong Prompt uses an unrelated residual.
Raw CFG often destabilizes structure, while Random Tangent and Wrong Prompt fail to reliably recover target layout, relations, or readable text.
The full method more consistently adds prompt-specified elements.

% ---------------------------------------------------------
\subsection{Operating boundary and cost}
\label{sec:exp:boundary_cost}
Table~\ref{tab:boundary_cost} summarizes the operating boundary and cost. The method remains stable under moderate CFG, while high/extreme CFG act as out-of-range stress tests and may overshoot. This supports viewing the method as a bounded local seed correction rather than a global steering mechanism. Its only extra cost is one conditional/unconditional cold-start probe, with modest runtime overhead and nearly unchanged peak memory. Diversity is largely preserved (Appendix~\ref{app:diversity}).

\section{Conclusion}
\label{sec:Con}

We presented a geometric view of seed sensitivity in diffusion and flow generation. Although the initialization prior is isotropic, the induced semantic map is not: many latent directions are nearly semantic-invariant, while meaningful variation concentrates in a low-dimensional sensitive subspace. This degenerate pullback geometry offers an explanatory account of the seed lottery. Motivated by this view, we proposed a training-free prompt-residual seed-shaping method. Rather than recovering the exact horizontal structure, the method uses a single high-noise prompt residual as a model-coupled proxy, updates the seed tangentially, and retracts it to the original Gaussian radius shell. The finite-difference sectional diagnostics further show that the shell-projected prompt residual has low vertical leakage relative to the empirical semantic-horizontal subspace. This empirically bridges the cone condition used in our local theory with the behavior of official-pipeline image generation backbones. The resulting intervention is prior-compatible, requires no retraining or cross-timestep aggregation, and improves alignment metrics across image, video, and 3D generation. Limitations see \ref{sec:limitations_impact}.

\clearpage

\bibliographystyle{IEEEtran}
\bibliography{example_paper}

\clearpage
\appendix

\section{Finite-Difference Sectional Cone Diagnostics}
\label{app:sectional_cone_diagnostics}

This appendix provides the detailed protocol behind Table~\ref{tab:sectional_cone_main}.
For each prompt--seed pair, we evaluate the official pipeline at the base seed and at radius-preserving perturbed seeds.
All perturbations stay on the Gaussian radius shell, so the measured differences are directional rather than radial.
We use $m=32$ random tangent controls.
The empirical horizontal space $\widehat{\mathcal H}_{z,k}^{\rm sh}$ is obtained from the top-$k$ right singular directions of the finite-difference semantic response matrix.

We report four quantities.
First, the empirical cone ratio
\begin{equation}
\widehat\rho_k
=
\frac{
\|\Proj_{\widehat{\mathcal V}_{z,k}^{\rm sh}}q_z\|
}{
\|\Proj_{\widehat{\mathcal H}_{z,k}^{\rm sh}}q_z\|+\varepsilon
}.
\end{equation}
Second, the horizontal energy fraction
\begin{equation}
E_{H,k}
=
\frac{
\|\Proj_{\widehat{\mathcal H}_{z,k}^{\rm sh}}q_z\|^2
}{
\|q_z\|^2+\varepsilon
}.
\end{equation}
Third, semantic gain compares the semantic displacement induced by the prompt residual against the mean displacement of random tangent controls.
Fourth, percentile measures the fraction of random tangent controls whose semantic displacement is below the prompt-residual displacement.

\begin{table}[htbp]
\centering
\caption{\textbf{Sectional cone diagnostics at $k=10$ and $k=16$.}
This table expands Table~\ref{tab:sectional_cone_main}; $k=5$ is omitted because the main theory concerns sufficiently wide empirical horizontal sections.}
\label{tab:sectional_cone_appendix}
\small
\setlength{\tabcolsep}{5pt}
\begin{tabular}{lccccc}
\toprule
Model & $k$ &
$\widehat\rho_k\downarrow$ &
Horizontal energy $\uparrow$ &
Semantic gain $\uparrow$ &
Percentile $\uparrow$ \\
\midrule
FLUX    & 10 & 0.176 & 97.0\% & 2.124 & 1.000 \\
FLUX    & 16 & 0.110 & 98.8\% & 2.124 & 1.000 \\
SDXL    & 10 & 0.251 & 96.1\% & 1.976 & 1.000 \\
SDXL    & 16 & 0.189 & 97.2\% & 1.976 & 1.000 \\
SD3.5M  & 10 & 0.379 & 87.4\% & 2.467 & 1.000 \\
SD3.5M  & 16 & 0.198 & 96.2\% & 2.467 & 1.000 \\
Z-Image & 10 & 0.288 & 92.3\% & 2.420 & 1.000 \\
Z-Image & 16 & 0.159 & 97.5\% & 2.420 & 1.000 \\
WAN & 10 & 0.126 & 96.7\% & 2.560 & 1.000 \\
WAN & 16 & 0.114 & 98.2\% & 2.560 & 1.000 \\
TRELLIS & 10 & 0.201 & 90.1\% & 2.133 & 1.000 \\
TRELLIS & 16 & 0.158 & 97.4\% & 2.133 & 1.000 \\
\bottomrule
\end{tabular}
\end{table}

The trend is consistent across the six backbones: increasing $k$ from $10$ to $16$ reduces the estimated vertical-to-horizontal ratio and increases the horizontal energy.
At $k=16$, all five models satisfy the strong empirical cone pattern reported in the main text.

\section{Sensitivity of the geometric diagnostics to the choice of semantic map $\pi$}
\label{app:pi_sensitivity}

This appendix studies the robustness of our \textbf{geometry-only} diagnostics to the choice of semantic map $\pi$ in $\Phi=\pi\circ\Psi$.
We focus on the diagnostic procedure itself---prompt stratification via local Jacobian-spectrum statistics---and evaluate whether its qualitative behavior remains consistent across a set of frozen, widely used semantic representations.

\paragraph{Notation.}
Throughout this appendix, $\Psi$ (a.k.a.\ $G$ in the main text) denotes the generator mapping base noise $z$ to an output image $x=\Psi(z)$.
We instantiate $\Psi$ with the Z-Image generator~\cite{team2025zimage} and study the composed map $\Phi=\pi\circ\Psi$.

\paragraph{Semantic maps $\pi$ considered.}
We evaluate six frozen choices of $\pi$ spanning (i) CLIP\footnote{\url{https://github.com/openai/CLIP}}, (ii) OpenCLIP/VLM-style embeddings\footnote{\url{https://github.com/mlfoundations/open_clip}}, and (iii) DINOv2 image embeddings fused with CLIP-text\footnote{\url{https://github.com/facebookresearch/dinov2}}.
For each family, we use two fusion rules: elementwise image--text product (\texttt{itprod}) and concatenation (\texttt{concat}).

CLIP: clip\_ViT-L-14\_openai\_itprod, clip\_ViT-L-14\_openai\_concat.

VLM: vlm\_ViT-H-14\_laion2b\_s32b\_b79k\_itprod, vlm\_ViT-H-14\_laion2b\_s32b\_b79k\_concat.

DINOv2 + CLIP-text: dinov2\_vitb14\_cliptext\_itprod, dinov2\_vitb14\_cliptext\_concat.

\paragraph{Jacobian estimator and spectrum statistics.}
For each prompt, we estimate a randomized local sketch of the Jacobian of $\Phi=\pi\circ\Psi$ at a fixed base noise using symmetric finite differences along random directions.
We probe \textbf{$r=6$} random directions, so the maximum observable rank, effective rank, and participation ratio are upper-bounded by $6$.
Accordingly, the measured spectrum should be interpreted as a finite local section of semantic sensitivity, not as a full-rank estimate of $D\Phi$.
We report four statistics from the probed singular values:
(i) smallest singular value $s_{\min}$,
(ii) effective rank,
(iii) participation ratio, and
(iv) condition number $\kappa$ reported as \texttt{cond}.

\paragraph{How to read the plots.}
The evaluator produces three plot types:
\begin{itemize}
    \item \textbf{Violin plots} (\texttt{fig\_spectrum\_stats\_violin\_*}): prompt-level mean distributions, with seeds aggregated per prompt.
    \item \textbf{Heatmaps} (\texttt{fig\_pi\_spearman\_heatmap\_*}): pairwise Spearman correlations of prompt-level means across different $\pi$.
    \item \textbf{Seed-stability boxplots} (\texttt{fig\_seed\_stability\_box\_*}): within-prompt seed variability, e.g., per-prompt standard deviation.
\end{itemize}
The conclusions below correspond to these plots.

\paragraph{Main observation.}
Across all six choices of $\pi$, the diagnostics show a consistent finite-section spectral pattern.
Although the numerical scale of some quantities, especially $s_{\min}$ and \texttt{cond}, naturally depends on the representation, the rank-type statistics are stable across semantic maps.
Effective rank remains close to the probe ceiling $r=6$, and participation ratio stays in a narrow range below that ceiling.
This indicates that the probed local semantic response is not dominated by an idiosyncratic single direction and that the diagnostic procedure remains well-behaved across different choices of $\pi$.
Thus, these diagnostics can be used for prompt-level stratification without committing to one particular semantic encoder.

\paragraph{Scale and structural consistency.}
Table~\ref{tab:pi_geom_summary} reports prompt-averaged metrics with 95\% bootstrap CIs over $211$ HPD prompts.
The results show two complementary trends:
\begin{itemize}
    \item \textbf{Representation-dependent scale.}
    The prompt-averaged $s_{\min}$ varies across representations, roughly from $\approx 0.040$ to $\approx 0.066$, reflecting the different sensitivity units induced by each $\pi$.

    \item \textbf{Stable finite-section spectral structure.}
    Across $\pi$, effective rank remains around $\approx 5.37$--$5.44$ out of the maximum observable value $6$, while participation ratio remains around $\approx 4.82$--$4.98$.
    This shows that the local six-direction Jacobian sketch has a stable spectral shape across semantic maps.
    The result should not be read as an estimate of the full semantic rank of $D\Phi$; rather, it indicates that the same finite-section diagnostic behaves consistently under different semantic representations.
\end{itemize}
Thus, while different semantic maps induce different measurement scales, the relative spectral structure used by our diagnostics remains consistent.

\paragraph{Seed stability.}
Table~\ref{tab:pi_geom_summary} also reports the mean within-prompt standard deviation across seeds for each metric and $\pi$.
The rank-type metrics show similar seed-stability behavior across semantic maps, while scale-dependent quantities such as $s_{\min}$ and \texttt{cond} reflect the expected representation-specific units.
Overall, the diagnostics remain stable enough to support prompt-level comparisons.

\paragraph{Geometry--$\Delta$ correlations.}
We additionally compute prompt-level Spearman correlations between geometry metrics and $\Delta$ under a fixed evaluator.
These results provide a complementary view of how geometric diagnostics interact with downstream scoring.
Since $\Delta$ depends on the evaluator, its relationship with a particular semantic representation can vary across $\pi$.
This is consistent with our interpretation: the geometry-only diagnostics capture local latent sensitivity, while downstream score prediction also depends on evaluator alignment.

\paragraph{Scope of interpretation.}
The diagnostics are randomized local sketches, not full measurements of the global Jacobian spectrum.
They are most informative when the semantic map provides a meaningful representation of the prompt-relevant attributes.
Prompts involving fine-grained counting, exact attribute conjunctions, or text-heavy content can exhibit higher seed sensitivity, which is reflected in larger within-prompt variability.
Similarly, agreement across $\pi$ is strongest for well-aligned, high-capacity semantic maps.
These observations clarify the intended scope of the diagnostics and support their use as practical tools for analyzing latent-space anisotropy, while showing that the qualitative behavior is not tied to a single semantic representation in this Z-Image study.

\graphicspath{{diagnostics_eval_geom/}}

\begin{figure}[t]
\centering
\begin{subfigure}[t]{0.49\textwidth}
  \centering
  \includegraphics[width=\linewidth]{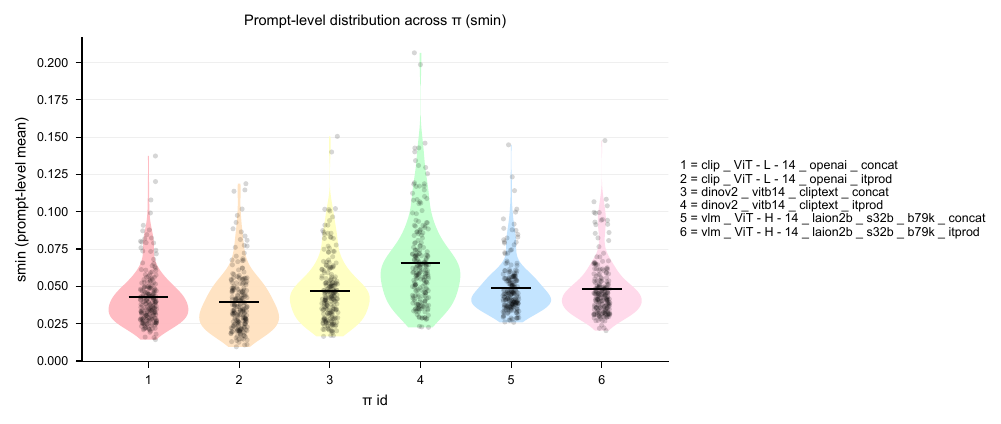}
  \caption{$s_{\min}$ (prompt-level means)}
\end{subfigure}\hfill
\begin{subfigure}[t]{0.49\textwidth}
  \centering
  \includegraphics[width=\linewidth]{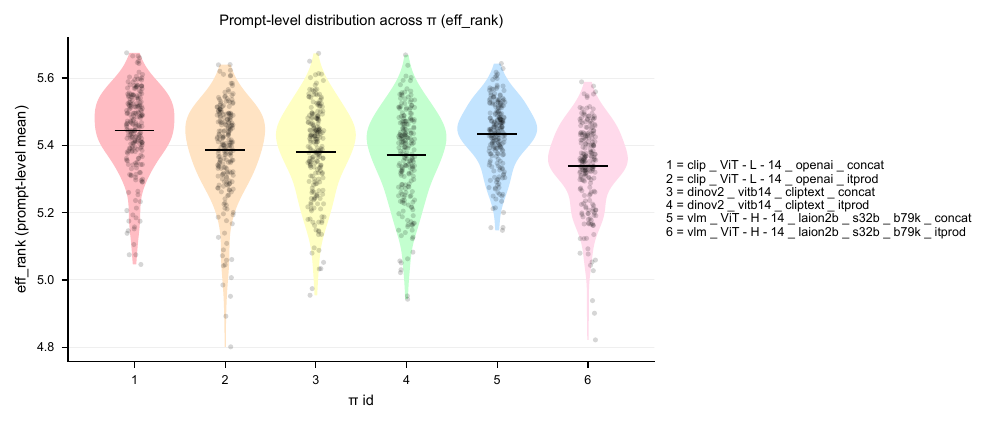}
  \caption{effective-rank (prompt-level means)}
\end{subfigure}\\[1.5mm]
\begin{subfigure}[t]{0.49\textwidth}
  \centering
  \includegraphics[width=\linewidth]{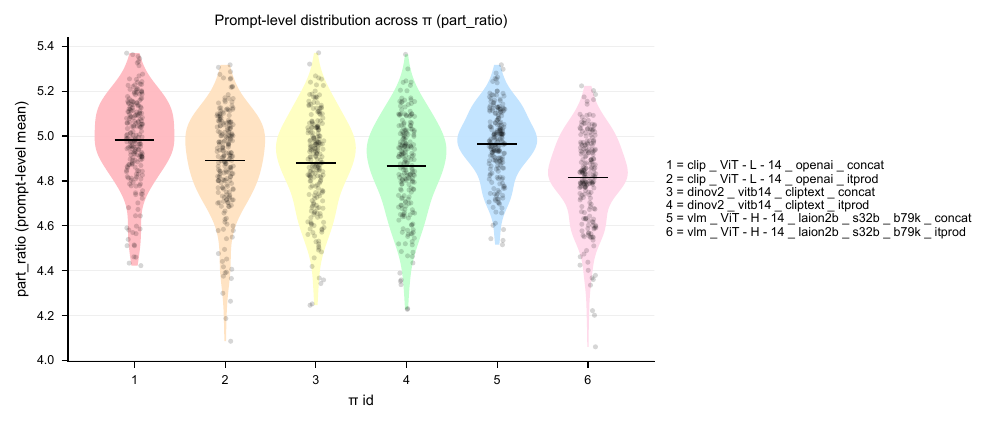}
  \caption{participation ratio (prompt-level means)}
\end{subfigure}\hfill
\begin{subfigure}[t]{0.49\textwidth}
  \centering
  \includegraphics[width=\linewidth]{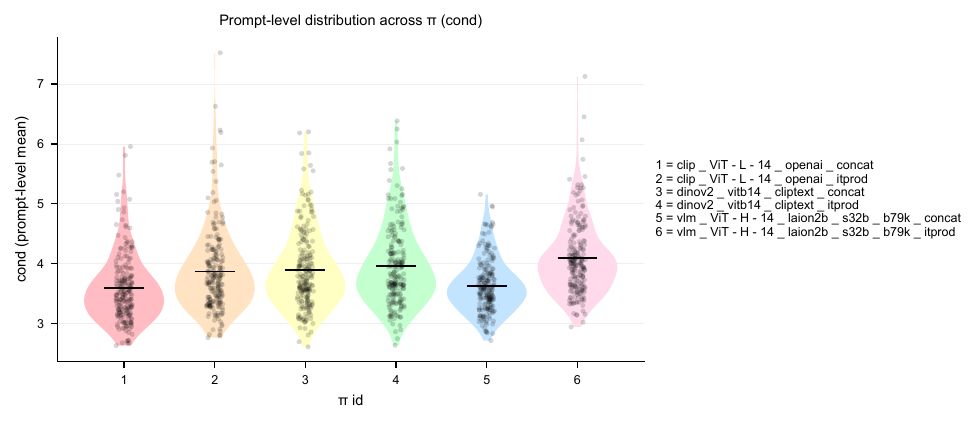}
  \caption{condition number \texttt{cond} (prompt-level means)}
\end{subfigure}
\caption{\textbf{Prompt-level distributions of Jacobian-spectrum statistics across $\pi$.}
Each panel shows the distribution across prompts of the prompt-level mean, with seeds aggregated per prompt, for a given metric and semantic map.
These plots diagnose how the \emph{scale} and \emph{spread} of each local finite-section statistic change with $\pi$.}
\label{fig:pi_violin_all}
\end{figure}

\begin{figure}[t]
\centering
\begin{subfigure}[t]{0.49\textwidth}
  \centering
  \includegraphics[width=\linewidth]{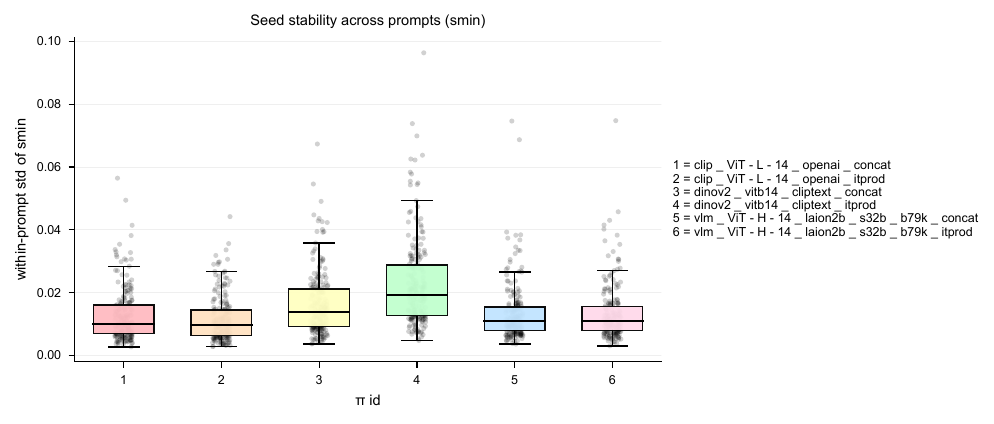}
  \caption{$s_{\min}$: within-prompt seed std}
\end{subfigure}\hfill
\begin{subfigure}[t]{0.49\textwidth}
  \centering
  \includegraphics[width=\linewidth]{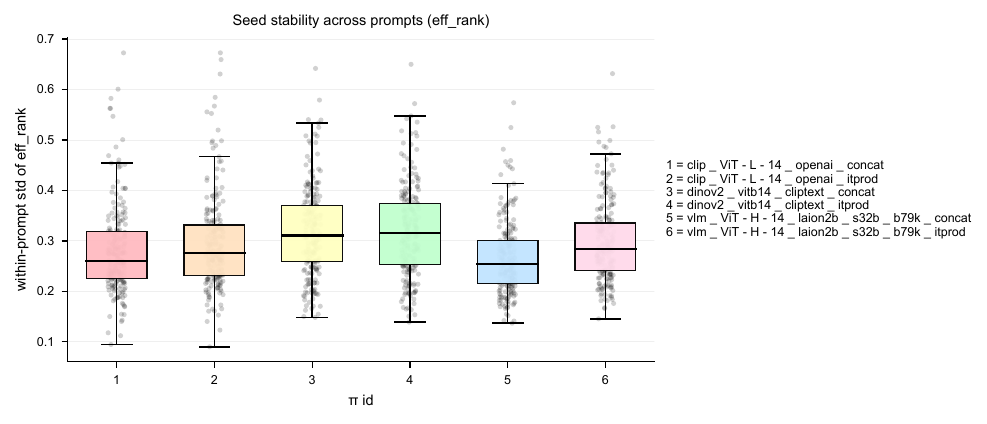}
  \caption{effective-rank: within-prompt seed std}
\end{subfigure}\\[1.5mm]
\begin{subfigure}[t]{0.49\textwidth}
  \centering
  \includegraphics[width=\linewidth]{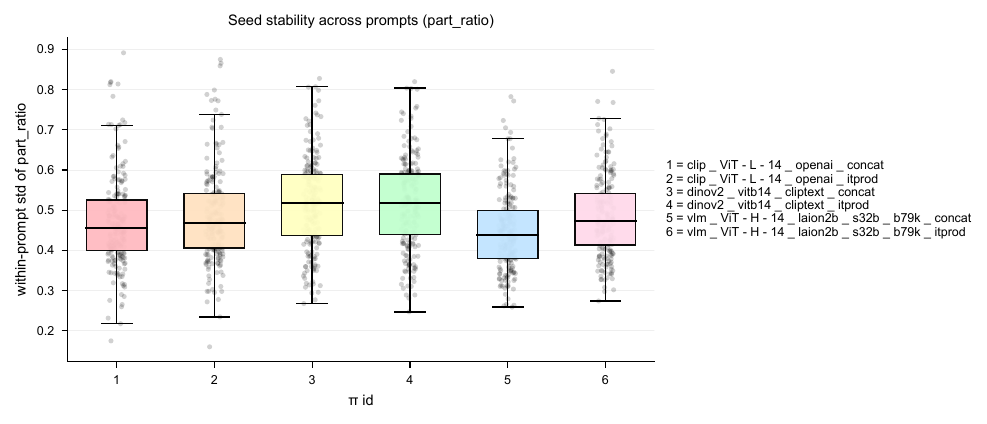}
  \caption{participation ratio: within-prompt seed std}
\end{subfigure}\hfill
\begin{subfigure}[t]{0.49\textwidth}
  \centering
  \includegraphics[width=\linewidth]{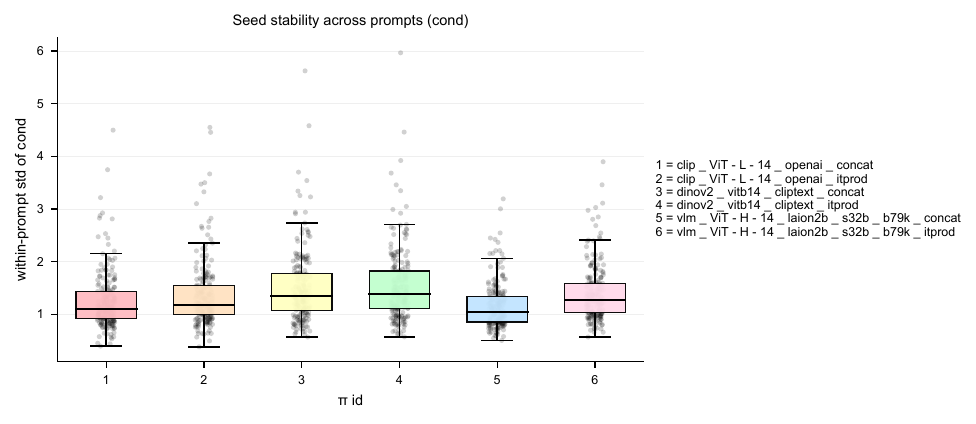}
  \caption{\texttt{cond}: within-prompt seed std}
\end{subfigure}
\caption{\textbf{Seed stability of geometry diagnostics across $\pi$.}
Each panel shows the distribution across prompts of the within-prompt standard deviation over seeds for the indicated metric.
This quantifies how much seed variation affects the diagnostic at the prompt level, and how that sensitivity varies with the semantic map.}
\label{fig:pi_seed_stability_all}
\end{figure}

\begin{figure}[t]
\centering
\begin{subfigure}[t]{0.49\textwidth}
  \centering
  \includegraphics[width=\linewidth]{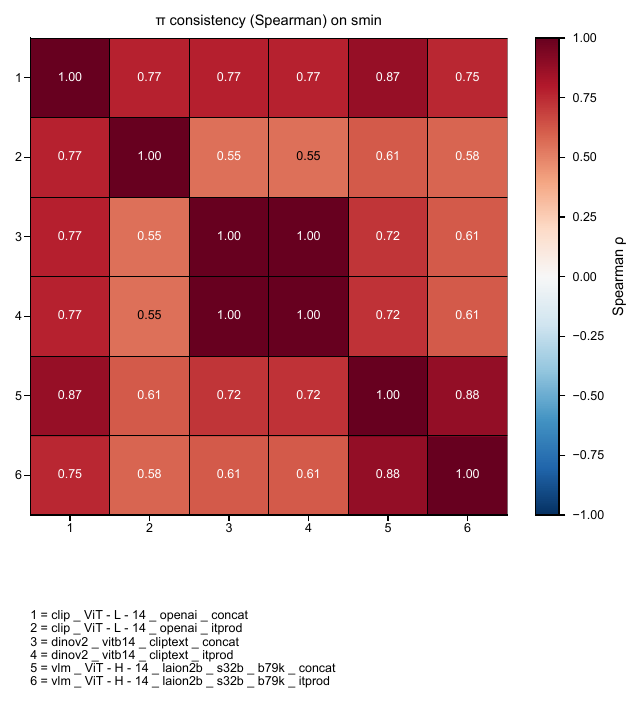}
  \caption{$s_{\min}$: cross-$\pi$ agreement}
\end{subfigure}\hfill
\begin{subfigure}[t]{0.49\textwidth}
  \centering
  \includegraphics[width=\linewidth]{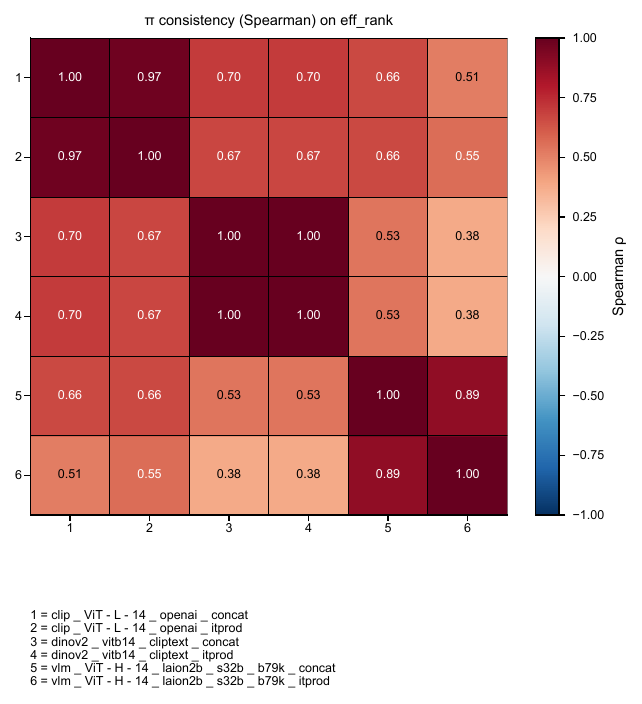}
  \caption{effective-rank: cross-$\pi$ agreement}
\end{subfigure}\\[1.5mm]
\begin{subfigure}[t]{0.49\textwidth}
  \centering
  \includegraphics[width=\linewidth]{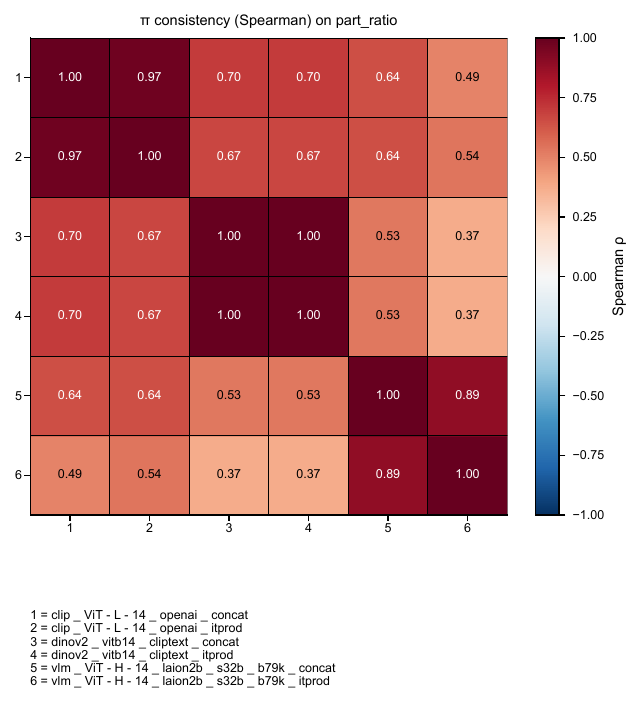}
  \caption{participation ratio: cross-$\pi$ agreement}
\end{subfigure}\hfill
\begin{subfigure}[t]{0.49\textwidth}
  \centering
  \includegraphics[width=\linewidth]{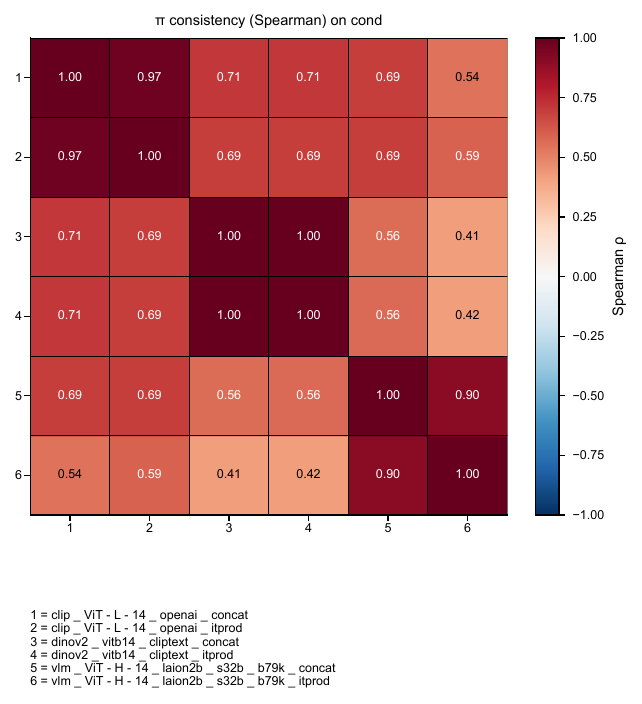}
  \caption{\texttt{cond}: cross-$\pi$ agreement}
\end{subfigure}
\caption{\textbf{Pairwise Spearman agreement across $\pi$ using prompt-level means.}
Each heatmap reports promptwise rank correlation between two choices of $\pi$ for the same diagnostic statistic.
Positive agreement supports the use of the diagnostics for prompt stratification without relying on a single semantic representation.}
\label{fig:pi_heatmaps_all}
\end{figure}

% Optional: geometry--Delta correlation (two representative pi's, four metrics each)
\begin{figure}[t]
\centering
\begin{subfigure}[t]{0.24\textwidth}
  \centering
  \includegraphics[width=\linewidth]{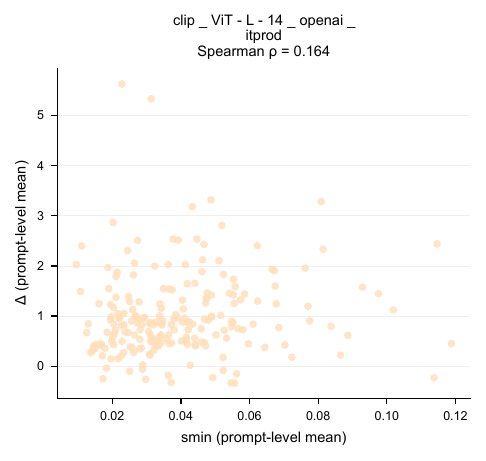}
  \caption{\small CLIP itprod: $s_{\min}$ vs.\ $\Delta$}
\end{subfigure}\hfill
\begin{subfigure}[t]{0.24\textwidth}
  \centering
  \includegraphics[width=\linewidth]{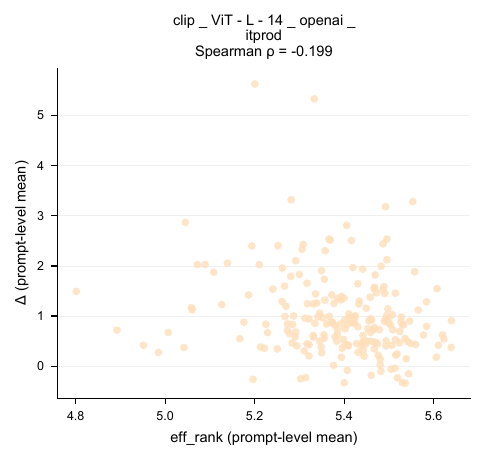}
  \caption{\small CLIP itprod: erank vs.\ $\Delta$}
\end{subfigure}\hfill
\begin{subfigure}[t]{0.24\textwidth}
  \centering
  \includegraphics[width=\linewidth]{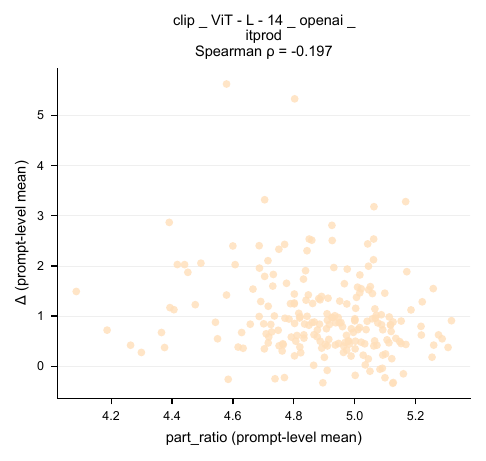}
  \caption{\small CLIP itprod: PR vs.\ $\Delta$}
\end{subfigure}\hfill
\begin{subfigure}[t]{0.24\textwidth}
  \centering
  \includegraphics[width=\linewidth]{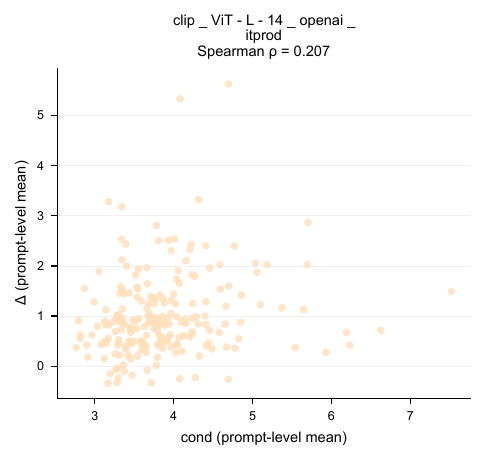}
  \caption{\small CLIP itprod: cond vs.\ $\Delta$}
\end{subfigure}\\[1.5mm]
\begin{subfigure}[t]{0.24\textwidth}
  \centering
  \includegraphics[width=\linewidth]{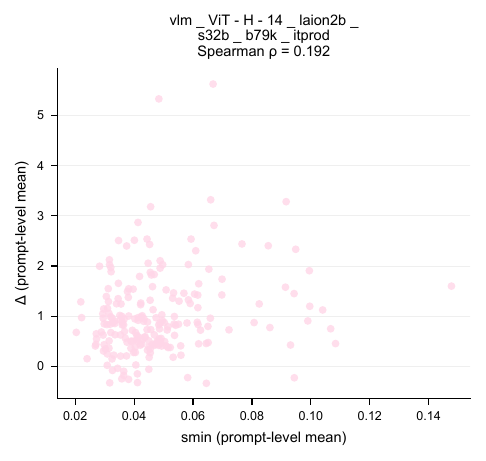}
  \caption{\small VLM itprod: $s_{\min}$ vs.\ $\Delta$}
\end{subfigure}\hfill
\begin{subfigure}[t]{0.24\textwidth}
  \centering
  \includegraphics[width=\linewidth]{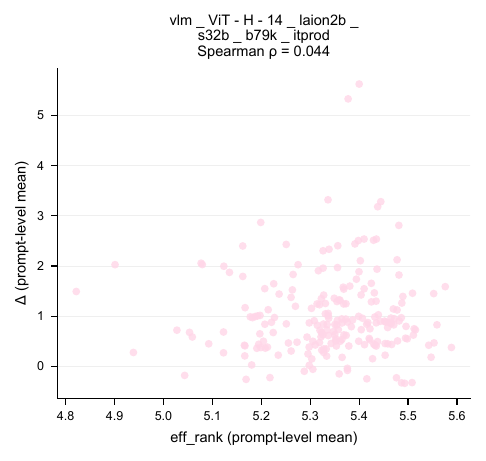}
  \caption{\small VLM itprod: erank vs.\ $\Delta$}
\end{subfigure}\hfill
\begin{subfigure}[t]{0.24\textwidth}
  \centering
  \includegraphics[width=\linewidth]{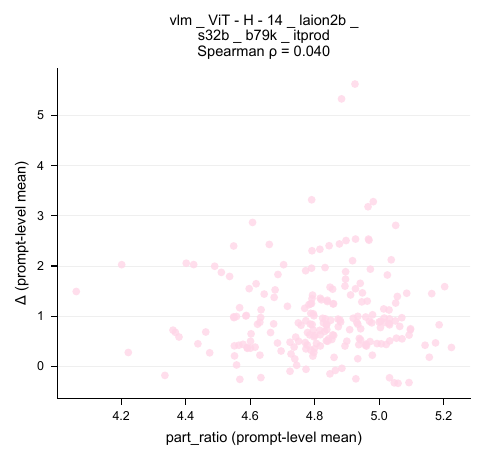}
  \caption{\small VLM itprod: PR vs.\ $\Delta$}
\end{subfigure}\hfill
\begin{subfigure}[t]{0.24\textwidth}
  \centering
  \includegraphics[width=\linewidth]{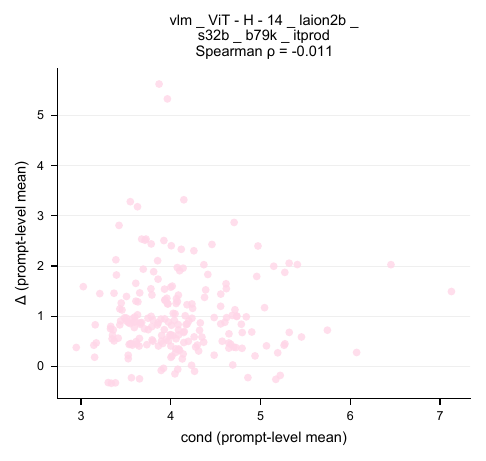}
  \caption{\small VLM itprod: cond vs.\ $\Delta$}
\end{subfigure}
\caption{\textbf{Geometry--$\Delta$ correlations depend on evaluator alignment.}
Each scatter uses prompt-level means and reports Spearman correlation in the title.
Variability across $\pi$ is expected because $\Delta$ is evaluator-dependent; the geometry-only conclusions rely instead on prompt stratification and finite-section spectrum shape.}
\label{fig:geom_delta_examples}
\end{figure}

\begin{figure}[t]
\centering
% ---- DINO itprod ----
\begin{subfigure}[t]{0.24\textwidth}
  \centering
  \includegraphics[width=\linewidth]{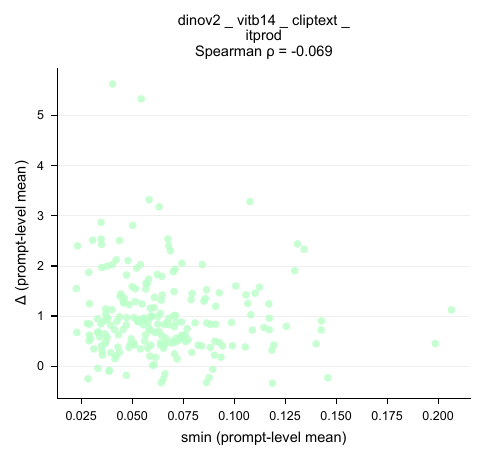}
  \caption{\small DINO itprod: $s_{\min}$ vs.\ $\Delta$}
\end{subfigure}\hfill
\begin{subfigure}[t]{0.24\textwidth}
  \centering
  \includegraphics[width=\linewidth]{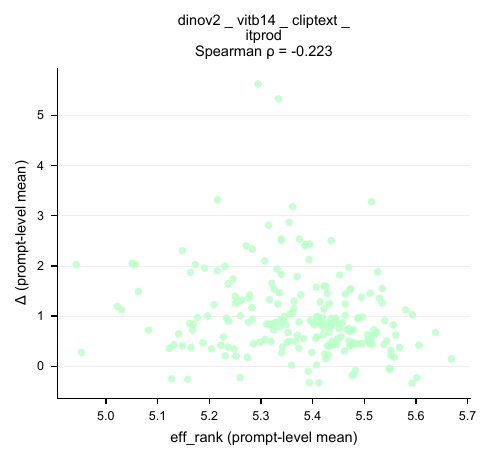}
  \caption{\small DINO itprod: erank vs.\ $\Delta$}
\end{subfigure}\hfill
\begin{subfigure}[t]{0.24\textwidth}
  \centering
  \includegraphics[width=\linewidth]{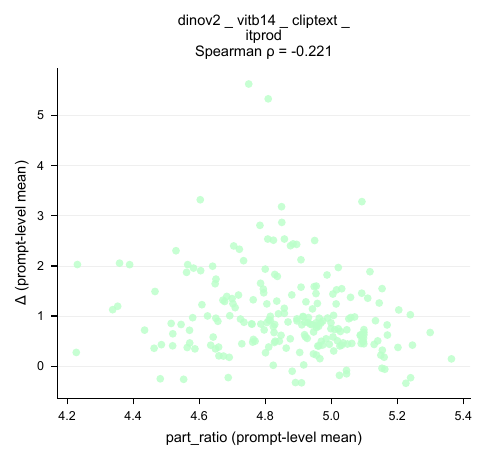}
  \caption{\small DINO itprod: PR vs.\ $\Delta$}
\end{subfigure}\hfill
\begin{subfigure}[t]{0.24\textwidth}
  \centering
  \includegraphics[width=\linewidth]{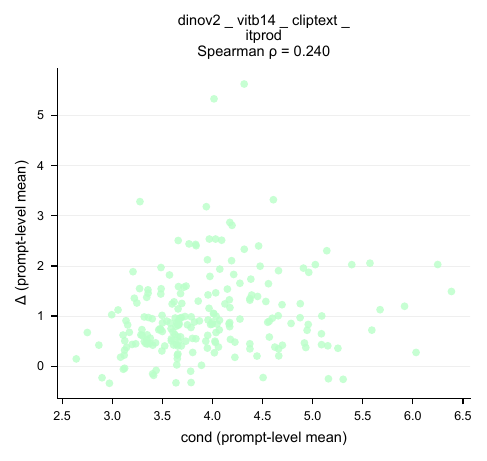}
  \caption{\small DINO itprod: cond vs.\ $\Delta$}
\end{subfigure}\\[1.5mm]

% ---- DINO concat ----
\begin{subfigure}[t]{0.24\textwidth}
  \centering
  \includegraphics[width=\linewidth]{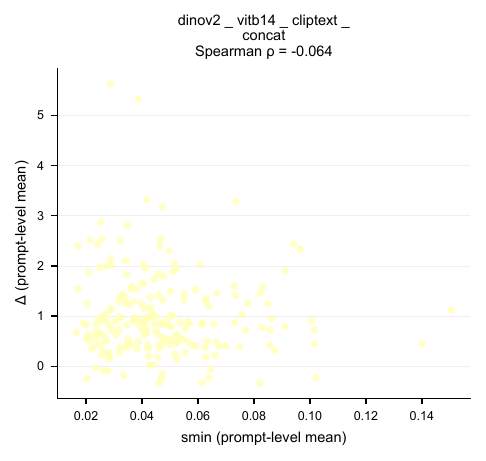}
  \caption{\small DINO concat: $s_{\min}$ vs.\ $\Delta$}
\end{subfigure}\hfill
\begin{subfigure}[t]{0.24\textwidth}
  \centering
  \includegraphics[width=\linewidth]{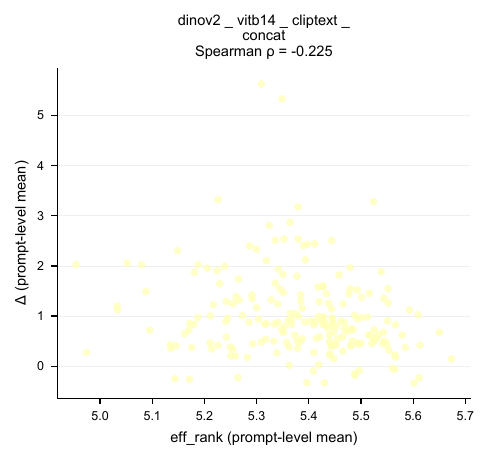}
  \caption{\small DINO concat: erank vs.\ $\Delta$}
\end{subfigure}\hfill
\begin{subfigure}[t]{0.24\textwidth}
  \centering
  \includegraphics[width=\linewidth]{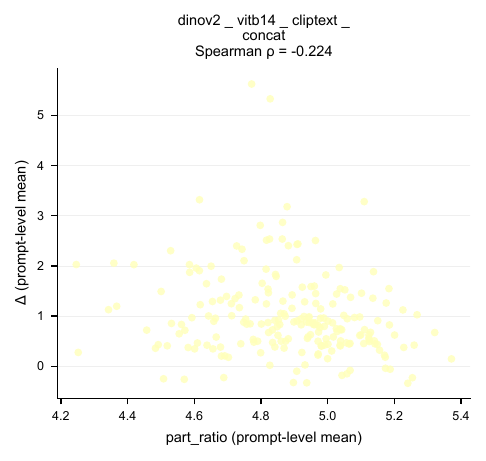}
  \caption{\small DINO concat: PR vs.\ $\Delta$}
\end{subfigure}\hfill
\begin{subfigure}[t]{0.24\textwidth}
  \centering
  \includegraphics[width=\linewidth]{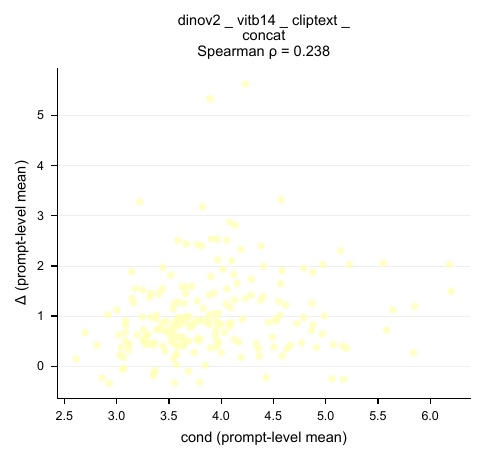}
  \caption{\small DINO concat: cond vs.\ $\Delta$}
\end{subfigure}

\caption{\textbf{Geometry--$\Delta$ correlations for DINO-based $\pi$.}
Same as Fig.~\ref{fig:geom_delta_examples}, but for DINOv2 image features fused with CLIP-text.
These plots are included for completeness; variability across $\pi$ is expected because $\Delta$ is evaluator-dependent.}
\label{fig:geom_delta_dino}
\end{figure}

\begin{table}[t]
\centering
\caption{\textbf{Prompt-averaged geometry diagnostics across $\pi$ (this run).}
Mean and 95\% bootstrap CI are computed across prompts using prompt-level means, with seeds aggregated per prompt.
Seed stability is summarized by the mean within-prompt standard deviation across seeds.
All spectrum statistics are computed from the $r=6$ finite-difference Jacobian sketch.}
\label{tab:pi_geom_summary}

\setlength{\tabcolsep}{5pt}
\renewcommand{\arraystretch}{1.15}

\begin{tabular}{lcccc}
\toprule
$\pi$ (name) &
\makecell{$s_{\min}$\\\footnotesize mean\\\footnotesize [95\% CI]} &
\makecell{erank\\\footnotesize mean\\\footnotesize [95\% CI]} &
\makecell{PR\\\footnotesize mean\\\footnotesize [95\% CI]} &
\makecell{cond\\\footnotesize mean\\\footnotesize [95\% CI]} \\
\midrule

\texttt{\detokenize{clip_ViT-L-14_itprod}} &
\makecell{0.03965\\\footnotesize[0.03703, 0.04239]} &
\makecell{5.3866\\\footnotesize[5.3674, 5.4051]} &
\makecell{4.8914\\\footnotesize[4.8619, 4.9202]} &
\makecell{3.8715\\\footnotesize[3.7811, 3.9691]} \\

\texttt{\detokenize{clip_ViT-L-14_concat}} &
\makecell{0.04267\\\footnotesize[0.04020, 0.04530]} &
\makecell{5.4439\\\footnotesize[5.4269, 5.4600]} &
\makecell{4.9820\\\footnotesize[4.9550, 5.0080]} &
\makecell{3.5917\\\footnotesize[3.5167, 3.6722]} \\

\texttt{\detokenize{vlm_ViT-H-14_itprod}} &
\makecell{0.04799\\\footnotesize[0.04547, 0.05066]} &
\makecell{5.3393\\\footnotesize[5.3217, 5.3559]} &
\makecell{4.8153\\\footnotesize[4.7881, 4.8411]} &
\makecell{4.0897\\\footnotesize[4.0105, 4.1768]} \\

\texttt{\detokenize{vlm_ViT-H-14_concat}} &
\makecell{0.04917\\\footnotesize[0.04682, 0.05160]} &
\makecell{5.4338\\\footnotesize[5.4206, 5.4462]} &
\makecell{4.9647\\\footnotesize[4.9433, 4.9846]} &
\makecell{3.6332\\\footnotesize[3.5763, 3.6945]} \\

\texttt{\detokenize{dinov2_vitb14_itprod}} &
\makecell{0.06581\\\footnotesize[0.06177, 0.06997]} &
\makecell{5.3703\\\footnotesize[5.3519, 5.3889]} &
\makecell{4.8652\\\footnotesize[4.8362, 4.8941]} &
\makecell{3.9588\\\footnotesize[3.8689, 4.0507]} \\

\texttt{\detokenize{dinov2_vitb14_concat}} &
\makecell{0.04712\\\footnotesize[0.04416, 0.05012]} &
\makecell{5.3813\\\footnotesize[5.3628, 5.3998]} &
\makecell{4.8816\\\footnotesize[4.8520, 4.9107]} &
\makecell{3.8933\\\footnotesize[3.8036, 3.9844]} \\

\bottomrule
\end{tabular}

\small
\noindent \\\textbf{Seed stability (mean within-prompt std across seeds).}
$s_{\min}$: \{0.0116, 0.0128, 0.0134, 0.0135, 0.0163, 0.0228\};
erank: \{0.2818--0.3195\};
PR: \{0.4465--0.5166\};
cond: \{1.1624--1.5354\}.
\end{table}

\noindent\textbf{Takeaway.}
Across six strong and diverse semantic maps $\pi$, the geometry-only diagnostics remain well-defined and practically usable.
Within the probed $r=6$ tangent section, the local Jacobian-spectrum statistics show stable finite-section structure across semantic representations, enabling prompt stratification without committing to any single embedding choice.
Downstream geometry--$\Delta$ correlations vary with $\pi$ as expected because $\Delta$ is evaluator-dependent, but this does not affect the core geometry-only conclusion.

\newpage
\clearpage
% =========================================
% =========================================
\section{Additional Proofs for Preliminaries and Method}
\label{app:proofs_extra}
% =========================================

% ============================================================

% ============================================================
\subsection{Proof of Theorem~\ref{thm:degenerate_geometry_main}: Degenerate geometry}
\label{app:proofs:degenerate_geometry}

\begin{proof}
Let $J:=D\Phi_z$ and let $G_{\mathcal M}(\Phi(z))$ be positive definite in a local chart. For any $w\in\mathcal V_z=\Ker(J)$,
\[
\|w\|_{g_{\rm lat}}^2=\langle Jw,Jw\rangle_{g_{\mathcal M}}=0.
\]
Conversely, if $\|u\|_{g_{\rm lat}}=0$, positive definiteness of $g_{\mathcal M}$ implies $Ju=0$, hence $u\in\mathcal V_z$. Thus the nullspace of $g_{\rm lat}$ is exactly $\mathcal V_z$.
Since $\mathcal H_z=\mathcal V_z^\perp$, no nonzero vector in $\mathcal H_z$ lies in this nullspace; hence $g_{\rm lat}$ is positive definite on $\mathcal H_z$.
In matrix form, $G_{\rm lat}(z)=J^\top G_{\mathcal M}J$. Because multiplication by the invertible matrix $G_{\mathcal M}^{1/2}$ does not change rank or kernel after composition with $J$, we have
\[
\Ker(G_{\rm lat}(z))=\Ker(J)=\mathcal V_z,
\qquad
\rank(G_{\rm lat}(z))=\rank(J)=k
\]
on the regular set.
\end{proof}

\subsection{Proof of Lemma~\ref{lem:local_sensitivity_main}: Local semantic sensitivity}
\label{app:proofs:local_sensitivity}

\begin{proof}
Work in a local semantic chart around $\Phi(z)$ and write $J:=D\Phi_z$. Since $\Phi$ is $C^2$ and $J$ is locally Lipschitz, Taylor's theorem gives, for sufficiently small $\Delta z$,
\[
\Phi(z+\Delta z)-\Phi(z)=J\Delta z+R(\Delta z),
\qquad
\|R(\Delta z)\|\le C\|\Delta z\|^2.
\]
Decompose $\Delta z=\Delta z_H+\Delta z_V$ with $\Delta z_H=\Proj_{\mathcal H_z}\Delta z$ and $\Delta z_V=\Proj_{\mathcal V_z}\Delta z$. Since $J\Delta z_V=0$, the linear term is $J\Delta z_H$. On $\mathcal H_z$, the nonzero singular values of $J$ satisfy
\[
\sigma_k(z)\|\Delta z_H\|
\le
\|J\Delta z_H\|
\le
\sigma_1(z)\|\Delta z_H\|.
\]
Local equivalence between the chart norm and the manifold distance $d_{\mathcal M}$ gives constants $c_1,c_2>0$, yielding the stated upper and lower bounds after absorbing Taylor constants into $C$.
If $\|\Delta z_H\|\ge\beta\|\Delta z\|$, then the quadratic term is dominated by the linear term for sufficiently small $\|\Delta z\|$, which gives Eq.~\eqref{eq:local_sens_cone}.
\end{proof}

\subsection{Proof of Theorem~\ref{thm:coarea_main}: Coarea density transport}
\label{app:proofs:coarea}

\begin{proof}
We derive the stated density formula directly from the coarea formula.

\paragraph{Step 1: Coarea formula.}
Let $\Phi:\R^d\to\R^k$ be locally Lipschitz with $k\le d$.
Denote by $J_k(\Phi,z)$ the $k$-dimensional Jacobian (the product of the top $k$ singular values of $D\Phi_z$ on the regular set).
The coarea formula states that for any nonnegative measurable function $h$,
\begin{equation}
\label{eq:app_coarea_formula}
\int_{\R^d} h(z)\,J_k(\Phi,z)\,dz
=
\int_{\R^k}\left(\int_{\Phi^{-1}(x)} h(z)\,d\mathcal H^{d-k}(z)\right)dx,
\end{equation}
where $\mathcal H^{d-k}$ is $(d-k)$-dimensional Hausdorff measure on the level set $\Phi^{-1}(x)$.

In our setting $\Phi:\mathcal Z\to\mathcal M$ maps into a manifold, but locally in a chart $\varphi$ we can apply \eqref{eq:app_coarea_formula} to $\tilde\Phi=\varphi\circ\Phi$; this yields the same expression up to chart Jacobian factors, which are absorbed into the density on $\mathcal M$. The theorem statement is the intrinsic version of this local expression.

\paragraph{Step 2: Express the pushforward probability using an indicator.}
Let $Z\sim p_{\mathcal Z}$ with density w.r.t.\ Lebesgue measure on $\mathcal Z\subset\R^d$.
Let $X=\Phi(Z)$.
For any measurable set $A\subset\mathcal M$,
\begin{align}
\mathbb P(X\in A)
&=\mathbb P(\Phi(Z)\in A)
=\int_{\mathcal Z} \mathbf 1_A(\Phi(z))\,p_{\mathcal Z}(z)\,dz.
\label{eq:app_pushforward_indicator}
\end{align}

\paragraph{Step 3: Insert $J_k(\Phi,z)$ and apply coarea.}
Rewrite the integrand as
\[
\mathbf 1_A(\Phi(z))\,p_{\mathcal Z}(z)
=
\left(\mathbf 1_A(\Phi(z))\frac{p_{\mathcal Z}(z)}{J_k(\Phi,z)}\right)\,J_k(\Phi,z),
\]
on the regular set where $J_k(\Phi,z)>0$ (the singular set is negligible by assump submersion).
Define
\[
h(z):=\mathbf 1_A(\Phi(z))\frac{p_{\mathcal Z}(z)}{J_k(\Phi,z)}.
\]
Then \eqref{eq:app_pushforward_indicator} becomes
\[
\mathbb P(X\in A)
=
\int_{\mathcal Z} h(z)\,J_k(\Phi,z)\,dz.
\]
Applying the coarea formula \eqref{eq:app_coarea_formula} (in local coordinates) gives
\begin{align}
\mathbb P(X\in A)
&=
\int_{\mathcal M}\left(\int_{\Phi^{-1}(x)} h(z)\,d\mathcal H^{d-k}(z)\right)d\mathcal H^k(x) \nonumber\\
&=
\int_{\mathcal M}\mathbf 1_A(x)\left(\int_{\Phi^{-1}(x)}
\frac{p_{\mathcal Z}(z)}{J_k(\Phi,z)}\,d\mathcal H^{d-k}(z)\right)d\mathcal H^k(x).
\label{eq:app_density_read_off}
\end{align}

\paragraph{Step 4: Read off the density.}
Since \eqref{eq:app_density_read_off} holds for all measurable $A$, the term in parentheses must be the density $p_{\mathcal M}(x)$ (w.r.t.\ $k$-dimensional Hausdorff/Riemannian volume measure on $\mathcal M$):
\[
p_{\mathcal M}(x)
=
\int_{\Phi^{-1}(x)}
\frac{p_{\mathcal Z}(z)}{J_k(\Phi,z)}\,d\mathcal H^{d-k}(z),
\]
which is exactly \eqref{eq:coarea_main}.
\end{proof}

% ============================================================
\subsection{Proof of Corollary~\ref{cor:energy_surrogate}: Local ``compression energy'' surrogate}
\label{app:proofs:energy_surrogate}

\begin{proof}
Start from Theorem~\ref{thm:coarea_main}, which implies that along a fiber point $z$ contributes to $p_{\mathcal M}(x)$ with weight
\[
\frac{p_{\mathcal Z}(z)}{J_k(\Phi,z)}.
\]
Taking negative log yields a local ``energy'':
\[
-\log\!\left(\frac{p_{\mathcal Z}(z)}{J_k(\Phi,z)}\right)
=
-\log p_{\mathcal Z}(z) + \log J_k(\Phi,z).
\]

\paragraph{Gaussian prior term.}
For $p_{\mathcal Z}(z)=\mathcal N(0,I_d)$,
\[
p_{\mathcal Z}(z) = (2\pi)^{-d/2}\exp\!\left(-\frac{1}{2}\|z\|^2\right),
\]
so
\[
-\log p_{\mathcal Z}(z)=\frac{1}{2}\|z\|^2 + \text{const.}
\]

\paragraph{Jacobian term and pseudodeterminant.}
On the regular set $\rank(J_z)=k$, the $k$-Jacobian satisfies
\[
J_k(\Phi,z)=\prod_{i=1}^k \sigma_i(z),
\]
where $\{\sigma_i(z)\}_{i=1}^k$ are the nonzero singular values of $J_z$ in local coordinates.
Equivalently,
\[
J_k(\Phi,z)^2 = \det(J_zJ_z^\top).
\]
Since $J_zJ_z^\top\in\R^{k\times k}$ is positive definite at regular points, its determinant equals the product of its (nonzero) eigenvalues; in invariant notation this is the pseudodeterminant $\pdet(J_zJ_z^\top)$, and here $\pdet=\det$.
Hence
\[
\log J_k(\Phi,z)=\frac{1}{2}\log\!\left(J_k(\Phi,z)^2\right)
=
\frac{1}{2}\log \det(J_zJ_z^\top)
=
\frac{1}{2}\log \pdet(J_zJ_z^\top).
\]
Putting the two terms together yields
\[
-\log\!\left(\frac{p_{\mathcal Z}(z)}{J_k(\Phi,z)}\right)
=
\frac{1}{2}\|z\|^2 + \frac{1}{2}\log \pdet(J_zJ_z^\top) + \text{const.},
\]
which is exactly \eqref{eq:eff_energy}.
\end{proof}

\subsection{Derivation of the score-factorized bridge}
\label{app:proofs:score_factorized_bridge}

This subsection makes explicit how the prompt residual used by the algorithm relates to the horizontal--vertical geometry.
The goal is not to prove that every pretrained generator exactly materializes the true horizontal space, but to identify a sufficient score-factorization condition under which the residual has a horizontal component, and to state precisely what the remaining terms represent.

\paragraph{From conditional scores to prompt residuals.}
For an $\epsilon$-parameterized diffusion model with noisy latent $z_t=\alpha_t z_0+\sigma_t\epsilon$, the learned noise predictor determines the conditional score through
\begin{equation}
\label{eq:app_eps_score_relation}
s_\theta(z_t,t,y):=\nabla_{z_t}\log p_\theta(z_t\mid y)
\approx -\frac{1}{\sigma_t}\epsilon_\theta(z_t,t,y),
\end{equation}
up to the usual approximation error of the learned score model.
By Bayes' rule,
\begin{align}
\nabla_{z_t}\log p_\theta(y\mid z_t)
&=\nabla_{z_t}\log p_\theta(z_t\mid y)-\nabla_{z_t}\log p_\theta(z_t) \\
&=s_\theta(z_t,t,y)-s_\theta(z_t,t,\emptyset),
\end{align}
where the unconditional branch is used as the model prior score.
Combining this identity with Eq.~\eqref{eq:app_eps_score_relation} gives
\begin{equation}
\epsilon_\theta(z_t,t,y)-\epsilon_\theta(z_t,t,\emptyset)
\approx -\sigma_t\nabla_{z_t}\log p_\theta(y\mid z_t).
\end{equation}
Thus, for $\epsilon$-prediction, the prompt residual is a time-scaled prompt-likelihood gradient.
The scalar factor is immaterial after normalization; the sign convention is fixed by defining the local potential $\ell_y$ so that the semantic component of the implemented residual has positive coefficient $c_t>0$.

\paragraph{Semantic factorization implies horizontal gradients.}
Assume first that the prompt likelihood depends on the latent only through the semantic map,
\begin{equation}
\label{eq:app_exact_factorization}
\log p_\theta(y\mid z)=\ell_y(\Phi(z)).
\end{equation}
By the chain rule, with $J_z:=D\Phi_z$,
\begin{equation}
\label{eq:app_exact_horizontal_gradient}
\nabla_z\log p_\theta(y\mid z)
=J_z^\top\nabla_\Phi \ell_y(\Phi(z))
\in \Img(J_z^\top).
\end{equation}
At a regular point, $\Img(J_z^\top)=\Ker(J_z)^\perp=\mathcal H_z$, so a purely semantic prompt-likelihood gradient is horizontal.
This is the clean mathematical case behind the proxy construction.

\paragraph{Relaxed factorization and Eq.~\eqref{eq:factorization_to_proxy}.}
In practice, the likelihood need not factor exactly through the chosen semantic map.
Write
\begin{equation}
\label{eq:app_relaxed_factorization}
\log p_\theta(y\mid z)=\ell_y(\Phi(z))+r_y(z),
\end{equation}
where $r_y$ captures prompt-related dependence not represented by $\Phi$.
Then
\begin{equation}
\label{eq:app_relaxed_gradient}
\nabla_z\log p_\theta(y\mid z)=J_z^\top\nabla_\Phi\ell_y+\nabla_z r_y.
\end{equation}
For $\epsilon$-prediction, substituting Eq.~\eqref{eq:app_relaxed_gradient} into the score-residual identity yields
\begin{equation}
\label{eq:app_eps_to_eq19}
g_\theta
= -\sigma_t J_z^\top\nabla_\Phi\ell_y
   -\sigma_t\nabla_z r_y
=J_z^\top a_t(y)+e_t,
\end{equation}
with $a_t(y):=-\sigma_t\nabla_\Phi\ell_y$ and $e_t:=-\sigma_t\nabla_z r_y$.
For velocity- or flow-parameterized backbones, there is no scheduler-independent exact scalar relating the velocity residual to the conditional score.
We therefore write
\begin{equation}
\label{eq:app_flow_mismatch}
g_\theta^{\rm flow}
=c_t\nabla_z\log p_\theta(y\mid z)+\xi_t,
\end{equation}
where $c_t$ is the effective time-dependent calibration induced by the chosen parameterization and scheduler, and $\xi_t$ is the remaining mismatch.
Combining Eq.~\eqref{eq:app_flow_mismatch} with Eq.~\eqref{eq:app_relaxed_gradient} gives
\begin{equation}
\label{eq:app_flow_to_eq19}
g_\theta^{\rm flow}
=c_tJ_z^\top\nabla_\Phi\ell_y+\big(c_t\nabla_z r_y+\xi_t\big)
=J_z^\top a_t(y)+e_t.
\end{equation}
This is Eq.~\eqref{eq:factorization_to_proxy}; it should be read as a sufficient structural approximation, not as an unconditional property of the pretrained model.
A convenient normalized measure of the flow-specific mismatch is
\begin{equation}
\eta_{\rm flow}
:=\frac{\|\xi_t\|}{\|c_t\nabla_z\log p_\theta(y\mid z_t)\|+\varepsilon}.
\end{equation}
When either $\nabla_z r_y$ or $\xi_t$ is large, the vertical or non-semantic component of the proxy can also be large; the method therefore relies on operational validation rather than exact recovery of $\mathcal H_z$.

\paragraph{What the residual term $r_y$ represents.}
The term $r_y$ is not an arbitrary error symbol: it collects prompt-likelihood dependence that is not encoded by the semantic projection $\Phi$.
Examples include rendering artifacts, typography failures, fine-grained counting, spatial relations missed by the chosen embedding, and finite-capacity biases of the backbone.
If $\nabla_z r_y$ dominated the semantic term $J_z^\top\nabla_\Phi\ell_y$, the normalized prompt residual would behave more like a non-semantic or random tangent direction under the radius-preserving probe.
Conversely, for a small spherical perturbation $z(\theta;u)$,
\begin{equation}
D_\pi(z,u;\theta)=\theta\|J_zu\|+O(\theta^2),
\end{equation}
so vertical directions contribute only at second order.
The semantic-displacement percentiles in Table~\ref{tab:semantic_displacement_percentile} therefore provide an indirect constraint: they do not prove $\|\nabla_z r_y\|$ is small or that the vertical component vanishes, but they show that the normalized prompt residual has a semantic-active component far stronger than random tangent controls.
This operational evidence is sufficient for the seed-shaping objective, which only requires a useful tangent direction rather than a certified horizontal projection.

\subsection{Proof of Lemma~\ref{lem:score_horizontal_main}: Shell-projected proxy compatibility}
\label{app:proofs:score_horizontal}

\begin{proof}
Let $J:=D\Phi_z$ and $Q:=Q_z$.
Projecting Eq.~\eqref{eq:factorization_to_proxy} onto the shell tangent space gives
\[
Qg_\theta=c_tQJ^\top\nabla_\Phi\ell_y+Qe_t.
\]
By Lemma~\ref{lem:shell_prompt_gradient}, $QJ^\top\nabla_\Phi\ell_y\in\Img(QJ^\top)=\mathcal H_z^{\rm sh}$.
Therefore its projection onto $\mathcal V_z^{\rm sh}$ is zero, and
\[
\Proj_{\mathcal V_z^{\rm sh}}(Qg_\theta)
=
\Proj_{\mathcal V_z^{\rm sh}}(Qe_t),
\]
which gives Eq.~\eqref{eq:shell_vertical_error_identity}.
For the cone bound, write
\[
Qg_\theta=s+r,
\qquad
s:=\Proj_{\mathcal H_z^{\rm sh}}(Qg_\theta),
\quad
r:=\Proj_{\mathcal V_z^{\rm sh}}(Qg_\theta).
\]
The two components are orthogonal.
If $\|r\|\le\rho\|s\|$ and $u=(s+r)/\|s+r\|$, then
\[
\|\Proj_{\mathcal V_z^{\rm sh}}u\|
=
\frac{\|r\|}{\sqrt{\|s\|^2+\|r\|^2}}
\le
\frac{\rho}{\sqrt{1+\rho^2}},
\]
and the corresponding angle to $\mathcal H_z^{\rm sh}$ is at most $\arctan\rho$.
\end{proof}

\subsection{Proof of Proposition~\ref{prop:cold_start_shell_transfer}: Cold-start shell-proxy transfer}
\label{app:proofs:cold_start_shell_transfer}

\begin{proof}
By adding and subtracting $\widetilde G_{t^\star}(z^{\rm ref})$ and $c_{t^\star}\nabla_{\mathbb S_R}F_y(z^{\rm ref})$, we obtain
\begin{align*}
&\widetilde G_{t^\star}(z_T)-c_{t^\star}\nabla_{\mathbb S_R}F_y(z_T)\\
&\quad=
\big(\widetilde G_{t^\star}(z^{\rm ref})-c_{t^\star}\nabla_{\mathbb S_R}F_y(z^{\rm ref})\big)
+\big(\widetilde G_{t^\star}(z_T)-\widetilde G_{t^\star}(z^{\rm ref})\big)
-c_{t^\star}\big(\nabla_{\mathbb S_R}F_y(z_T)-\nabla_{\mathbb S_R}F_y(z^{\rm ref})\big).
\end{align*}
Taking norms and using the Lipschitz assumptions gives
\[
\|r_T\|
\le
\|r_{\rm ref}\|+(L_G^{\rm sh}+c_{t^\star}L_F^{\rm sh})\|z_T-z^{\rm ref}\|.
\]
If the reference discrepancy is controlled by Proposition~\ref{prop:cold_start_deviation}, the last term is $\mathcal O(T-t^\star)$.
\end{proof}

\subsection{Proof of Theorem~\ref{thm:algorithmic_bridge}: Local algorithmic bridge}
\label{app:proofs:algorithmic_bridge}

\begin{proof}
Let $f:=\nabla_{\mathbb S_R}F_y(z_T)$, $c:=c_{t^\star}$, and $r:=r_T$.
With exact normalization, the update direction is
\[
u_0=\frac{cf+r}{\|cf+r\|}.
\]
The cone assumption gives $\|r\|\le\rho_c c\|f\|$.
Hence
\[
\langle f,cf+r\rangle
\ge
c\|f\|^2-\|f\|\|r\|
\ge
c(1-\rho_c)\|f\|^2,
\]
and
\[
\|cf+r\|
\le
c(1+\rho_c)\|f\|.
\]
Therefore
\begin{equation}
\label{eq:app_directional_gain}
\langle f,u_0\rangle
\ge
\frac{1-\rho_c}{1+\rho_c}\|f\|.
\end{equation}
If the $\varepsilon$-stabilized normalization is used and $\|\bar g_T\|\ge m_g>0$ locally, then the direction changes by $O(\varepsilon/m_g)$, producing the $O(\delta\varepsilon/m_g)$ term. With exact normalization this term is absent.

Because $u_0\perp z_T$ and $\|u_0\|=1$, the spherical retraction satisfies
\[
z_T^\star
=
z_T+
\delta u_0
-
\frac{\delta^2}{2\|z_T\|^2}z_T
+O\!\left(\frac{\delta^3}{\|z_T\|^2}\right).
\]
Taylor expanding $F_y$ around $z_T$ gives
\[
F_y(z_T^\star)-F_y(z_T)
=
\delta\langle \nabla_{\mathbb S_R}F_y(z_T),u_0\rangle+O(\delta^2).
\]
Combining this expansion with Eq.~\eqref{eq:app_directional_gain} proves Eq.~\eqref{eq:algorithmic_bridge_gain}.
The statement about cold-start error follows by substituting the residual bound of Proposition~\ref{prop:cold_start_shell_transfer} into the cone residual term.
\end{proof}

\subsection{Proof of Proposition~\ref{prop:retraction_opt}: Retraction as projection onto the sphere}
\label{app:proofs:retraction_opt}

\begin{proof}
Let $R:=\|z_T\|$ and $u:=z_T+\delta\hat v$.
We need to solve
\[
\min_{z\in\R^d}\ \|z-u\|^2
\quad\text{s.t.}\quad \|z\|=R.
\]

\paragraph{Method 1: Geometry (maximize inner product).}
Expand:
\[
\|z-u\|^2=\|z\|^2+\|u\|^2-2\langle z,u\rangle
=
R^2+\|u\|^2-2\langle z,u\rangle.
\]
Since $R^2+\|u\|^2$ is constant under the constraint, minimizing $\|z-u\|^2$ is equivalent to maximizing $\langle z,u\rangle$
subject to $\|z\|=R$.
By Cauchy--Schwarz, $\langle z,u\rangle\le \|z\|\|u\|=R\|u\|$, with equality iff $z$ is a positive scalar multiple of $u$.
Thus the maximizer is
\[
z^\star = R\frac{u}{\|u\|}
=
\|z_T\|\cdot\frac{z_T+\delta\hat v}{\|z_T+\delta\hat v\|},
\]

\paragraph{Method 2: Lagrange multipliers.}
Consider $\mathcal L(z,\lambda)=\|z-u\|^2+\lambda(\|z\|^2-R^2)$.
Stationarity in $z$ gives $2(z-u)+2\lambda z=0$, i.e.\ $(1+\lambda)z=u$.
Thus $z=\frac{1}{1+\lambda}u$. Enforce $\|z\|=R$ to get $|1+\lambda|=\|u\|/R$ and choose the sign giving the minimizer (same direction as $u$), yielding $z^\star=R u/\|u\|$.
\end{proof}

% ============================================================

% ============================================================
\subsection{Proof of Lemma~\ref{lem:small_delta}: First-order shell retraction}
\label{app:proofs:small_delta}

\begin{proof}
Let $R=\|z_T\|$, $\hat z=z_T/R$, and let $u\perp z_T$ with $\|u\|=1$.
Then
\[
\|z_T+\delta u\|
=
\sqrt{R^2+\delta^2}
=
R\sqrt{1+\delta^2/R^2}.
\]
Therefore
\[
z_T^+
=
R\frac{z_T+\delta u}{\sqrt{R^2+\delta^2}}
=
(z_T+\delta u)\left(1+\frac{\delta^2}{R^2}\right)^{-1/2}.
\]
Using $(1+s)^{-1/2}=1-s/2+O(s^2)$ gives
\[
z_T^+
=
z_T+
\delta u
-
\frac{\delta^2}{2R^2}z_T
+O\!\left(\frac{\delta^3}{R^2}\right),
\]
where the remainder includes the product of $\delta u$ with $O(\delta^2/R^2)$.
Exact radius preservation follows directly from the definition.
Finally, for any $C^2$ function $F_y$, Taylor's theorem gives
\[
F_y(z_T^+)-F_y(z_T)
=
\langle \nabla F_y(z_T),z_T^+-z_T\rangle+O(\|z_T^+-z_T\|^2).
\]
The first-order displacement is the tangent vector $\delta u$, and $u\in T_{z_T}\mathbb S_R$.
Thus $\langle\nabla F_y(z_T),u\rangle=\langle\nabla_{\mathbb S_R}F_y(z_T),u\rangle$, while all remaining terms are $O(\delta^2)$.
\end{proof}

\subsection{Proof of Proposition~\ref{prop:cold_start_deviation}: 
Cold-start state deviation}
\label{app:proofs:cold_start_deviation}

\begin{proof}
We use a reverse-time reparameterization and Gr\"onwall's inequality 
in integral form to avoid differentiating $\|e_\tau\|$ at $\tau=0$.

\paragraph{Step 1: Reverse-time reparameterization.}
Set $\tau:=T-t$, and define
\[
    \tilde z_\tau:=z_{T-\tau}^{\mathrm{traj}},
    \qquad
    \tilde z_\tau^{\mathrm{cs}}:=\rho(T-\tau)z_T,
    \qquad \tau\in[0,h].
\]
Then $\tilde z_0=\tilde z_0^{\mathrm{cs}}=z_T$, and the chain rule gives
\[
    \frac{d\tilde z_\tau}{d\tau}
    =-v_\theta(\tilde z_\tau,T-\tau;y),
    \qquad
    \frac{d\tilde z_\tau^{\mathrm{cs}}}{d\tau}
    =-\rho'(T-\tau)\,z_T.
\]

\paragraph{Step 2: Integral form of the deviation.}
Let $e_\tau:=\tilde z_\tau-\tilde z_\tau^{\mathrm{cs}}$. Since $e_0=0$, 
the fundamental theorem of calculus yields
\[
    e_\tau=\int_0^\tau \frac{de_s}{ds}\,ds
    =\int_0^\tau \big[-v_\theta(\tilde z_s,T-s;y)+\rho'(T-s)\,z_T\big]\,ds.
\]
Adding and subtracting $v_\theta(\tilde z_s^{\mathrm{cs}},T-s;y)$ inside 
the integrand and applying the triangle inequality:
\begin{align*}
\|e_\tau\| 
&\le \int_0^\tau \|v_\theta(\tilde z_s,T-s;y) - v_\theta(\tilde z_s^{\mathrm{cs}},T-s;y)\|\,ds \\
&\quad + \int_0^\tau \|v_\theta(\tilde z_s^{\mathrm{cs}},T-s;y) - \rho'(T-s)\,z_T\|\,ds.
\end{align*}

\paragraph{Step 3: Apply the Lipschitz and boundedness assumptions.}
By the $L_v$-Lipschitz assumption on $v_\theta(\cdot,t;y)$,
\[
\|v_\theta(\tilde z_s,T-s;y) - v_\theta(\tilde z_s^{\mathrm{cs}},T-s;y)\|
\le L_v\|e_s\|.
\]
By the boundedness assumptions $\|v_\theta\|\le B_v$ and $|\rho'|\le L_\rho$,
\[
\|v_\theta(\tilde z_s^{\mathrm{cs}},T-s;y) - \rho'(T-s)\,z_T\|
\le B_v + L_\rho\|z_T\|.
\]
Substituting:
\[
\|e_\tau\| \le L_v\int_0^\tau \|e_s\|\,ds + (B_v+L_\rho\|z_T\|)\,\tau.
\]

\paragraph{Step 4: Gr\"onwall's inequality (integral form).}
The inequality $\|e_\tau\|\le \alpha(\tau)+L_v\int_0^\tau \|e_s\|\,ds$, with $\alpha(\tau)=(B_v+L_\rho\|z_T\|)\tau$, gives by the Gr\"onwall--Bellman inequality
\[
\|e_\tau\| \le \alpha(\tau)+L_v\int_0^\tau \alpha(s)e^{L_v(\tau-s)}\,ds.
\]
Since
\[
L_v\int_0^\tau s e^{L_v(\tau-s)}\,ds
= \frac{e^{L_v\tau}-1}{L_v}-\tau,
\]
we obtain
\[
\|e_\tau\| \le (B_v+L_\rho\|z_T\|)\left[\tau+\frac{e^{L_v\tau}-1}{L_v}-\tau\right]
= (B_v+L_\rho\|z_T\|)\frac{e^{L_v\tau}-1}{L_v}.
\]

\paragraph{Step 5: Conclude.}
Setting $\tau=h$ gives Eq.~\eqref{eq:cold_start_state_bound}; the case $L_v=0$ follows from Taylor expansion of $e^{L_vh}$.
If $\rho\equiv 1$ on $[t^\star,T]$, the direct bound
$\|z_{t^\star}^{\rm traj}-z_T\|\le\int_{t^\star}^{T}\|v_\theta(z_t,t;y)\|dt\le B_v(T-t^\star)$ also applies.
\end{proof}
\clearpage

% ============================
\section{Additional Experiments: Visualizing Degenerate Latent Geometry}
\label{app:degeneracy_viz}
% ============================

This appendix details the empirical diagnostics used to visualize the pullback geometry induced by the semantic map $\Phi=\pi\circ\Psi$.
Our theory predicts that semantic sensitivity is anisotropic on the Gaussian typical-radius shell: most tangent directions induce small semantic changes, while a smaller set of prompt-relevant directions produces stronger semantic movement.
We verify this through three complementary diagnostics: direction-sensitivity distributions, local spectrum estimation, and prompt-residual proxy comparisons.

\subsection{Experimental Setup and Notation}
\label{app:degeneracy_viz:setup}

\textbf{Backbone and sampler.}
We use FLUX.1-Dev with the official inference configuration in Appendix~\ref{E_D}.
All probes modify only the initial seed while keeping the prompt and sampler fixed.

\textbf{Prompts and seeds.}
We evaluate complex prompts involving multi-object composition, typography, and spatial relations, with multiple random seeds per prompt.
Reported curves and distributions aggregate over prompts and seeds, while strip visualizations show representative examples.

\textbf{Semantic features and metrics.}
Let $x(z_T)$ denote the generated output from initial noise $z_T$ under prompt $y$.
We use CLIP image/text embeddings $e_I(x)$ and $e_T(y)$, normalized as $\hat e_I$ and $\hat e_T$.

\begin{equation}
  \mathrm{CLIPSim}(x,y)=\langle \hat e_I(x), \hat e_T(y)\rangle .
\end{equation}

For a baseline seed $z_T$ and perturbed seed $z_T'$, we measure semantic displacement by
\begin{equation}
  \|\Delta\mathrm{CLIP}\|
  =
  \left\|\hat e_I(x(z_T'))-\hat e_I(x(z_T))\right\|_2 .
  \label{eq:app_delta_clip}
\end{equation}

\subsection{Typical-Radius Shell Probing}
\label{app:degeneracy_viz:geodesic}

To probe sensitivity while staying on the Gaussian typical-radius shell, we fix $R=\|z_T\|$.
Let $\hat z=z_T/\|z_T\|$. A random tangent direction is obtained by
\begin{equation}
  \tilde u := u-\langle u,\hat z\rangle \hat z,
  \qquad
  \hat u := \tilde u/\|\tilde u\|.
  \label{eq:app_tangent_proj}
\end{equation}
We then use the radius-preserving spherical geodesic
\begin{equation}
  z_T(\theta)=R\big(\cos\theta\,\hat z+\sin\theta\,\hat u\big),
  \label{eq:app_geodesic}
\end{equation}
which satisfies $\|z_T(\theta)\|=R$ for all $\theta$.
For small $\delta/R$, this is equivalent to a tangent step followed by spherical retraction,
\begin{equation}
  z_T' = R\cdot \frac{z_T+\delta \hat u}{\|z_T+\delta \hat u\|},
\end{equation}
matching the retraction used in our method.

\subsection{Direction Sensitivity Distribution}
\label{app:degeneracy_viz:dir_sens_method}

For each prompt--seed pair, we sample $M$ random tangent directions, perturb each seed by a small angle $\theta_0$, and compute $\|\Delta\mathrm{CLIP}\|$.
We aggregate the resulting semantic displacements and visualize both the histogram and CCDF.

For small $\theta_0$, the semantic change follows the local linear approximation
\begin{equation}
  \Phi(z_T(\theta_0))-\Phi(z_T)
  \approx
  D\Phi_{z_T}\big(R\theta_0\,\hat u\big).
\end{equation}
Thus, the distribution of $\|\Delta\mathrm{CLIP}\|$ across random tangent directions reveals the anisotropy of semantic sensitivity.

\begin{figure}[t]
  \centering
  \begin{subfigure}[t]{0.49\textwidth}
    \centering
    \includegraphics[width=\linewidth]{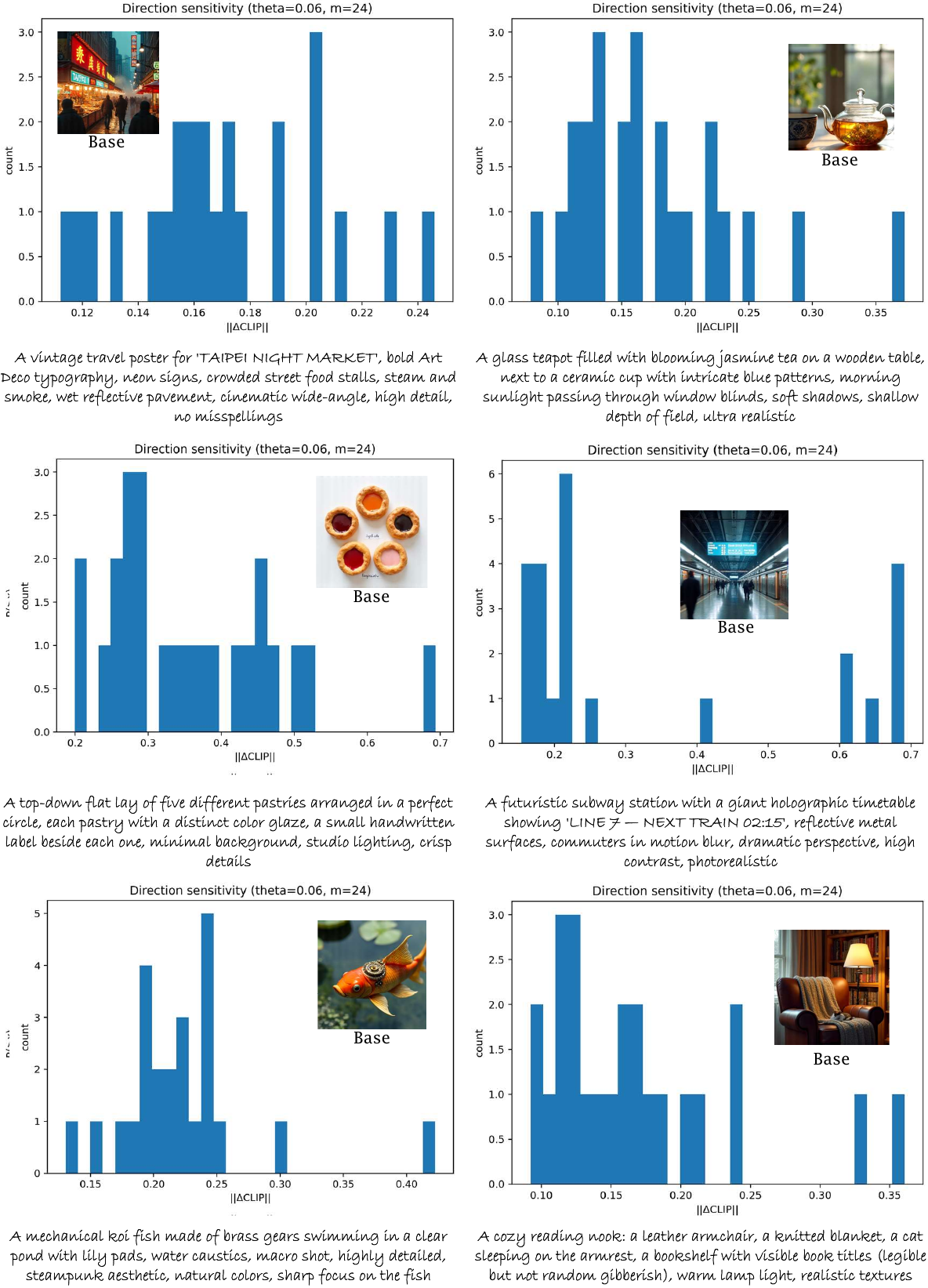}
    \caption{Histogram of $\|\Delta\mathrm{CLIP}\|$ over random tangent directions.}
    \label{fig:app_dir_sens_hist}
  \end{subfigure}
  \hfill
  \begin{subfigure}[t]{0.49\textwidth}
    \centering
    \includegraphics[width=\linewidth]{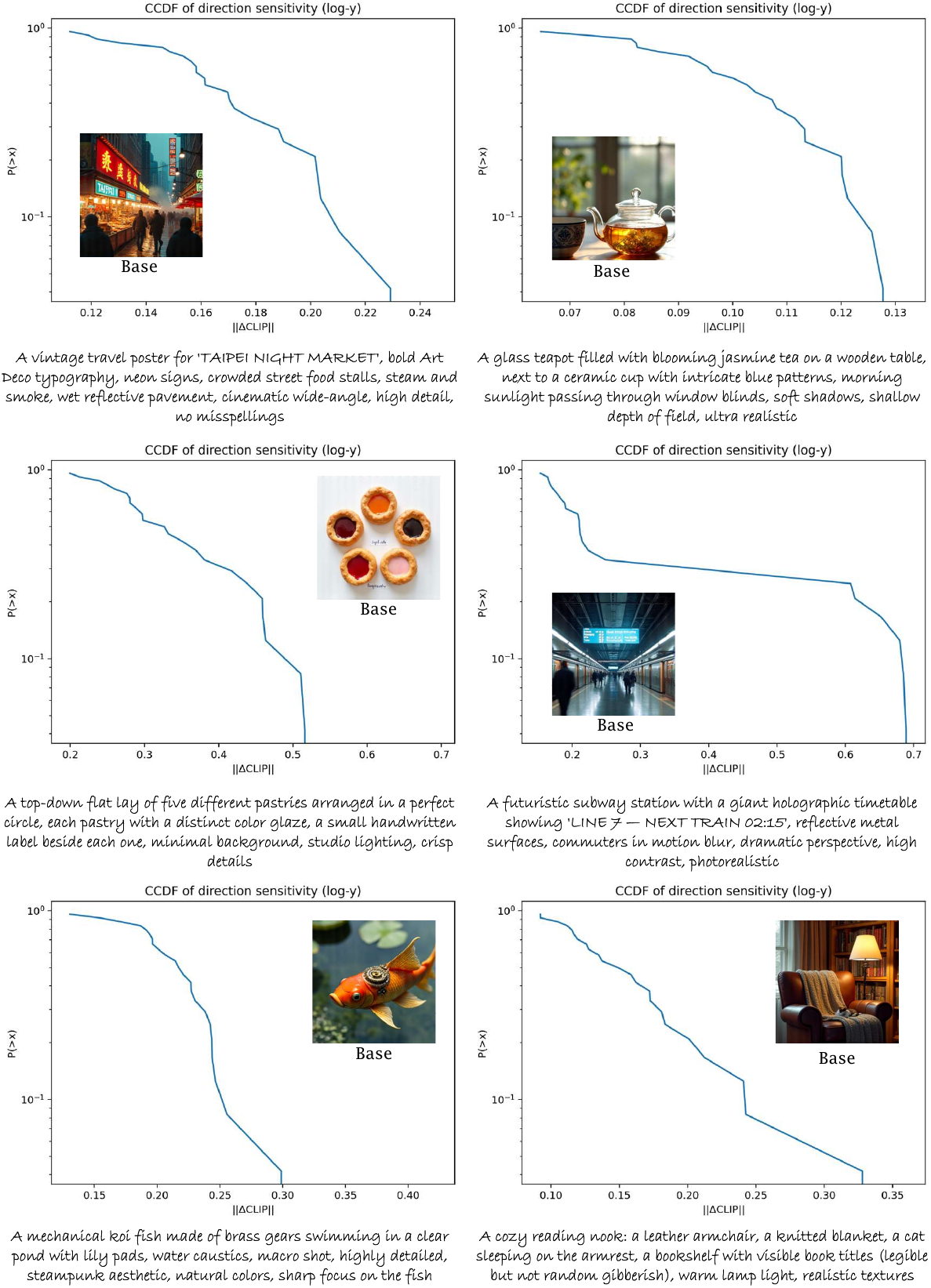}
    \caption{CCDF $\mathbb P(\|\Delta\mathrm{CLIP}\|\ge t)$ showing the tail of large semantic changes.}
    \label{fig:app_dir_sens_ccdf}
  \end{subfigure}
  \caption{\textbf{Anisotropic semantic sensitivity on the Gaussian radius shell.}
  Most tangent directions induce small semantic displacement, while a subset produces strong semantic changes, consistent with the pullback-geometry view.}
  \label{fig:app_dir_sens}
\end{figure}

\subsection{Local Spectrum Estimation}
\label{app:degeneracy_viz:spectrum_method}

We further estimate the effective local spectrum of semantic sensitivity.
Let $\phi(z_T):=\hat e_I(x(z_T))$ be the normalized CLIP image embedding.
For each tangent direction $\hat u_j$, we compute the symmetric geodesic finite difference
\begin{equation}
  g_j
  =
  \frac{\phi(z_T(\theta_0;\hat u_j))-\phi(z_T(-\theta_0;\hat u_j))}{2\theta_0}
  \in \mathbb R^D .
  \label{eq:app_dir_deriv}
\end{equation}
Stacking $g_j$ as columns of $G=[g_1,\ldots,g_M]$, the spectrum of $G^\top G$ provides a randomized estimate of the dominant local sensitivity modes.

\begin{figure}[t]
  \centering
  \includegraphics[width=\linewidth]{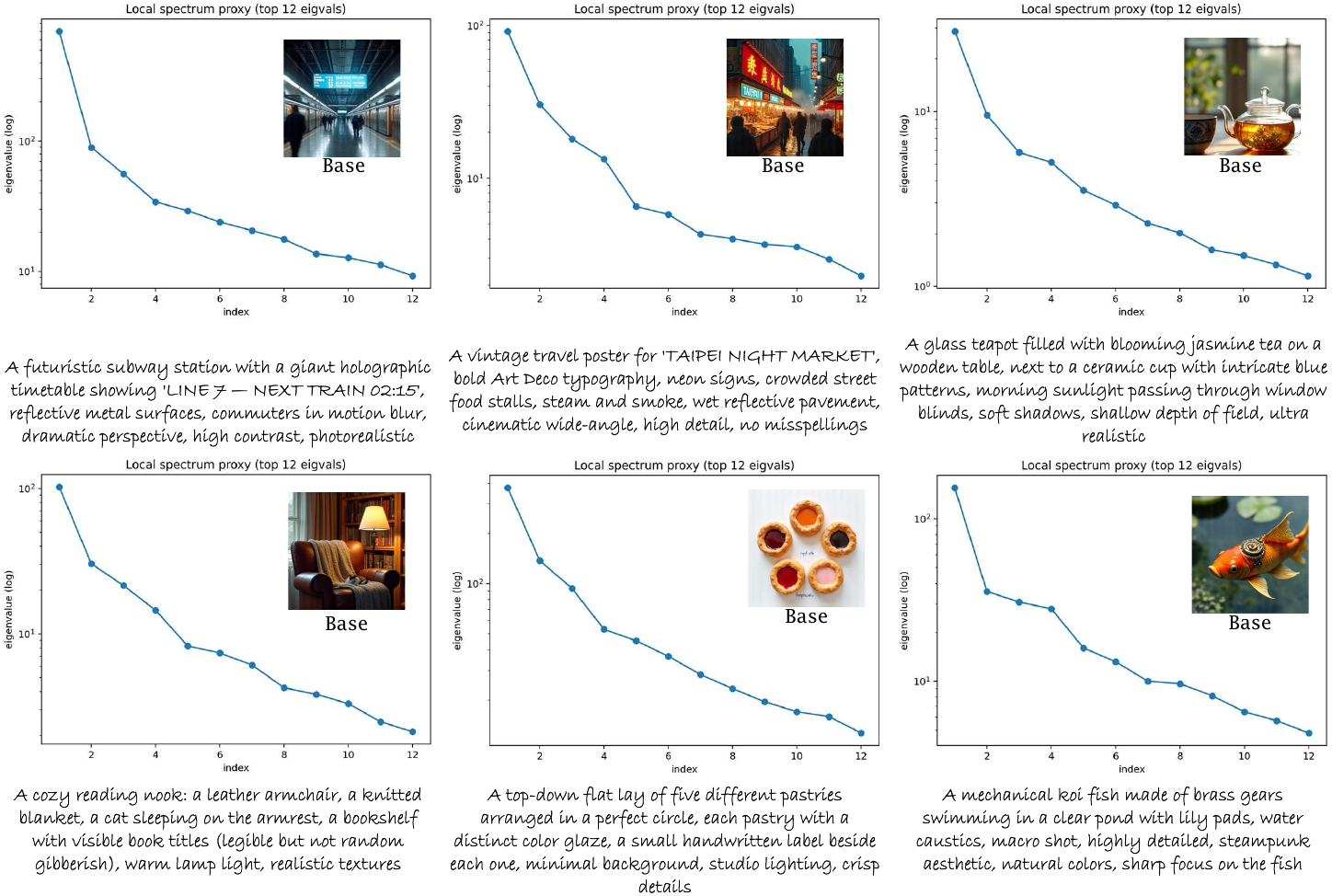}
  \caption{\textbf{Local spectrum proxy.}
  The spectrum summarizes how semantic sensitivity concentrates across dominant tangent directions.}
  \label{fig:app_local_spectrum}
\end{figure}

A concentrated spectrum indicates that semantic changes are governed by a small set of dominant directions.
Across prompts, the spectral shape also reflects conditioning-dependent structure, especially for typography, layout, and compositional prompts.

\subsection{Prompt-Residual Proxy vs.\ Control Directions}
\label{app:degeneracy_viz:hv_method}

Since $D\Phi$ is not explicitly available during sampling, we use the model-derived prompt residual as an operational proxy direction.
At probing timestep $t$, we compute
\begin{equation}
  r := \epsilon_\theta(z_t,t,y)-\epsilon_\theta(z_t,t,\varnothing),
\end{equation}
normalize it, and project it to the tangent space:
\begin{equation}
  \hat u_{\mathrm H}
  =
  \frac{\hat r-\langle \hat r,\hat z\rangle \hat z}
       {\|\hat r-\langle \hat r,\hat z\rangle \hat z\|}.
  \label{eq:app_uH}
\end{equation}
We compare this prompt-residual proxy with an independently sampled random tangent control $\hat u_{\mathrm C}$ under matched radius-preserving perturbations.

\subsection{Operational Diagnostics for the Prompt-Residual Proxy}
\label{app:proxy_diagnostics}

We evaluate whether the prompt-residual proxy behaves as a structured tangent direction on FLUX.1-Dev.

\paragraph{Semantic-displacement percentile.}
For each matched prompt--seed pair, we compare the proxy direction against random tangent controls.
Table~\ref{tab:semantic_displacement_percentile} shows that the proxy lies in the upper tail of random controls for CLIP displacement metrics.

\begin{table}[t]
\centering
\caption{\textbf{Semantic-displacement percentile of the prompt-residual proxy on FLUX.}
Percentiles are computed over 48 matched prompt--seed pairs against random tangent controls.}
\label{tab:semantic_displacement_percentile}
\small
\setlength{\tabcolsep}{5pt}
\begin{tabular}{lccc}
\toprule
Metric & Mean percentile $\uparrow$ & Median percentile $\uparrow$ & Top-2 rate $\uparrow$ \\
\midrule
CLIP L2 displacement & 0.995 & 1.000 & 1.000 \\
CLIP cosine displacement & 0.997 & 1.000 & 1.000 \\
Alignment absolute change & 0.922 & 1.000 & 0.854 \\
Alignment signed change & 0.529 & 0.578 & 0.438 \\
\bottomrule
\end{tabular}
\end{table}

\paragraph{Residual concentration.}
For local prompt variants at $t^\star=900$, Table~\ref{tab:residual_concentration} shows that prompt-residual directions form a concentrated family compared with random tangent controls.
The tangential projection retains nearly all residual norm, with mean radial ratio $0.066$ and retained tangent norm ratio $0.997$.

\begin{table}[t]
\centering
\caption{\textbf{Concentration of prompt-residual directions at $t^\star=900$.}
Local prompt variants induce a concentrated residual family compared with random tangent controls.}
\label{tab:residual_concentration}
\small
\setlength{\tabcolsep}{5pt}
\begin{tabular}{lccc}
\toprule
Direction family & Pairwise cosine $\uparrow$ & Effective rank $\downarrow$ & Top-5 energy $\uparrow$ \\
\midrule
Local prompt variants & 0.723 & 9.373 & 0.766 \\
Random tangent controls & 0.000 & 28.999 & 0.175 \\
\bottomrule
\end{tabular}
\end{table}

\paragraph{High-noise direction consistency.}
We compare prompt-residual directions at different high-noise timesteps with the default $t^\star=900$ direction.
Table~\ref{tab:high_noise_consistency} shows consistent alignment above the random tangent baseline around the default probe.
Across these timesteps, the retained tangent norm ratio remains high, ranging from $0.998$ to $1.000$.

\begin{table}[t]
\centering
\caption{\textbf{High-noise consistency of prompt-residual directions on FLUX.}
Cosine similarity is computed with the $t=900$ direction as reference.}
\label{tab:high_noise_consistency}
\small
\setlength{\tabcolsep}{6pt}
\begin{tabular}{lccc}
\toprule
Timestep & Cosine with $v_{900}$ $\uparrow$ & 95\% CI & Random baseline \\
\midrule
1000 & 0.150 & [0.132, 0.166] & $\approx 0$ \\
950  & 0.461 & [0.421, 0.501] & $\approx 0$ \\
900  & 1.000 & [1.000, 1.000] & $\approx 0$ \\
850  & 0.215 & [0.187, 0.243] & $\approx 0$ \\
800  & 0.095 & [0.077, 0.113] & $\approx 0$ \\
\bottomrule
\end{tabular}
\end{table}

Together, these diagnostics show that the prompt-residual proxy is structured, concentrated, and semantically active on the Gaussian radius shell.

\subsection{Geodesic Curves and Strip Visualizations}
\label{app:degeneracy_viz:curves_method}

For each $(y,z_T)$, we evaluate a grid of angles $\theta$ along both the prompt-residual proxy direction $\hat u_{\mathrm H}$ and the random control direction $\hat u_{\mathrm C}$.
We report semantic displacement $\|\Delta\mathrm{CLIP}\|(\theta)$ and alignment $\mathrm{CLIPSim}(\theta)$.

\begin{figure}[t]
  \centering
  \begin{subfigure}[t]{0.49\textwidth}
    \centering
    \includegraphics[width=\linewidth]{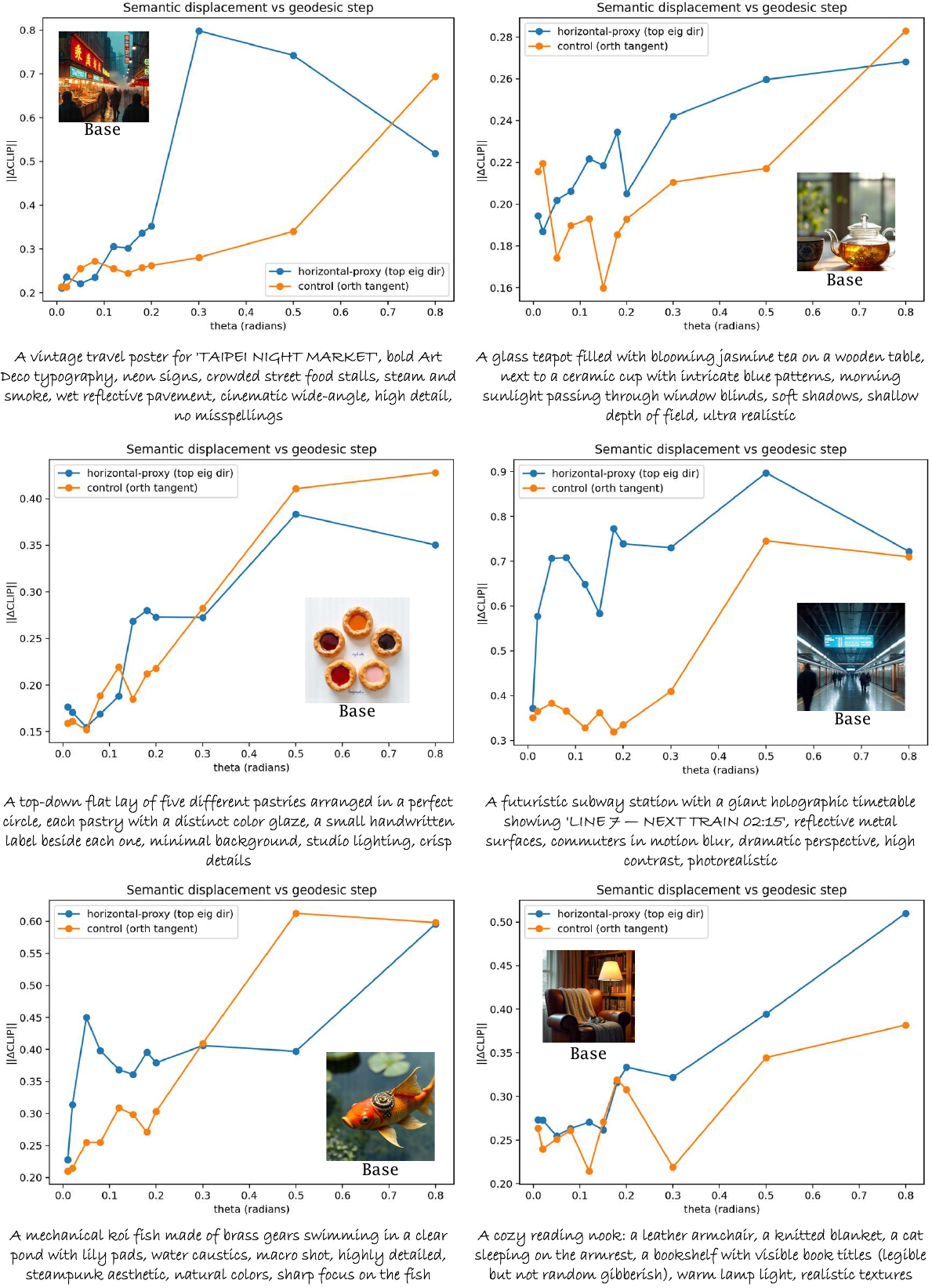}
    \caption{Semantic displacement $\|\Delta\mathrm{CLIP}\|$ vs.\ geodesic angle $\theta$.}
    \label{fig:app_hv_disp}
  \end{subfigure}
  \hfill
  \begin{subfigure}[t]{0.49\textwidth}
    \centering
    \includegraphics[width=\linewidth]{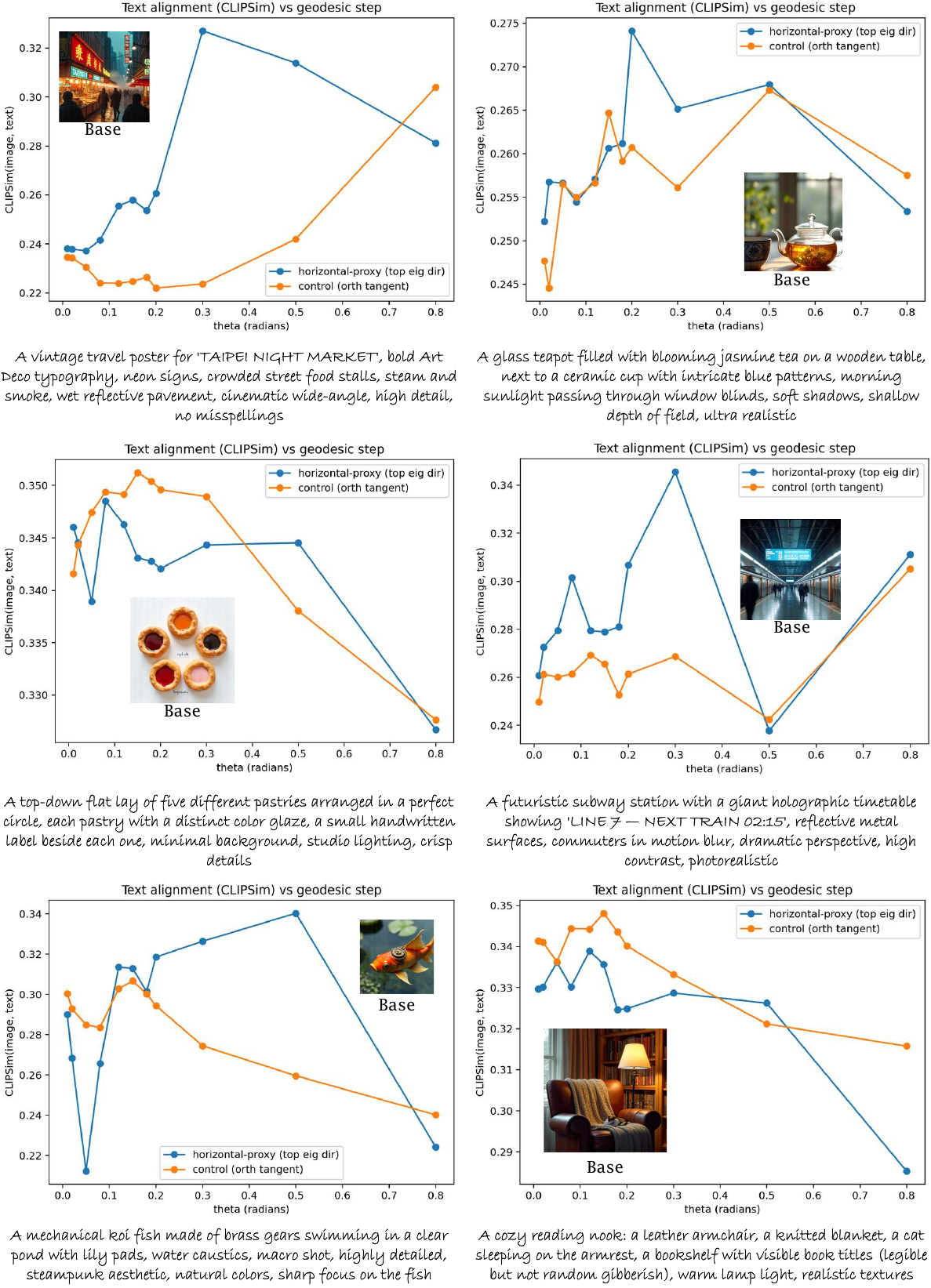}
    \caption{Alignment $\mathrm{CLIPSim}$ vs.\ geodesic angle $\theta$.}
    \label{fig:app_hv_sim}
  \end{subfigure}
  \caption{\textbf{Prompt-residual proxy vs.\ control perturbations.}
  Prompt-residual proxy directions typically yield faster semantic displacement and improved alignment at moderate angles.}
  \label{fig:app_hv_curves}
\end{figure}

The main signature is an early-slope gap: semantic displacement increases faster along $\hat u_{\mathrm H}$ than along $\hat u_{\mathrm C}$.
Alignment curves further show that moderate prompt-residual movement can improve prompt agreement, supporting the controlled injection strength used in our method.

We also visualize generation strips along increasing $\theta$.
Each column corresponds to one radius-preserving geodesic step, with $\theta=0$ as the baseline.

\begin{figure}[htbp]
  \centering
  \begin{subfigure}[t]{0.49\textwidth}
    \centering
    \includegraphics[width=\linewidth]{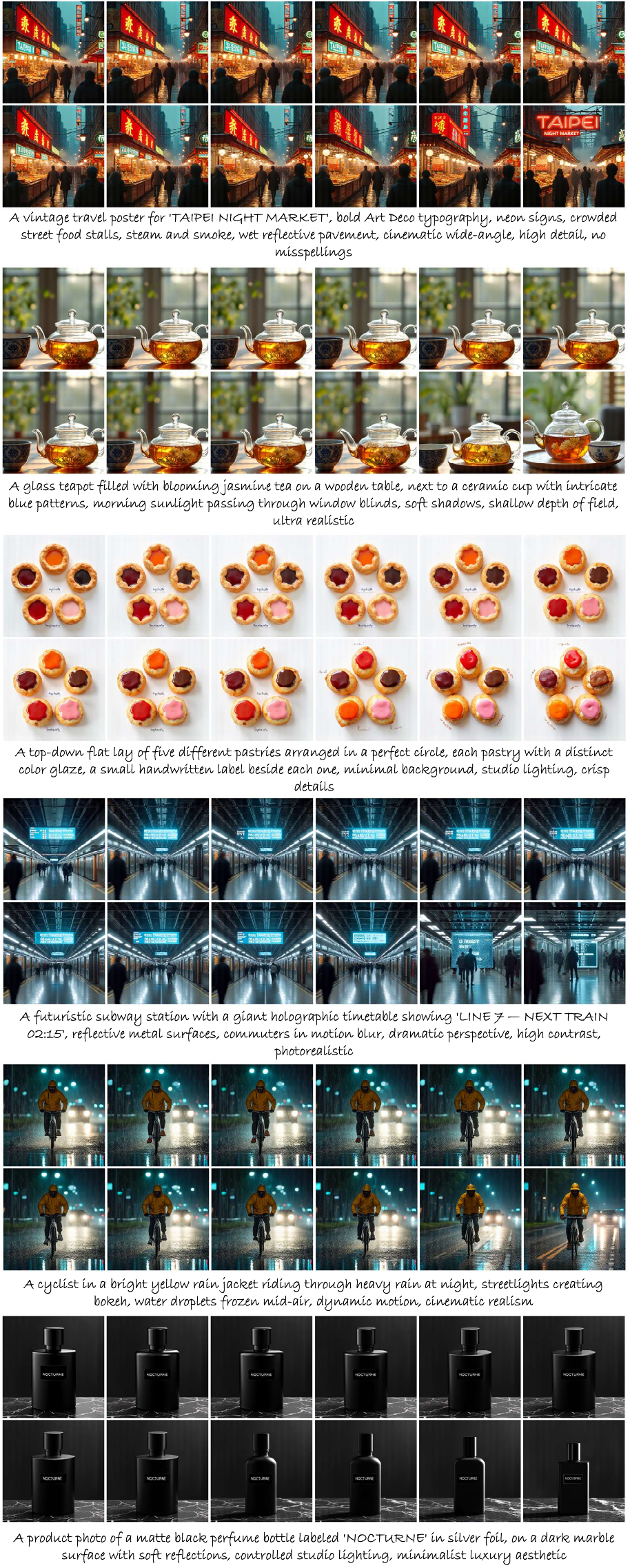}
    \caption{Control strip along $\hat u_{\mathrm C}$.}
    \label{fig:app_strip_control}
  \end{subfigure}
  \hfill
  \begin{subfigure}[t]{0.49\textwidth}
    \centering
    \includegraphics[width=\linewidth]{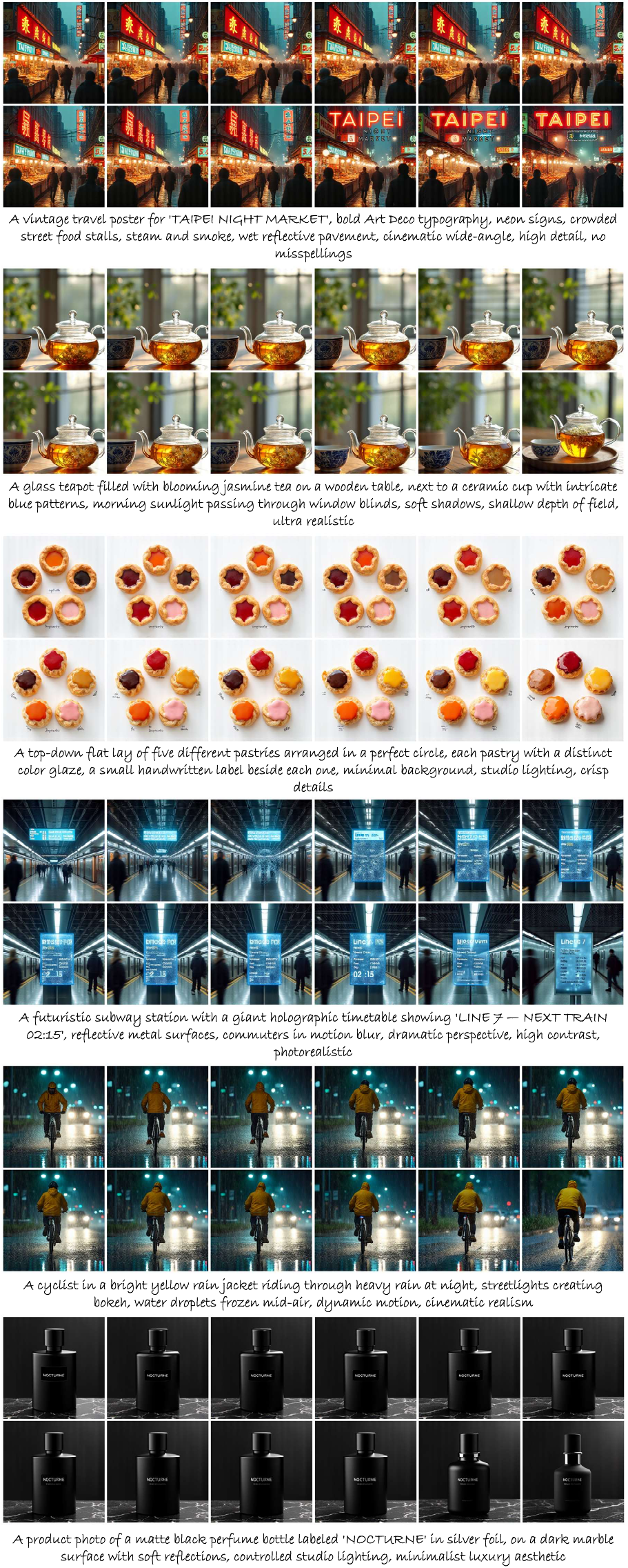}
    \caption{Prompt-residual proxy strip along $\hat u_{\mathrm H}$.}
    \label{fig:app_strip_horizontal}
  \end{subfigure}
  \caption{\textbf{Strip visualization along spherical geodesics.}
  Prompt-residual proxy movement induces coherent, prompt-relevant semantic changes earlier than random control directions.}
  \label{fig:app_strips}
\end{figure}

\subsection{Diagnostic Evaluation of Geometric Assumptions}
\label{app:diagnostic-geometry}

We further quantify the geometry diagnostics over $N=200$ prompts.
For each prompt, we generate one baseline image and one image with our method under matched settings.
We report prompt-level CLIPScore improvement
\begin{equation}
\Delta
=
\mathrm{CLIPScore}(\text{ours})
-
\mathrm{CLIPScore}(\text{baseline}).
\end{equation}
All confidence intervals are bootstrap 95\% intervals over prompts using 2,000 resamples.

\paragraph{Local Jacobian spectrum.}
We approximate the Jacobian of $\pi\circ G$ around the baseline initial noise $z_0$ using symmetric finite differences:
\begin{equation}
J_i
\approx
\frac{\pi(G(z_0+\varepsilon u_i))-\pi(G(z_0-\varepsilon u_i))}{2\varepsilon}.
\end{equation}
We use the minimum singular value $s_{\min}$ as a local geometry proxy.

\paragraph{Overall gains.}
Across 200 prompts, our method improves CLIPScore with:
\begin{itemize}
  \item Mean improvement: $\Delta=+0.312$.
  \item Bootstrap 95\% CI: $[0.153,\;0.466]$.
  \item Prompt-level win rate: $65.0\%$.
  \item Effect size: Cohen's $d=0.266$.
\end{itemize}

\paragraph{Regime analysis by $s_{\min}$.}
We stratify prompts into three quantile buckets by $s_{\min}$.
Table~\ref{tab:diag-buckets} shows that gains remain positive across all buckets, with the largest improvement in the low-$s_{\min}$ regime.

\begin{table}[t]
\centering
\small
\caption{Prompt-level gains stratified by the geometry proxy $s_{\min}$.
$\Delta$ denotes CLIPScore(ours) $-$ CLIPScore(baseline).
Confidence intervals are bootstrap 95\% over prompts within each bucket.}
\begin{tabular}{lcccc}
\toprule
Bucket by $s_{\min}$ & \# Prompts & $\mathbb{E}[\Delta]$ & 95\% CI & Win rate \\
\midrule
Low  & 67 & 0.424 & [0.150,\;0.699] & 0.716 \\
Mid  & 66 & 0.236 & [-0.016,\;0.484] & 0.606 \\
High & 67 & 0.274 & [-0.059,\;0.573] & 0.627 \\
\bottomrule
\end{tabular}
\label{tab:diag-buckets}
\end{table}

\begin{figure}[t]
  \centering
  \includegraphics[width=0.56\linewidth]{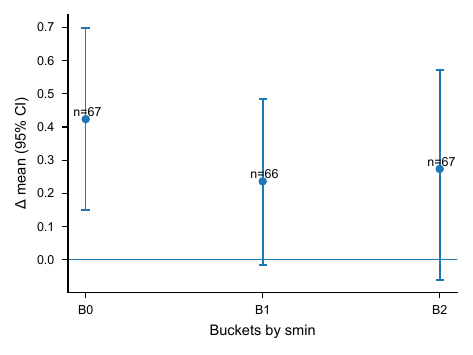}
  \caption{Bucketed analysis: mean $\Delta$ with bootstrap 95\% CI across three $s_{\min}$ quantile buckets.}
  \label{fig:diag-buckets}
\end{figure}

\subsection{Summary}
\label{app:degeneracy_viz:summary}

Together, the direction-sensitivity distribution, local spectrum proxy, prompt-residual diagnostics, geodesic curves, strip visualizations, and prompt-level gain analysis provide a consistent empirical picture:
under an isotropic Gaussian prior, semantic sensitivity is anisotropic and concentrated in structured directions.
This supports the pullback-geometry viewpoint and motivates the prompt-residual seed shaping method in Sec.~\ref{sec:method}.
% =========================
% Appendix: Diversity Analysis (No var metric)
% =========================

\section{Diversity Analysis: Alignment Gains Without Diversity Sacrifice}
\label{app:diversity}
\begin{figure}[htbp]
    \centering
    \includegraphics[width=0.6\linewidth]{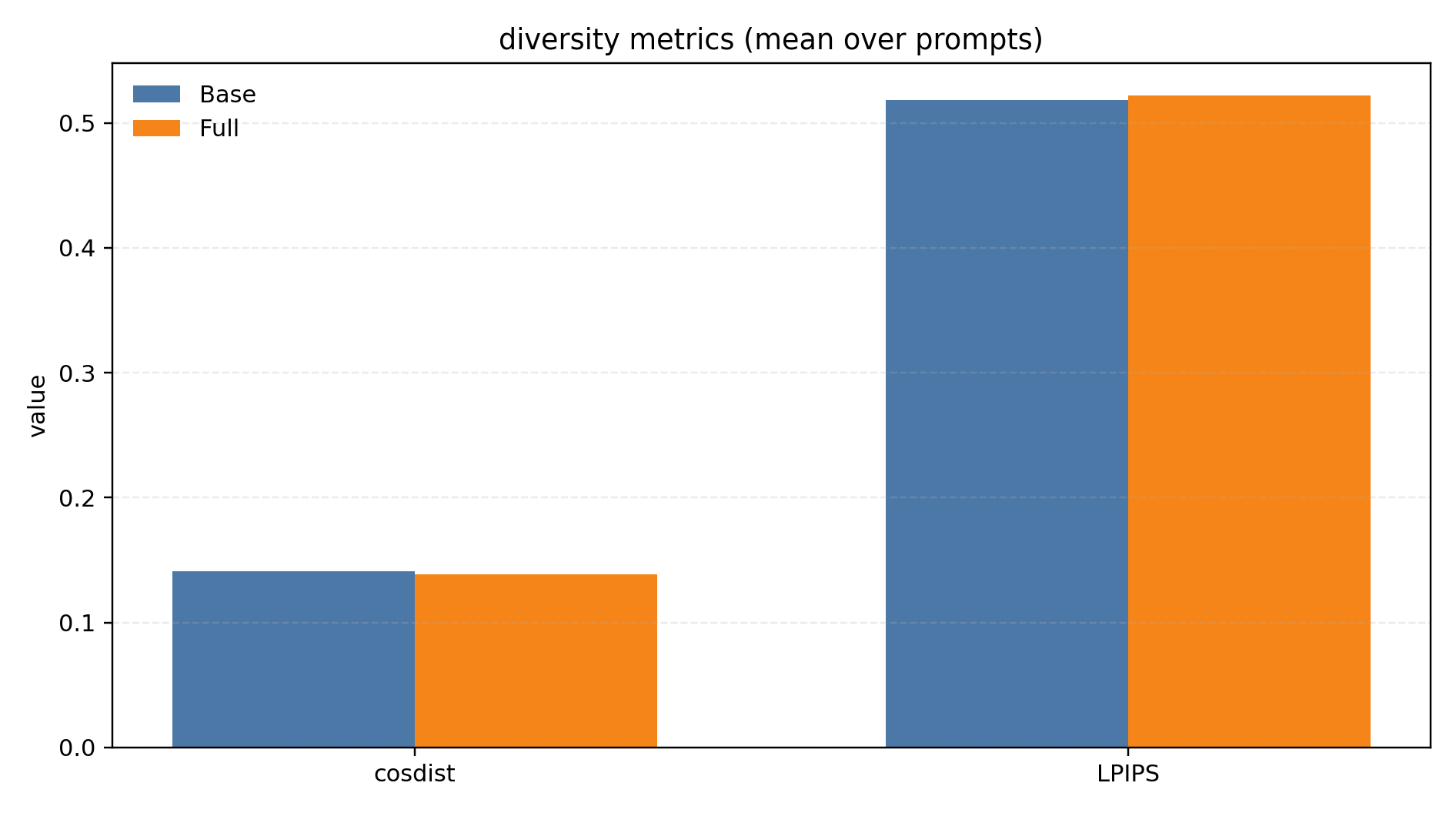}
    \caption{\textbf{Diversity metrics (mean over prompts).}
    Full improves alignment without reducing diversity: semantic diversity measured by embedding cosine distance remains essentially unchanged, while perceptual diversity measured by LPIPS is comparable or slightly higher.}
    \label{fig:diversity_metrics_novar}
\end{figure}
A potential concern for inference-time alignment improvements is that they might arise from reduced output diversity (e.g., implicitly collapsing samples to fewer modes). To rule out this explanation, we compare seed-to-seed diversity between the \textbf{Base} sampler (standard generation) and the \textbf{Full} method (our inference-time procedure). As shown in Fig.~\ref{fig:diversity_metrics_novar}, we observe no meaningful reduction in diversity under our method, suggesting that alignment gains are \emph{not} obtained by sacrificing diversity.

\subsection{Evaluation Protocol}
\label{app:diversity_protocol}
For each prompt, we generate $n$ samples using $n$ distinct random seeds under identical generation settings (model, resolution, guidance, sampler, and number of steps). For a diversity metric $d(\cdot,\cdot)$, we compute the average pairwise distance across the $n$ samples:
\begin{equation}
D(\mathcal{S}) \;=\; \frac{2}{n(n-1)} \sum_{1 \le i < j \le n} d(x_i, x_j),
\end{equation}
where $\mathcal{S}=\{x_1,\ldots,x_n\}$ denotes the set of samples for a fixed prompt. We then report the mean of $D(\mathcal{S})$ over all prompts.

\subsection{Diversity Metrics}
\label{app:diversity_metrics}
We use complementary metrics capturing semantic and perceptual diversity:
\begin{itemize}
    \item \textbf{Embedding cosine distance (cosdist).} The average cosine distance in a semantic embedding space (e.g., CLIP), capturing \emph{semantic diversity} such as changes in composition and scene semantics.
    \item \textbf{LPIPS.} A perceptual similarity metric; higher values indicate larger perceptual differences and thus greater \emph{appearance-level diversity}.
\end{itemize}

\subsection{Results and Discussion}
\label{app:diversity_results}
Figure~\ref{fig:diversity_metrics_novar} shows that diversity is preserved when switching from Base to Full:
\begin{itemize}
    \item \textbf{Semantic diversity is unchanged.} The mean embedding cosine distance is nearly identical for Base and Full (cosdist\_base vs.\ cosdist\_full), indicating that our method does not compress variation in semantic space.
    \item \textbf{Perceptual diversity is maintained (or slightly increased).} LPIPS under Full is comparable to, and marginally higher than, Base (lpips\_full vs.\ lpips\_base), suggesting that outputs do not become more similar in appearance across seeds.
\end{itemize}
Overall, these results support that the alignment improvements of Full are not explained by reduced randomness or mode collapse, but rather by structured inference-time manipulation that improves alignment while preserving seed-to-seed diversity.

% =========================
\section{Additional Diagnostics for the Single-Probe Method}
\label{app:additional_single_probe_diagnostics}
% =========================

This appendix reports two diagnostics for Algorithm~\ref{alg:sdno}. 
Our method uses a \emph{single} scheduler-consistent cold-start probe at $t^\star=900$, followed by tangential injection and spherical retraction. 
All experiments below keep this default single-probe configuration unchanged.

\subsection{Single-Probe Cold-Start Diagnostic}
\label{app:single_probe_cold_start_diag}

We directly evaluate the default configuration on FLUX.1-Dev: one cold-start residual at $t^\star=900$, one tangential update, and one radius-shell retraction. 
All settings use the same sampler, prompt set, resolution, and matched random seeds. 
The diagnostic set contains 48 prompts with 8 seeds per prompt, giving 384 matched prompt--seed cases. 
We report HPSv3 and Q20, the 20th percentile over matched cases, to measure both average quality and lower-tail robustness.

\begin{table}[htbp]
\centering
\caption{\textbf{Single-probe cold-start diagnostic on FLUX.}
The method uses the default single high-noise probe at $t^\star=900$.}
\label{tab:single_probe_cold_start_diag}
\small
\setlength{\tabcolsep}{7pt}
\renewcommand{\arraystretch}{1.05}
\begin{tabular}{@{}lccc@{}}
\toprule
Setting & Probe configuration & HPSv3 $\uparrow$ & Q20 $\uparrow$ \\
\midrule
Standard & none & 10.466 & 8.613 \\
Ours & single cold-start probe, $t^\star=900$ & 10.667 & 8.897 \\
\bottomrule
\end{tabular}
\end{table}

The single-probe configuration improves both the mean score and Q20, supporting the effectiveness of the cold-start residual as a model-coupled initialization-side correction.

\subsection{Complementarity with Feature-Side Inference Control}
\label{app:inference_complementarity}

We also test whether seed-side shaping composes with training-free feature-side inference control. 
Our method changes only the initial seed direction on the Gaussian radius shell, while feature-side controls modify denoising behavior during sampling. 
We use FreeU as a representative feature-side method and keep our single-probe procedure unchanged in the combined setting.

\begin{table}[htbp]
\centering
\caption{\textbf{Complementarity between single-probe seed shaping and feature-side inference control on FLUX.}
FreeU modifies denoising-side feature behavior, while our method modifies initialization geometry.}
\label{tab:freeu_complementarity}
\small
\setlength{\tabcolsep}{7pt}
\renewcommand{\arraystretch}{1.05}
\begin{tabular}{@{}lccc@{}}
\toprule
Setting & HPSv3 $\uparrow$ & Gain vs. standard & Q20 $\uparrow$ \\
\midrule
Standard & 10.466 & -- & 8.613 \\
FreeU only & 10.502 & $+0.036$ & -- \\
Ours only, single probe $t^\star=900$ & 10.667 & $+0.201$ & 8.897 \\
Ours $+$ FreeU & 11.167 & $+0.700$ & 9.202 \\
\bottomrule
\end{tabular}
\end{table}

The combined setting achieves the strongest mean score and Q20, indicating that the proposed seed-side intervention is compatible with denoising-side improvements.

 \section{Implementation Details}
	\label{E_D}
	\subsection{Model Details}
	\textbf{FLUX.1 dev} \cite{flux2024} is an open-weight rectified-flow transformer for high-resolution text-to-image synthesis.  
	The network stacks 19 MMDiT \cite{DBLP:conf/icml/EsserKBEMSLLSBP24} blocks that jointly attend to 4\,096 T5-XXL \cite{DBLP:journals/jmlr/RaffelSRLNMZLL20} text embeddings and a $1\,024\!\times\!1\,024$ pixel latent grid.

	\noindent\textbf{SDXL} \cite{SDXL} is a cascaded latent diffusion model that generates $1\,024\!\times\!1\,024$ images through a base$+$refiner ensemble.  
	
	\noindent\textbf{SD3.5 Medium} is a text-to-image diffusion transformer used in our sectional cone diagnostics under its official inference setting.
	
	\noindent\textbf{Z-Image} \cite{team2025zimage} is a text-to-image generation backbone used in our finite-difference cone diagnostics under its official base-model setting.
	
	\noindent\textbf{Wan 2.1 1.3B} \cite{DBLP:journals/corr/abs-2503-20314} is a bilingual diffusion transformer optimized for Chinese--English text-to-video generation.

	\noindent\textbf{Trellis text Xlarge} \cite{DBLP:conf/cvpr/XiangLXDWZC0Y25} is a sparse-voxel rectified-flow transformer that maps natural-language prompts directly to 3-D assets without 2-D distillation.

% =========================
% Appendix: Baselines
% =========================
\subsection{Baseline}
\label{app:baselines}

\textbf{InitNO} is a \emph{training-free} initial-noise optimization approach, but it is \emph{designed specifically for UNet-based diffusion pipelines} and is \emph{image-only}. It serves as the closest training-free baseline when comparing within UNet diffusion settings.

\textbf{NPNet} requires \emph{additional training} with a \emph{separately customized dataset} and is \emph{image-only}. Thus, NPNet is \emph{not a strictly comparable training-free baseline}. We include it as a representative reference point for the line of work that improves generation via \emph{learned, data-dependent optimization modules}, and we report/interpret it separately from training-free comparisons.

% Using simple marks without extra packages:
\newcommand{\cmark}{\ensuremath{\checkmark}}
\newcommand{\xmark}{\ensuremath{\times}}

\begin{table}[t]
  \centering
  \small
  \setlength{\tabcolsep}{4pt}
  \renewcommand{\arraystretch}{1.15}
  \caption{Applicability and comparability of initial-noise optimization methods.
  ``Training-free'' indicates no additional optimization of model weights and no extra training dataset.
  ``Arch.-agnostic'' means not tied to UNet-only design choices and can be instantiated across different backbones/formulations given the standard inference interface.
  \label{tab:baseline_applicability}}
  \resizebox{\linewidth}{!}{%
  \begin{tabular}{lccccccc}
    \toprule
    \textbf{Method} &
    \textbf{Training-free} &
    \textbf{Requires training} &
    \textbf{Custom dataset} &
    \textbf{Optimizes init noise} &
    \textbf{UNet diffusion} &
    \textbf{DiT Flow} &
    \textbf{Multimodal}  \\
    \midrule
    ours &
    \cmark & \xmark & \xmark & \cmark &
    \cmark & \cmark & \cmark \\
    \midrule
    InitNO &
    \cmark & \xmark & \xmark & \cmark &
    \cmark & \xmark & \xmark  \\
    \midrule
    NPNet &
    \xmark & \cmark & \cmark & \cmark/-- &
    \cmark/-- & \xmark & \xmark \\
    \bottomrule
  \end{tabular}%
  }
\end{table}

	\subsection{Benchmarks}

	\textbf{Pick-a-Pic} \cite{10.5555/3666122.3667716} benchmark is a large-scale, open dataset designed to evaluate text-to-image generation models based on real user preferences. 
	
	\noindent\textbf{DrawBench} \cite{NEURIPS2022_ec795aea} benchmark is a comprehensive evaluation suite designed to assess the performance of text-to-image generation models in rendering complex spatial relationships and compositional scenes. 
	
	\noindent\textbf{HPD} \cite{wu2023human} benchmark is a large-scale, human-annotated dataset designed to evaluate the alignment of text-to-image generative models with human aesthetic preferences.

	\noindent\textbf{Vbench} \cite{huang2023vbench} is a comprehensive evaluation suite designed to assess the performance of video generative models across a multitude of dimensions. 
	
	\noindent\textbf{Toy4K} \cite{huang2023vbench} benchmark is a synthetic 3D object dataset designed to facilitate the evaluation of 3D object recognition and generation models. 
	
	\subsection{Metrics}
	
	\noindent\textbf{Pickscore} \cite{10.5555/3666122.3667716} is a human preference prediction model designed to evaluate the alignment of images generated from textual prompts with human aesthetic judgments.  
	
	\noindent\textbf{ImageReward} \cite{DBLP:conf/nips/XuLWTLDTD23} is a human preference reward model designed to evaluate and improve text-to-image generation models by aligning their outputs with human aesthetic judgments. 
	
	\noindent\textbf{CLIPScore} \cite{DBLP:conf/emnlp/HesselHFBC21} is a reference-free evaluation tool designed to assess the alignment between generated captions and their corresponding images. 
	
	\noindent\textbf{VQAScore with Qwen-2.5-VL} \cite{lin2024evaluating, bai2025qwen25vltechnicalreport} To assess the semantic alignment between video content and textual descriptions, we employ VQAScore. This facilitates the computation of similarity between video frames and their corresponding textual prompts. 
	
	\noindent\textbf{Vbench} \cite{huang2023vbench} is a comprehensive and versatile benchmark suite designed to address the critical challenge of evaluating video generative models by decomposing ``video generation quality" into 16 specific, hierarchical, and disentangled dimensions, including subject consistency, motion smoothness, spatial relationship, and aesthetic quality. 
	
	\noindent\textbf{VQAScore with Qwen-2.5-VL} \cite{lin2024evaluating, bai2025qwen25vltechnicalreport} computes the VQAScore of rendered images and input prompts. It is averaged over multiple views to ensure consistency. Higher scores indicate better alignment, reflecting the model's ability to adhere to prompts.
	
	\noindent\textbf{Fréchet Distance \cite{DBLP:conf/nips/HeuselRUNH17} (FD) with Inception-v3 \cite{DBLP:conf/cvpr/SzegedyVISW16}} measures the similarity between feature distributions of generated and real images using Inception-v3. Lower FD values indicate that generated assets are closer to the real data distribution in feature space, reflecting higher visual quality.
	
	\noindent\textbf{Kernel Distance \cite{DBLP:journals/corr/abs-1103-1625} (KD) with Inception-v3 \cite{DBLP:conf/cvpr/SzegedyVISW16}} uses kernel methods (e.g., MMD) to compare feature distributions from Inception-v3. It assesses distribution alignment, with lower scores suggesting that generated outputs are realistic and diverse.
	
	\noindent\textbf{Fréchet Distance \cite{DBLP:conf/nips/HeuselRUNH17} (FD) with DINOv2 \cite{DBLP:journals/tmlr/OquabDMVSKFHMEA24}}
	This variant employs DINOv2 features, which capture rich semantic information. Lower FD values indicate better semantic consistency with real-world objects, leveraging the model's strong representation power.
	
	\noindent\textbf{Kernel Distance \cite{DBLP:journals/corr/abs-1103-1625} (KD) with DINOv2 \cite{DBLP:journals/tmlr/OquabDMVSKFHMEA24}}
	By applying kernel distance to DINOv2 features, this metric emphasizes structural and semantic similarities. Lower scores denote that generated assets align well with expected visual characteristics.
	
	\subsection{Model Setting}
	For the standard method, we adopted the hyperparameters presented in Table~\ref{tab:combined_inference_settings}, we increased the number of inference steps for the standard pipeline so that its total inference time matches ours, which incurs additional computational cost due to our design.
	\begin{table}[htbp]
		\centering
        \caption{Inference Settings for SDXL, FLUX.1 Dev, Wan2.1 T2V 1.3B, and TRELLIS Text XLarge}
		\begin{tabular}{lcc}
			\toprule
			\textbf{Parameter} & \textbf{SDXL} & \textbf{FLUX.1 Dev} \\
			\midrule
			Number of Inference Steps & 51 (Before: 50) & 29 (Before: 28) \\
			Guidance Scale & 7.5 & 3.5 \\
			Eta (for DDIM Scheduler) & 0.0 & - \\
			Output Image Size & $1024 \times 1024$ & $1024 \times 1024$ \\
			True CFG Scale & - & 1 \\
			\midrule
			\textbf{Parameter} & \textbf{Wan2.1 T2V 1.3B} & \textbf{TRELLIS Text XLarge} \\
			\midrule
			Resolution & 832$\times$480 (480p) & - \\
			Sample Guide Scale & 6 & - \\
			Sample Shift & 10 & - \\
			Sparse Sampler Steps & - & 26 (Before: 25) \\
			Sparse Sampler CFG Strength & - & 7.5 \\
			SLAT Sampler Steps & - & 26 (Before: 25) \\
			SLAT Sampler CFG Strength & - & 7.5 \\
			\bottomrule
		\end{tabular}
		
		\label{tab:combined_inference_settings}
	\end{table}
    \begin{table}[htbp]
\centering
\caption{Recommended hyperparameters for Algorithm~\ref{alg:sdno}.}
\label{tab:sdno_hparams}
\small
\setlength{\tabcolsep}{4pt}
\begin{tabular}{p{0.18\linewidth} p{0.18\linewidth} p{0.18\linewidth} p{0.40\linewidth}}
\hline
Hyperparam & Rec. & Range & Notes / tuning tips \\
\hline
$t^\star$ & $900\sim1000$ & high-noise timesteps &
Single cold-start probe used for the prompt residual. \\

$\delta$ & $5\sim6$ & $0\sim15$ &
Stronger injection can improve alignment but may overshoot.\\

$\rho(t)$ & model-native & model-dep. &
Use native schedule for time-consistent probing. \\

Orthogonalize & on & on/off &
Decouple from $\mathbf{z}_T$; turn off if effect weakens. \\

Normalize & on & on/off &
Stabilizes the single-step residual direction. \\
\hline
\end{tabular}
\end{table}
	\subsubsection{Hyperparameter}
\label{app:hparam}

This section analyzes the two key hyperparameters of our method: the single cold-start probe timestep $t^\star$ and the injection strength $\delta$.

\paragraph{Single cold-start probe timestep $t^\star$}
The main method uses one high-noise cold-start probe and does not aggregate residuals across timesteps. All main experiments extract the prompt residual at a fixed high-noise timestep \textbf{$t^\star=900$}. This single-step probe provides a stable semantic signal across both simple and complex prompts while requiring only one additional conditional/unconditional probing pair.

\paragraph{Injection Strength $\delta$}
The injection strength $\delta$ controls the magnitude of proxy-guided tangential injection before retraction, and thus directly trades off semantic enhancement and visual quality. We sweep $\delta$ over a wide range and evaluate performance on both simple prompts and complex prompts.

Overall, we observe that small $\delta$ leads to limited semantic gain, while overly large $\delta$ may introduce artifacts and degrade aesthetic quality. Across our tests, \textbf{$\delta=6$ consistently provides strong improvements for both simple and complex prompts while maintaining aesthetic metrics without degradation}. We therefore adopt $\delta=6$ as the default setting.

	\section{Visualization Method in Figure~\ref{ssp}}
	\label{VMsspa}
\begin{figure}[htbp]
		\centering
		\includegraphics[width=0.9\textwidth]{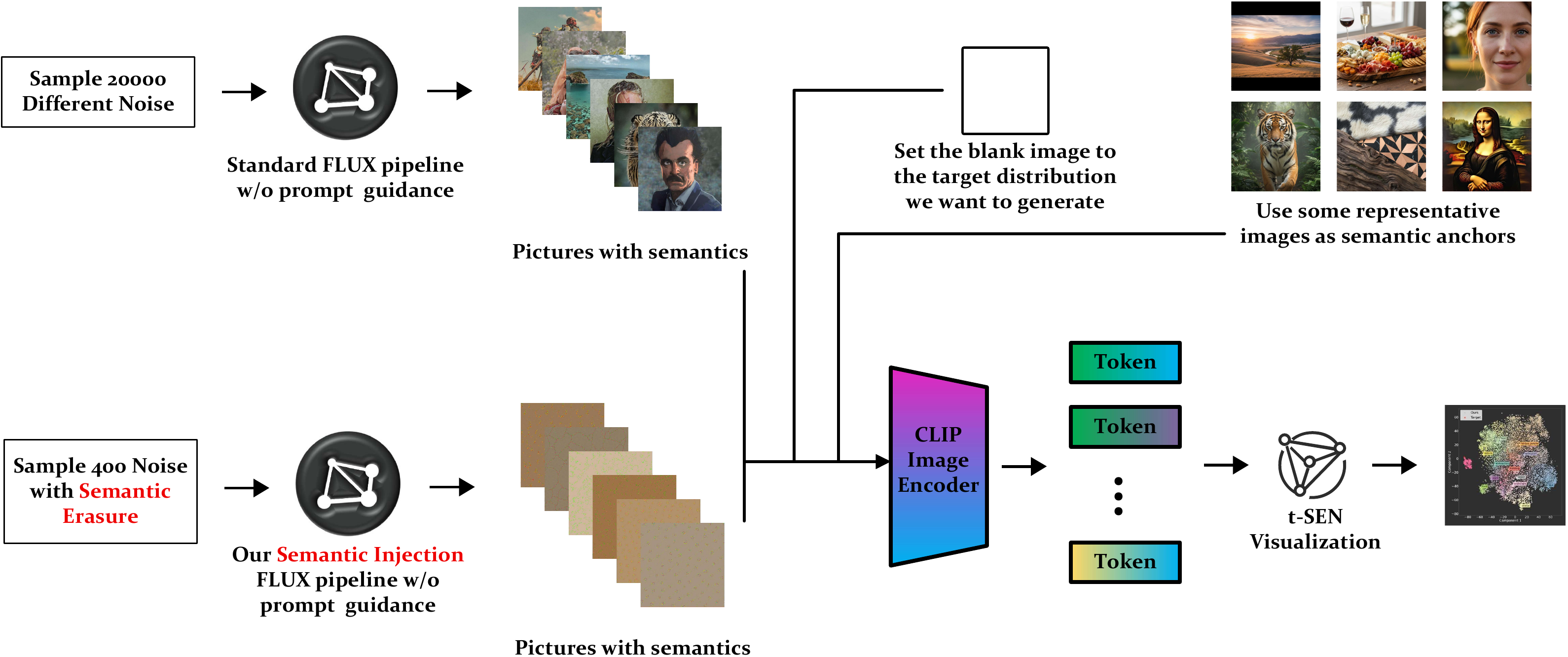}  
		\caption{Visualization Method in Fig.~\ref{ssp}.}  \label{VMssp}
	\end{figure}

	The overall visualization method in Figure~\ref{ssp} is summarized in Fig.~\ref{VMssp}.
\section{Details about User Study}
\label{US}

\subsection{Participants and Demographics}

To evaluate generation quality and prompt alignment from a human-preference perspective, we conducted a user study across image, video, and 3D generation tasks.
We report only responses from adult participants who self-reported being at least 18 years old.
Responses from participants who reported being under 18 were excluded before analysis.
The final analyzed pool contains 432 adult participants.
The study was voluntary and anonymous, and participants were asked to choose which result better matched the given textual description.

\begin{table*}[t]
    \centering
    \caption{\textbf{User study results and adult participant demographics.}}
    \label{tab:user_study_combined}

    \begin{minipage}[t]{0.50\textwidth}
        \centering
        \subcaption{\textbf{Preference results across generation modalities.}}
        \label{tab:user_study_modalities}
        \setlength{\tabcolsep}{4pt}
        \resizebox{\linewidth}{!}{
        \begin{tabular}{lcc}
            \toprule
            \textbf{Category} & \textbf{Ours (\%)} & \textbf{Standard (\%)} \\
            \midrule
            \textbf{T2I} Basic Objects \& Colors & 92.5 & 7.5 \\
            \textbf{T2I} Counterfactual Concepts & 85.0 & 15.0 \\
            \textbf{T2I} Spatial \& Textual Relations & 55.0 & 45.0 \\
            \textbf{T2I} Stylistic \& Artistic Rendering & 81.4 & 18.6 \\
            \textbf{T2I} Complex \& Specific Concepts & 72.9 & 27.1 \\
            \midrule
            \textbf{T2V} Dynamic Scenes & 66.4 & 33.6 \\
            \textbf{T2V} Static Scenes & 53.8 & 46.2 \\
            \midrule
            \textbf{Text to 3D} Detailed Objects & 58.6 & 41.4 \\
            \textbf{Text to 3D} Specialized Objects & 51.9 & 48.1 \\
            \midrule
            \textbf{Overall (T2I)} & \textbf{74.1} & 25.9 \\
            \textbf{Overall (T2V)} & \textbf{64.5} & 35.5 \\
            \textbf{Overall (Text to 3D)} & \textbf{55.3} & 44.7 \\
            \bottomrule
        \end{tabular}}
    \end{minipage}
    \hfill
    \begin{minipage}[t]{0.47\textwidth}
        \centering
        \subcaption{\textbf{Adult participant demographics.}}
        \label{tab:gender_age_distribution}
        \setlength{\tabcolsep}{4pt}
        \resizebox{\linewidth}{!}{
        \begin{tabular}{lccc}
            \toprule
            \textbf{Age Group} & \textbf{Male} & \textbf{Female} & \textbf{Percentage} \\
            \midrule
            18--25   & 64 & 42 & 24.54\% \\
            26--30   & 62 & 46 & 25.00\% \\
            31--40   & 32 & 22 & 12.50\% \\
            41--50   & 30 & 24 & 12.50\% \\
            51--60   & 38 & 32 & 16.20\% \\
            Over 60  & 26 & 12 & 8.80\% \\
            \midrule
            \textbf{Total} & \textbf{254} & \textbf{178} & \textbf{100.00\%} \\
            \bottomrule
        \end{tabular}}
    \end{minipage}
\end{table*}

\subsection{Survey Methodology}

\begin{figure}[H]
    \centering
    \includegraphics[width=0.3\textwidth]{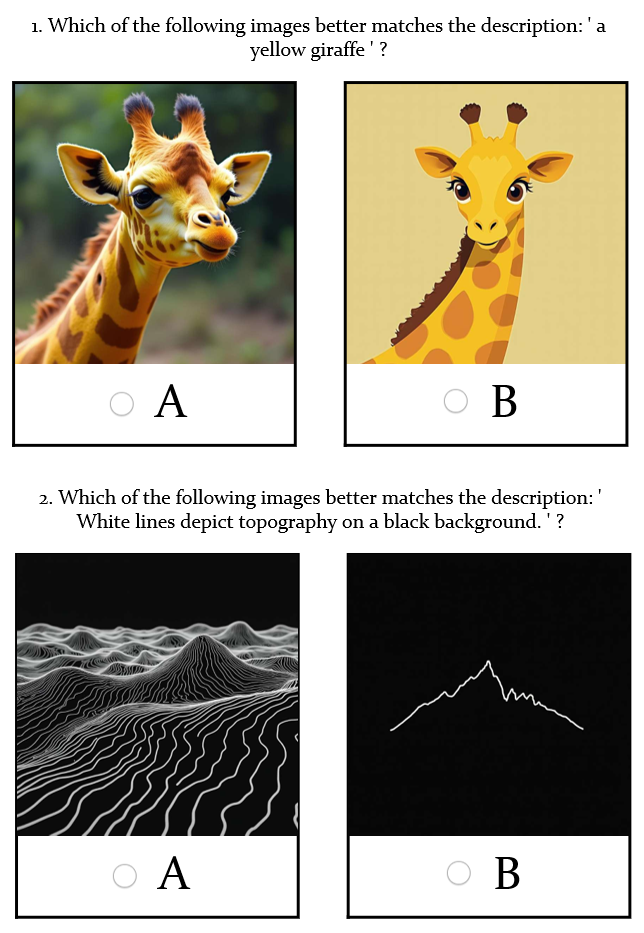}
    \caption{Schematic diagram of the questionnaire design.}
    \label{quen}
\end{figure}

As shown in Fig.~\ref{quen}, participants were presented with paired outputs generated by our method and the standard baseline, together with the corresponding textual description.
For each pair, participants were asked to select the image, video, or 3D asset that better matched the description.
The order of the two methods was randomized to reduce position bias.
The prompts were designed to evaluate several aspects of alignment, including object presence, color fidelity, spatial relationships, text rendering, motion consistency, and object-level detail.

\subsection{Results}

\subsubsection{T2I}
For text-to-image generation, our method is preferred over the standard baseline across all evaluated categories.
Preference rates range from $55.0\%$ on spatial and textual relations to $92.5\%$ on basic objects and colors, with an overall T2I preference rate of $74.1\%$.
The gains are strongest on prompts involving object presence, color fidelity, counterfactual concepts, and stylistic rendering.
Spatial and textual relation prompts remain more challenging, indicating that fine-grained layout and text rendering are still difficult failure modes.

\subsubsection{T2V}
For text-to-video generation, our method is also preferred over the standard baseline.
Preference rates are $66.4\%$ for dynamic scenes and $53.8\%$ for static scenes, with an overall T2V preference rate of $64.5\%$.
The stronger gain on dynamic prompts suggests improved text-video alignment for motion-centric scenes, while static or less visually distinctive scenes show smaller margins.

\subsubsection{Text to 3D}
For text-to-3D generation, our method obtains a moderate but consistent preference advantage.
Preference rates are $58.6\%$ for detailed objects and $51.9\%$ for specialized objects, with an overall Text-to-3D preference rate of $55.3\%$.
The results suggest that seed shaping can improve prompt adherence in 3D generation, especially for detailed object descriptions, while highly specialized objects remain challenging.
% ============================================================
% J. Limitations
% ============================================================
\section{Limitations}
\label{sec:limitations_impact}

Our framework separates three objects that are easy to conflate:
(i) the ideal geometric object induced by the semantic map,
(ii) the practical prompt-residual update used by the sampler, and
(iii) the empirical diagnostics used to test whether the practical update behaves like the ideal object.
This separation is important because our method is not intended to recover the exact semantic-horizontal subspace of the diffusion model.
Rather, it provides a training-free, inference-time approximation whose validity is supported empirically under the tested models, prompts, and schedulers.

\paragraph{The ideal horizontal space is not available at inference time.}
The geometric analysis in Section~\ref{sec:pre} characterizes the effective semantic map
$\Phi=\pi\circ\Psi_{T\to0}$ in Eq.~\eqref{eq:semantic_map}.
At a regular point, the Jacobian $J_z=D\Phi_z$ induces the vertical--horizontal decomposition in Eq.~\eqref{eq:HV_split}, where vertical directions are locally semantic-invariant and horizontal directions are locally semantic-sensitive.
On the Gaussian radius shell, this induces the shell split in Eq.~\eqref{eq:shell_HV_split}, with shell-horizontal directions lying in $\Img(Q_zJ_z^\top)$, where the shell projector $Q_z$ is defined in Eq.~\eqref{eq:shell_projector}.
If this shell-horizontal space were directly accessible, one could in principle design seed interventions that move preferentially along semantic-sensitive directions while preserving the Gaussian radius shell.

This ideal intervention can be written explicitly.
For the prompt-semantic potential $F_y(z)=\ell_y(\Phi(z))$ defined in Eq.~\eqref{eq:prompt_objective}, the geometry-matched shell direction is the shell gradient in Eq.~\eqref{eq:shell_prompt_gradient}:
\[
g_{\rm oracle}(z)
:=
\nabla_{\mathbb S_R}F_y(z)
=
Q_zJ_z^\top\nabla_\Phi\ell_y(\Phi(z)).
\]
Equivalently, if the semantic metric is kept explicit rather than absorbed into $\nabla_\Phi\ell_y$, this direction can be written as
\[
g_{\rm oracle}(z)
=
Q_zD(\pi\circ\Psi_{T\to0})_z^\top
G_{\mathcal M}(\Phi(z))
\nabla_\Phi\ell_y(\Phi(z)).
\]
This direction is exactly shell-horizontal by Lemma~\ref{lem:shell_prompt_gradient}.
The corresponding radius-preserving oracle update would use the same spherical retraction structure as Eq.~\eqref{eq:orthogonalize}:
\[
z^+
=
\|z\|\,
\frac{
z+\delta\, g_{\rm oracle}(z)/(\|g_{\rm oracle}(z)\|+\varepsilon)
}{
\left\|
z+\delta\, g_{\rm oracle}(z)/(\|g_{\rm oracle}(z)\|+\varepsilon)
\right\|+\varepsilon
}.
\]
By the first-order shell expansion in Eqs.~\eqref{eq:first_order_retraction}--\eqref{eq:first_order_objective_expansion}, this oracle update performs first-order ascent on $F_y$ for sufficiently small $\delta$, while remaining on the original Gaussian radius shell.

In practice, however, this oracle direction is computationally inaccessible.
Evaluating $\Phi$ in Eq.~\eqref{eq:semantic_map} at a single seed already requires a complete denoising trajectory followed by a semantic-encoder forward pass.
Computing $D\Phi_z^\top$ would require differentiating through the entire sampler and semantic encoder, or approximating the Jacobian with many radius-preserving finite-difference generations.
Both options are far more expensive than the intended use case: shaping one initial seed before generation.
Moreover, $D\Phi_z$ is local to a particular seed, prompt, scheduler, and model state, so a Jacobian computed at one point does not provide a reusable horizontal space for other seeds.
For this reason, our algorithm should not be interpreted as an exact estimator of $\mathcal H_z$ or $\mathcal H_z^{\rm sh}$.
Instead, it should be viewed as a low-cost approximation to the geometry-matched oracle direction above.

\paragraph{The prompt residual is a proxy, not an oracle.}
Our practical method replaces the unavailable oracle direction with a model-coupled prompt-residual proxy.
Specifically, we use the conditional--unconditional residual in Eq.~\eqref{eq:cfg_residual}, evaluated at the high-noise cold-start state in Eq.~\eqref{eq:single_probe_proxy}.
The motivation is that, under the local decomposition in Eq.~\eqref{eq:factorization_to_proxy}, the prompt residual contains a component aligned with the pullback of a prompt-semantic gradient, plus an error term $e_t$.
After shell projection, Eq.~\eqref{eq:shell_proxy_decomposition} gives
\[
Q_zg_\theta(z,t,y)
=
c_t\nabla_{\mathbb S_R}F_y(z)+Q_ze_t,
\]
so the implemented direction can be interpreted as a noisy approximation to the ideal shell gradient in Eq.~\eqref{eq:shell_prompt_gradient}.
The actual update then removes the radial component and retracts the seed to its original Gaussian radius shell as in Eq.~\eqref{eq:orthogonalize}.

This argument has an important limitation: Eq.~\eqref{eq:factorization_to_proxy} is a modeling assumption, not a runtime guarantee.
The residual may contain components unrelated to the semantic-horizontal space, and the residual-to-gradient relationship can vary across model families, parameterizations, schedulers, guidance scales, and prompt types.
Accordingly, we do not claim that the prompt residual exactly recovers the true horizontal direction.
Our claim is more limited: for the tested backbones and settings, the shell-projected residual exhibits the empirical signatures expected of a semantic-horizontal proxy.

\paragraph{The diagnostics provide evidence, not a formal certificate.}
Because $D\Phi_z$ is unavailable, the true vertical and horizontal spaces in Eq.~\eqref{eq:HV_split} and Eq.~\eqref{eq:shell_HV_split} cannot be directly measured.
We therefore use finite-difference semantic probes to construct an empirical local section of the radius shell and test whether the prompt-residual direction behaves differently from random tangent controls.
The cone diagnostics in Section~\ref{sec:exp:proxy_diagnostics} and Appendix~\ref{app:proxy_diagnostics} measure whether the shell-projected residual has low estimated vertical leakage, high concentration in the empirical semantic-sensitive subspace, and larger semantic displacement than random tangent directions.

These diagnostics are useful because they test the empirical version of the cone condition in Eq.~\eqref{eq:shell_cone_condition}.
In practice, we estimate the finite-dimensional cone ratio $\widehat\rho_k$ using Eq.~\eqref{eq:rho_hat_main}.
However, this remains an empirical diagnostic rather than a formal certificate.
The estimate depends on the chosen semantic projection $\pi$, the finite-difference scale, the number of tangent probes, and the evaluator used to measure semantic displacement.
Thus, values such as $\widehat\rho_k<0.2$, horizontal energy above $96\%$, and semantic-displacement percentile $1.000$ should be interpreted as strong empirical support for the proxy in the evaluated regimes, not as a mathematical certificate that the exact horizontal space has been recovered.

\paragraph{Failure modes and operating boundaries.}
The method can degrade when the prompt-residual proxy no longer approximates a semantic-horizontal direction.
We identify several such regimes.

\begin{enumerate}
    \item \textbf{Extreme classifier-free guidance.}
    Very large guidance scales can amplify the residual error term $e_t$ in Eq.~\eqref{eq:factorization_to_proxy} and push the update outside the local regime where the shell approximation is meaningful.
    In this case, the method may overshoot, distort structure, or reduce quality.
    The guidance stress tests in Table~\ref{tab:boundary_cost} illustrate this operating boundary.

    \item \textbf{Out-of-distribution backbones or schedulers.}
    The relationship between the prompt residual in Eq.~\eqref{eq:cfg_residual} and the shell-gradient proxy in Eq.~\eqref{eq:shell_proxy_decomposition} can change across model architectures, parameterizations, timestep schedules, and noise scalings.
    A probe timestep and update strength that work for one backbone may therefore require retuning for another.

    \item \textbf{Large update strengths.}
    The bridge in Theorem~\ref{thm:algorithmic_bridge} is local.
    If the tangential injection strength $\delta$ in Eq.~\eqref{eq:orthogonalize} is too large, the update may leave the neighborhood in which the residual is a reliable proxy.
    This is consistent with the first-order expansion in Eq.~\eqref{eq:first_order_objective_expansion} and with the ablation in Figure~\ref{time}, where moderate values of $\delta$ are beneficial while overly large values can degrade results.
\end{enumerate}

In these regimes, the method should be treated as a bounded local seed correction rather than a general steering mechanism.
When applying the method to a new model family or prompt distribution, we recommend re-running the cone diagnostics in Section~\ref{sec:exp:proxy_diagnostics} and strength ablations in Section~\ref{sec:exp:ablate} before drawing conclusions about effectiveness.

\paragraph{Scope of the contribution.}
These limitations do not undermine the central contribution, but they clarify its scope.
The geometric analysis explains why an isotropic Gaussian prior can induce highly anisotropic semantic behavior after generation: the semantic map in Eq.~\eqref{eq:semantic_map} induces the degenerate pullback geometry in Eqs.~\eqref{eq:pullback_metric}--\eqref{eq:pullback_matrix}, with semantic-sensitive directions concentrated in the horizontal component of Eq.~\eqref{eq:HV_split}.
The geometry-matched oracle update above shows the exact seed intervention implied by this theory, while also making clear why it is impractical for standard inference.
Our algorithm provides a cheap prompt-residual approximation to this unavailable shell-horizontal direction via Eqs.~\eqref{eq:cfg_residual}--\eqref{eq:shell_proxy_decomposition} and implements the final prior-compatible update through Eq.~\eqref{eq:orthogonalize}.
The diagnostics test whether this approximation behaves as predicted in practice.
Together, they support a practical and empirically validated seed-shaping procedure, while stopping short of claiming exact recovery of the latent semantic geometry or universal improvement across all generation regimes.
% ---------------- Optional: a compact comparison table ----------------
\begin{table}[t]
\centering
\caption{\textbf{Main generation results across image, video, and 3D generation.}
$\Delta$ and Win are computed against Standard using paired bootstrap over matched cases.
The primary metric is PickScore for image generation and VQA for video/3D generation.}
\label{tab:all_generation_results}

\begin{minipage}[t]{0.5\linewidth}
\vspace{0pt}
\centering
\subcaption{\textbf{Image generation.}}
\label{tab:afull_sdxl_main_exp}

\tiny
\renewcommand{\arraystretch}{1.2}
\setlength{\tabcolsep}{2pt}
\resizebox{\linewidth}{!}{%
\begin{tabular}{@{}lllccccc@{}}
\toprule
Model & Dataset & Method & Pick $\uparrow$ & ImgR $\uparrow$ & CLIP $\uparrow$
& $\Delta$ Pick [95\% CI] & Win \\
\midrule
SDXL & Pick-a-Pic & Standard
& 17.049 & -1.971 & 16.223 & -- & -- \\
SDXL & Pick-a-Pic & Initno$^\dagger$ \cite{DBLP:conf/cvpr/GuoLCLY024}
& 17.050 & -1.969 & 16.237 & -- & -- \\
SDXL & Pick-a-Pic & NPNet \cite{zhou2025golden}
& 17.051 & -1.968 & 16.250 & -- & -- \\
SDXL & Pick-a-Pic & Ours
& \textbf{17.371} & \textbf{-1.852} & \textbf{16.642}
& \textbf{+0.322 [0.241, 0.404]} & \textbf{68.4\%} \\
\midrule

SDXL & DrawBench & Standard
& 17.416 & -2.084 & 16.618 & -- & -- \\
SDXL & DrawBench & Initno$^\dagger$ \cite{DBLP:conf/cvpr/GuoLCLY024} 
& 17.419 & -2.085 & 16.632 & -- & -- \\
SDXL & DrawBench & NPNet \cite{zhou2025golden}
& 17.420 & -2.084 & 16.640 & -- & -- \\
SDXL & DrawBench & Ours
& \textbf{17.619} & \textbf{-1.984} & \textbf{16.957}
& \textbf{+0.203 [0.089, 0.314]} & \textbf{63.1\%} \\
\midrule

SDXL & HPD & Standard
& 16.748 & -1.955 & 14.986 & -- & -- \\
SDXL & HPD & Initno$^\dagger$ \cite{DBLP:conf/cvpr/GuoLCLY024} 
& 16.765 & -1.952 & 15.010 & -- & -- \\
SDXL & HPD & NPNet \cite{zhou2025golden}
& 16.780 & -1.948 & 15.040 & -- & -- \\
SDXL & HPD & Ours
& \textbf{17.041} & \textbf{-1.875} & \textbf{15.851}
& \textbf{+0.293 [0.171, 0.417]} & \textbf{66.2\%} \\
\midrule

FLUX & Pick-a-Pic & Standard
& 17.134 & -1.942 & 16.295 & -- & -- \\
FLUX & Pick-a-Pic & NPNet \cite{zhou2025golden}
& 17.140 & -1.936 & 16.305 & -- & -- \\
FLUX & Pick-a-Pic & Ours
& \textbf{17.732} & \textbf{-1.905} & \textbf{16.914}
& \textbf{+0.598 [0.481, 0.716]} & \textbf{74.5\%} \\
\midrule

FLUX & DrawBench & Standard
& 17.275 & -2.051 & 16.642 & -- & -- \\
FLUX & DrawBench & NPNet \cite{zhou2025golden}
& 17.282 & -2.044 & 16.655 & -- & -- \\
FLUX & DrawBench & Ours
& \textbf{17.626} & \textbf{-1.968} & \textbf{17.013}
& \textbf{+0.351 [0.236, 0.462]} & \textbf{69.8\%} \\
\midrule

FLUX & HPD & Standard
& 16.730 & -1.951 & 15.021 & -- & -- \\
FLUX & HPD & NPNet \cite{zhou2025golden} 
& 16.750 & -1.944 & 15.900 & -- & -- \\
FLUX & HPD & Ours 
& \textbf{17.239} & \textbf{-1.874} & \textbf{16.731}
& \textbf{+0.509 [0.374, 0.641]} & \textbf{72.1\%} \\
\bottomrule
\end{tabular}%
}
\end{minipage}
\hfill
\begin{minipage}[t]{0.45\linewidth}
\vspace{0pt}
\centering

\subcaption{\textbf{Video generation.}}
\label{tab:full_video_generation}
\vspace{0.5mm}

\tiny
\renewcommand{\arraystretch}{0.8}
\setlength{\tabcolsep}{2pt}
\resizebox{\linewidth}{!}{%
\begin{tabular}{@{}lccccc@{}}
\toprule
Method & AQ $\uparrow$ & AS $\uparrow$ & BC $\uparrow$ & DD $\uparrow$ & IQ $\uparrow$ \\
\midrule
Standard & 0.572 & 0.220 & 0.963 & 0.644 & 0.644 \\
Ours
& \textbf{0.651} & \textbf{0.341} & \textbf{0.985} & \textbf{0.701} & \textbf{0.664} \\
\midrule
Method & MS $\uparrow$ & OC $\uparrow$ & SC $\uparrow$ & TF $\uparrow$ & VQA $\uparrow$ \\
\midrule
Standard & 0.972 & 0.244 & 0.937 & 0.984 & 0.534 \\
Ours
& \textbf{0.997} & \textbf{0.303} & \textbf{0.994} & \textbf{0.997} & \textbf{0.600} \\
\midrule
\multicolumn{2}{@{}l}{Primary metric} & \multicolumn{2}{c}{$\Delta$ VQA [95\% CI]} & \multicolumn{2}{c@{}}{Win} \\
\multicolumn{2}{@{}l}{Ours vs. Standard}
& \multicolumn{2}{c}{\textbf{+0.066 [0.041, 0.091]}}
& \multicolumn{2}{c@{}}{\textbf{71.3\%}} \\
\bottomrule
\end{tabular}%
}

\vspace{2mm}

\subcaption{\textbf{3D generation.}}
\label{tab:full_3d_generation}
\vspace{2mm}

\tiny
\renewcommand{\arraystretch}{0.8}
\setlength{\tabcolsep}{2pt}
\resizebox{\linewidth}{!}{%
\begin{tabular}{@{}lccccc@{}}
\toprule
Method & FD$_I$ $\downarrow$ & KD$_I$ $\downarrow$ & FD$_D$ $\downarrow$ & KD$_D$ $\downarrow$ & VQA $\uparrow$ \\
\midrule
Standard & 29.544 & 0.011 & 340.530 & 0.802 & 0.876 \\
Ours
& \textbf{29.403} & \textbf{0.008} & \textbf{335.425} & \textbf{0.733} & \textbf{0.911} \\
\midrule
\multicolumn{2}{@{}l}{Primary metric} & \multicolumn{2}{c}{$\Delta$ VQA [95\% CI]} & \multicolumn{2}{c@{}}{Win} \\
\multicolumn{2}{@{}l}{Ours vs. Standard}
& \multicolumn{2}{c}{\textbf{+0.035 [0.018, 0.053]}}
& \multicolumn{2}{c@{}}{\textbf{64.7\%}} \\
\bottomrule
\end{tabular}%
}

\vspace{0.5mm}
{\tiny $I$: Inception; $D$: DINOv2.}
\end{minipage}
\end{table}

	\clearpage
	\section{Additional Main Experiment Results}
	\label{ER}
	
	\begin{figure}[htbp]
		\centering
\includegraphics[width=0.9\textwidth]{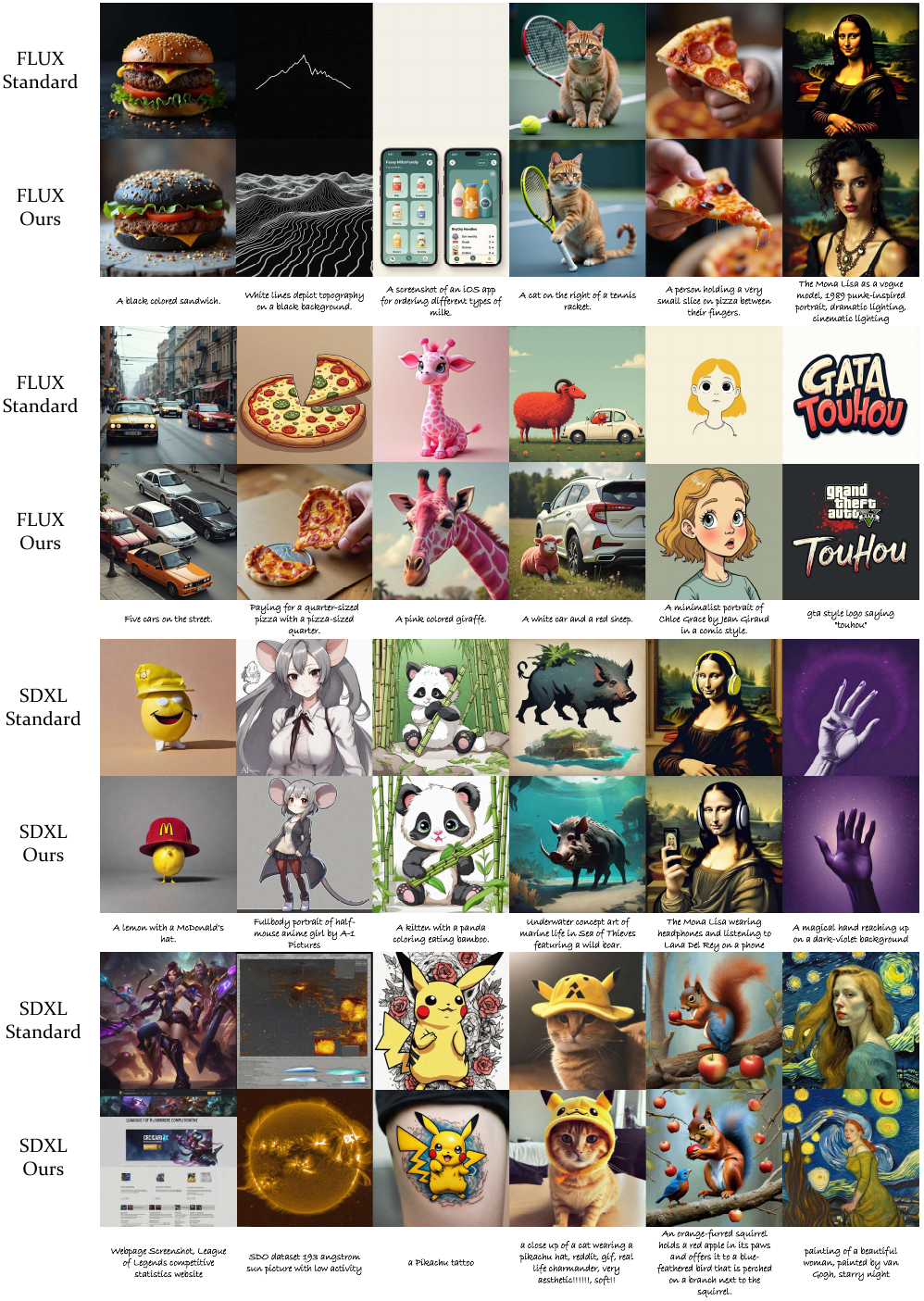}  
		\caption{More Results of SDXL and FLUX}  \label{s_m}
	\end{figure}
	
	\clearpage
\newpage
	\begin{figure}[t]
		\centering
		\includegraphics[width=\textwidth]{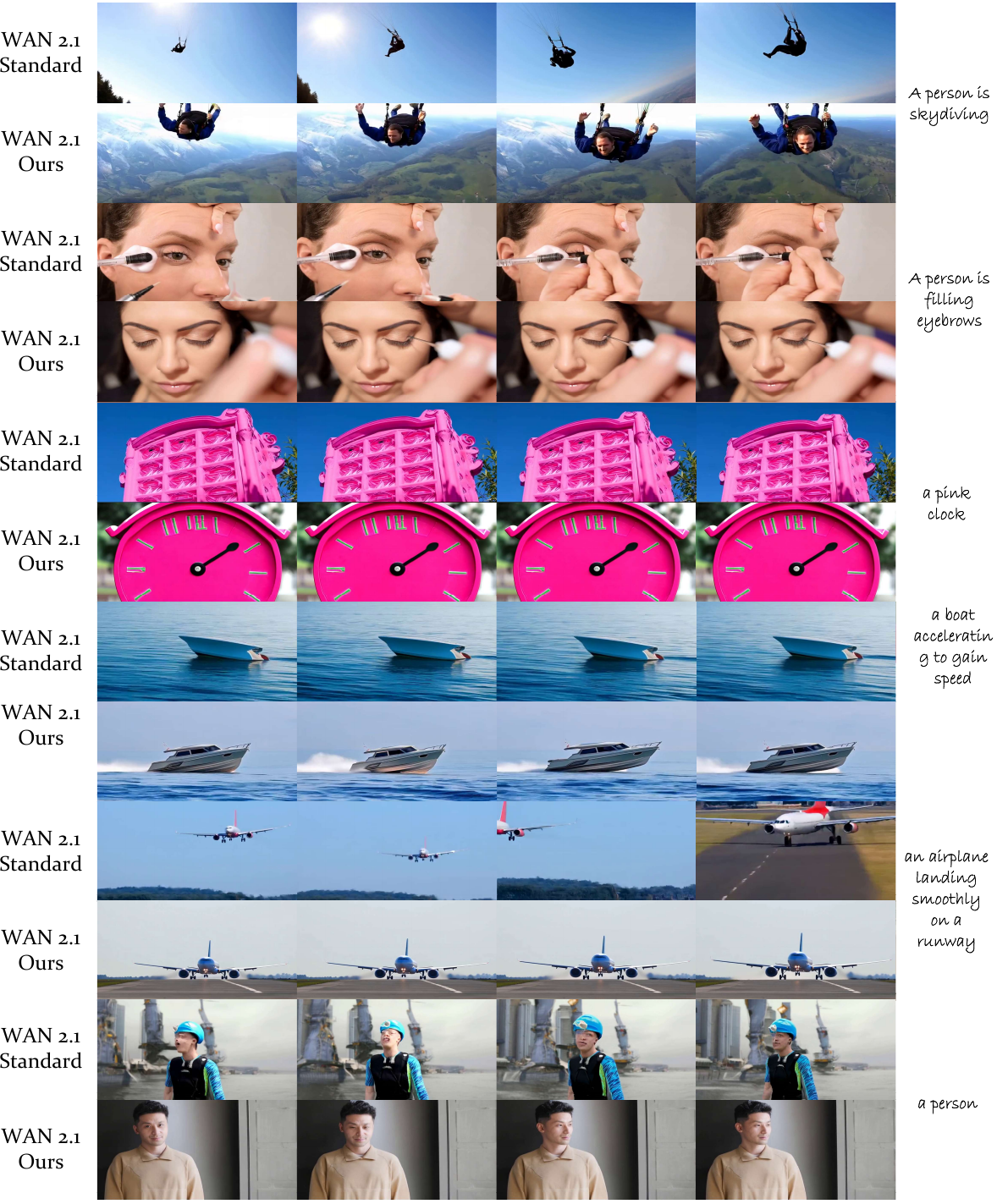}  
		\caption{More Results of WAN 2.1 1.3B}  \label{fig:wan_more1}
	\end{figure}
	
	\clearpage
\newpage	
	\begin{figure}[t]
		\centering
		\includegraphics[width=\textwidth]{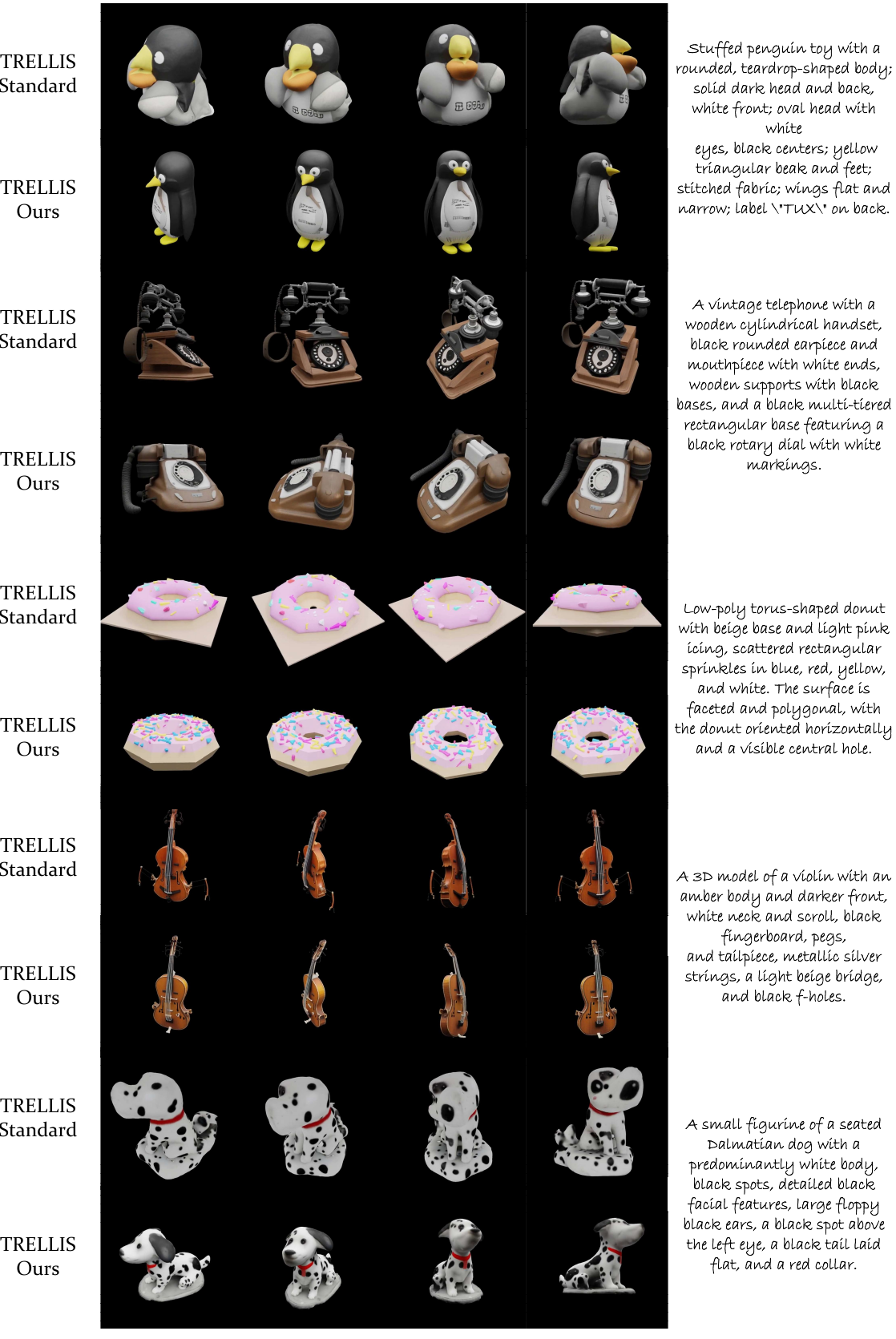}  
		\caption{More Results of TRELLIS Text Xlarge}  \label{3d_m1}
	\end{figure}

\clearpage

\end{document}